\documentclass{article}

\usepackage{microtype}
\usepackage{graphicx}
\usepackage{subcaption}
\usepackage{booktabs} 

\usepackage[preprint]{icml2026}

\usepackage{amsmath}
\usepackage{amssymb}
\usepackage{mathtools}
\usepackage{amsthm}

\usepackage[textsize=tiny]{todonotes}

\icmltitlerunning{Performative Policy Gradient}

\usepackage[colorlinks=true,urlcolor=blue,linkcolor=red,citecolor=teal,backref=page]{hyperref}
\usepackage{array}
\usepackage{makecell}
\usepackage{float,xcolor,todonotes,tcolorbox,longtable}
\usepackage{mathtools}
\usepackage{amsmath, amssymb, natbib, graphicx, url}
\usepackage{algorithm}
\usepackage{algorithmic}
\usepackage{wrapfig}
\usepackage{booktabs}
\newcommand{\subopt}{\mathrm{SubOpt}}
\usepackage{tcolorbox}
\usepackage{dsfont,enumitem}
\usepackage{footnote}
\usepackage{caption}
\usepackage{subcaption}
\usepackage{mwe}
\usepackage{cancel}
\usepackage{multirow}

\makesavenoteenv{tabular}
\makesavenoteenv{table}

\newcommand{\bx}{\mathbf{x}}
\newcommand{\by}{\mathbf{y}}

\newcommand \defn {\mathrel{\triangleq}}
\newcommand \dd {\,\mathrm{d}}
\DeclareMathOperator*{\argmax}{arg\,max}
\DeclareMathOperator*{\argmin}{arg\,min}

\newcommand \prob {\mathop{\mbox{\ensuremath{\mathbb{P}}}}\nolimits}

\newcommand \norm[1]{\left\|#1\right\|}

\newcommand \KL[2] {D_{\mathrm{KL}}\left( #1 ~\middle\|~ #2\right)}
\newcommand \TV[2] {\mathrm{TV}\left( #1 ~\middle\|~ #2\right)}

\newcommand{\identity}{\mathbf{I}}
\newcommand{\indicator}{\mathds{1}}

\newcommand{\opt}{\bpi^{\star}}

\newcommand{\reals}{\mathbb{R}}

\newcommand{\simplex}{\ensuremath{\Delta}_K}

\newcommand{\entropy}{\mathcal{H}}
\newcommand \occupancy[2]{\boldsymbol{d}_{#1}^{#2}}
\newcommand{\softV}{\tilde V}
\newcommand{\softA}{\tilde A}

\usepackage{amsthm}
\newtheorem{theorem}{Theorem}
\newtheorem{corollary}{Corollary}
\newtheorem{lemma}{Lemma}

\newtheorem{definition}{Definition}

\newtheorem{remark}{Remark}

\newtheorem{example}{Example}
\newtheorem{assumption}{Assumption}

\makeatletter
\newtheorem*{rep@theorem}{\rep@title}
\newcommand{\newreptheorem}[2]{%
	\newenvironment{rep#1}[1]{%
		\def\rep@title{\textbf{#2} \ref{##1}}%
		\begin{rep@theorem}}%
		{\end{rep@theorem}}}
\makeatother
\newreptheorem{theorem}{Theorem}
\newreptheorem{lemma}{Lemma}
\newreptheorem{proposition}{Proposition}
\newreptheorem{assumption}{Assumption}
\newreptheorem{corollary}{Corollary}

\DeclareRobustCommand{\bigO}{\text{\usefont{OMS}{cmsy}{m}{n}O}}
\allowdisplaybreaks%
\newif\ifsinglecol
\singlecolfalse

\usepackage[nohints]{minitoc}

\usepackage{mathrsfs}
\DeclareRobustCommand{\bigO}{%
  \text{\usefont{OMS}{cmsy}{m}{n}O}%
}
\newcommand{\deb}[1]{\todo[inline,color=blue!20]{{\textbf{DB:~}#1}}}

\newcommand{\udu}[1]{\todo[inline,color=yellow!20]{{\textbf{UDU:~}#1}}}

\usepackage{tikz}
\usetikzlibrary{automata}
\usetikzlibrary{bayesnet, arrows, calc, shapes, backgrounds,arrows,decorations.pathmorphing,fit,positioning}
\tikzset{
   container/.style = {rectangle, rounded corners, draw=yellow, dashed,
fit=#1, inner sep=6mm, node contents={}},
circle-label/.style = {circle, draw}
        }
\tikzset{box/.style={draw, diamond, thick, text centered, minimum height=0.5cm, minimum width=1cm, text width=0.9cm}}
\tikzset{line/.style={draw, thick, -latex'}}
\allowdisplaybreaks
\usepackage{pgfplots}

\usepackage{layouts}

\def\bV{\mathbf{V}}

\def\blambda{{\boldsymbol\lambda}}

\def\be{\mathbf{e}}

\def\bu{\mathbf{u}}

\def\bx{\mathbf{x}}
\def\by{\mathbf{y}}

\def\btheta{{\boldsymbol\theta}}

\def\blambda{{\boldsymbol\lambda}}

\def\bnu{{\boldsymbol\nu}}

\def\bpi{{\boldsymbol\pi}}
\def\brho{{\boldsymbol\rho}}

\def\cA{\mathcal{A}}

\def\cD{\mathcal{D}}

\def\cM{\mathcal{M}}

\def\cS{\mathcal{S}}

\def\Cov{{\sf Cov}}

\DeclareMathOperator*{\expectation}{\mathbb{E}}
\newcommand{\transition}{\mathbf{P}}
\newcommand{\reward}{r}
\newcommand{\lipreward}{L_\reward}
\newcommand{\liptransition}{L_\transition}
\newcommand{\liphellinger}{L_H}

\newcommand \hellinger[2] {D_{\mathrm{H}}\left( #1 \| #2\right)}

\newcommand{\framework}{{\color{blue}\ensuremath{\mathsf{PePG}}}}

\newcommand{\softreward}{\tilde{\reward}}

\begin{document}

\twocolumn[
  \icmltitle{Performative Policy Gradient:\\ Optimality in Performative Reinforcement Learning}



  \icmlsetsymbol{equal}{*}

  \begin{icmlauthorlist}
    \icmlauthor{Debabrota Basu}{inria}
    \icmlauthor{Udvas Das}{inria}
    \icmlauthor{Brahim Driss}{inria}
    \icmlauthor{Uddalak Mukherjee}{isi}
  \end{icmlauthorlist}

  \icmlaffiliation{inria}{Univ. Lille, Inria, CNRS,
Centrale Lille, UMR 9189 – CRIStAL 
F-59000 Lille, France}
  \icmlaffiliation{isi}{ACMU, Indian Statistical Institute, Kolkata - 700108, India}

  \icmlcorrespondingauthor{Udvas Das}{udvas.das@inria.fr}

  \icmlkeywords{Policy Gradient methods, Performative Optimality, Reinforcement Learning}

  \vskip 0.3in
]


\printAffiliationsAndNotice{}  
\doparttoc
\faketableofcontents

\begin{abstract}
    Post-deployment machine learning algorithms often influence the environments that they act in, and thus, \textit{performatively shift} the underlying dynamics that the standard Reinforcement Learning (RL) ignores. While designing optimal algorithms in this \textit{performative} setting has been studied in supervised learning, the RL counterpart remains under-explored. 
    In this paper, we prove the performative counterparts of the performance difference lemma and the policy gradient theorem in RL, and introduce the \textbf{Performative Policy Gradient} algorithm (\framework). \framework~is the first policy gradient algorithm designed to account for performativity in RL. Under softmax parametrisation, and also with and without entropy regularisation, we prove that \framework~converges to \textit{performatively optimal policies}, i.e. policies that remain optimal under the distribution shifts induced by themselves. Thus, \framework~significantly extends the prior works in Performative RL that achieves  \textit{performative stability} but not optimality. Our empirical analysis on standard performative RL environments validate that \framework~outperforms the existing performative RL algorithms aiming for stability.\vspace*{-1em}
\end{abstract}
\section{Introduction}\label{sec:intro}
Reinforcement Learning (RL) studies the dynamic decision making problems under incomplete information~\citep{rlbook}. Since an RL algorithm tries and optimises a utility function over a sequence of interactions with an unknown environment, RL has emerged as a powerful tool for algorithmic decision making. 
Specially, in the last decade, RL has underpinned some of the celebrated successes of AI, such as championing Go with AlphaGo~\citep{silver2014deterministic}, \ifsinglecol\else\, \fi aligning Large Language Models (LLMs)~\citep{bai2022training}, reasoning~\citep{havrilla2024teaching} etc.
The classical paradigm of RL assumes the underlying environment to be \textit{static}, and the goal of RL algorithms is to find a utility-maximising, aka \textit{optimal policy}, for choosing actions over time. But \textit{the static environment assumption does not hold universally}.

\ifsinglecol
\else
\begin{figure}[t!]
    \centering
    \includegraphics[width=0.54\linewidth]{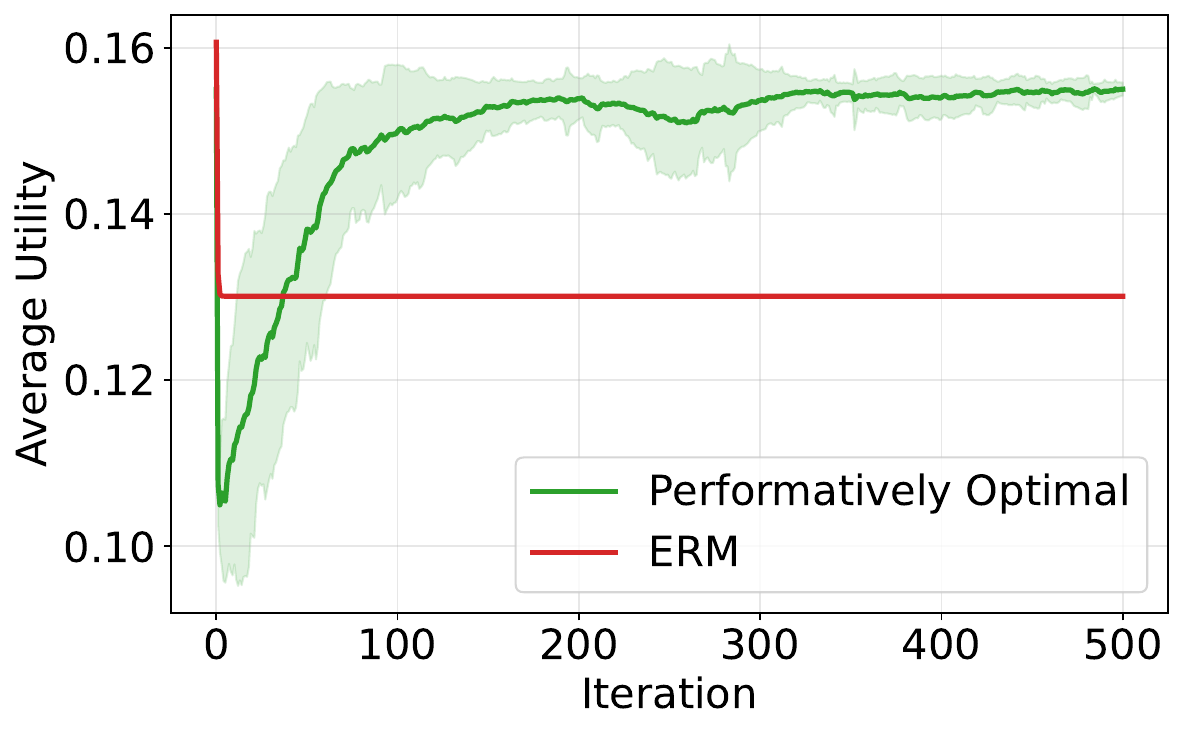}
    \caption{Average utility (over $20$ runs) obtained by ERM and Performatively Optimal policies for $\beta=0.5$.}\label{fig:loan}\vspace*{-2em}
\end{figure}
\fi

\textit{In this digital age, algorithms are not passive.} Their decisions also shape the environment that they interact with, inducing distribution shifts. 
This phenomenon in which predictive AI models trigger actions that influence their own outcomes is known as \textit{performativity}. 
In supervised learning, the study of \textit{performative prediction} is pioneered by~\citet{perdomo2020performative}, and followed by an extensive literature across optimisation, control, multi-agent RL, and games \citep{izzo2021learn,izzo2022learn,miller2021outside,li2022state,narang2023multiplayer,piliouras2023multi,gois2024performative}. 
There has been attempts to achieve performative optimality or stability for real-life tasks-- recommender systems~\citep{eilat2023performative}, measuring the power of firms~\citep{hardt2022performative,mofakhami2023performative}, healthcare~\citep{zhang2022shifting} etc. 
\textit{Performativity is also omnipresent in deployed RL systems.}
For example, an RL algorithm deployed in a recommender system does not only maximise the user satisfaction but also shifts the preferences of the users in long-term~\citep{chaney2018algorithmic,mansoury2020feedback}. 

\begin{example}[Performative RL in loan approval] 
A loan approval algorithm predicts whether an applicant should (or should not) obtain a loan according to their credit score $x$ that depends on the capitals of the applicant and the population. At each time $t$, a loan applicant arrives with a credit score $x_t \sim \mathcal{N}(\mu_t,\sigma^2)$. The bank chooses to give a loan by applying a softmax binary classifier $\bpi_{\theta}: \mathds{R} \rightarrow \{0,1\}$ on $x_t$ with threshold parameter $\theta$. This decision has two effects. (a) The bank receives a positive payoff $R$, if the loan applicant repays their loan, or else, loses by $L$. Thus, the bank’s expected utility for policy $\bpi_{\theta}$ is
$U(\theta, \mu) \triangleq \mathbb{E}_{x }\big[ \bpi_\btheta(x) (\prob(\text{repay}|x) R - (1-\prob(\text{repay}|x)) L) \big]\,.$
(b) Since the capitals of both the applicant and the population influence the credit score, we model the change in population mean $\mu_{t+1}$ by the bank’s policy, i.e. approval rate $\mathbb{E}_{x }\big[ \pi_\theta(x) \big]$. Specifically, 
$\mu_{t+1} \triangleq (1 - \beta)\mu_t + \beta  f\big(\mathbb{E}_{x_t}\big[ \pi_\theta(x_t) \big]\big)$,
where $\beta \in [0,1]$ is the performative strength and
$f: \mathds{R} \rightarrow [-M, M]$. 
Now, if one ignores the performative nature of this problem, and tries to find the optimal with respect to a static credit distribution, it obtains $\theta^{\rm{ERM}} \triangleq \argmax_\theta U\big(\theta, \mu_0\big).$ In contrast, if it considers performativity, it obtains $\theta^{\rm{Perf}} \triangleq \argmax_\theta U\big(\theta, \mu^*(\theta)\big)$. Figure~\ref{fig:loan} shows that the average utility obtained by $\theta^{\rm{ERM}}$ and $\theta^{\rm{Perf}}$ are different-- demonstrating performativity as a common phenomenon across decision making problems, and its effect on the desired optimal solution (Appendix~\ref{app:example}).
\end{example}\vspace*{-0.5em}

These practical problems have motivated the study of \textit{performative RL}. Though \citet{bell2021reinforcement} were the first to propose a setting where the transition and reward of an underlying MDP non-deterministically depend on the deployed policy, \cite{mandal2023performativereinforcementlearning} formally introduced \textit{Performative RL}, and its solution concepts, i.e., performatively stable and optimal policies. \textit{Performatively stable policies} do not change due to distribution shifts after deployment. \textit{Performatively optimal policies} yield the highest expected return once deployed in a performative RL environment. \citet{mandal2023performativereinforcementlearning} proposed direct optimization and ascent based techniques to attain performative stability upon repeated retraining. Extending this, \cite{rank2024performative,mandal2024performative} solved the same problem with delayed retraining for gradually shifting and linear MDPs. However, \textit{there exists no algorithm yet in performative RL that provably converges to the performative optimal policy}.  

In classical RL, Policy Gradient (PG) algorithms treat policy as a parametric function and update the parameters through gradient ascent algorithms~\citep{Williams1992,Sutton1999,kakade2001natural}. PGs are efficient and scalable. TRPO~\citep{Schulman2015TRPO}, PPO~\citep{Schulman2017PPO}, NPG~\citep{kakade2001natural}, DDPG~\citep{silver2014deterministic} are some of the PG algorithms widely used across modern RL. Recent theoretical advances also establish finite-sample convergence guarantees~\citep{PGtheory,yuan2022general} for different PG algorithms.
Motivated by the simplicity and efficiency of PG algorithms, we ask two questions. 
\begin{tcolorbox}[top=2pt,bottom=2pt,left=2pt,right=2pt]
    1. \textit{How to design PG-type algorithms for performative RL environments to achieve optimality?}\\
    2. \textit{What are the minimal conditions for PG-type algorithms to converge to the performatively optimal policy}?
\end{tcolorbox}
\textbf{Our contributions} address both the questions affirmatively, and showcase the difference of optimality-seeking and stability-seeking algorithms in performative RL.

\textbf{I. Algorithm Design:} \textit{We propose the first Performative Policy Gradient algorithm}, \framework, both with and without entropy-regularisation for performative RL environments (Section~\ref{sec:algo}). We derive the performative policy gradient theorem showing that the gradient of performative value function involves the classical policy gradient term and two novel gradient terms for environment shifts-- (a) the expected gradient of reward, and (b) the expected gradient of log-transition probabilities times its impact on the expected cumulative return. We deploy this result to estimate the performative policy gradient for \textit{any differentiable parametrisations of the policy and the environment}.

\textbf{II. Convergence to Performative Optimality.} We show that for any Performative Markov Decision Process (PeMDP) with smooth transition functions and rewards, \framework{} converges close to the optimal policy (Section~\ref{sec:analysis}). 
We provide a novel and generic recipe to prove convergence of \framework~via (a) smoothness of the performative value function, and (b) a performative gradient domination lemma capturing the per-step improvements due to performative policy gradients. 
As a concrete example, we show that, for PeMDPs with softmax policies, linear rewards and exponential family transitions, \framework~converges to an $\left(\epsilon +\frac{1}{1-\gamma} \right)$-ball around a performative optimal policy in $\bigO\left(\frac{|\cS||\cA|^2}{\epsilon^2(1-\gamma)^3}\right)$ iterations, where $|\cS|$ and $|\cA|$ are the number of states and actions, respectively, and $\gamma$ is discount factor.


\textbf{III. Stability- vs. Optimality-seeking Performative RL.} We further theoretically and numerically contrast the performances of stability-seeking and optimality-seeking algorithms. Theoretically, we derive the performative performance difference lemma that distinguishes the effect of policy update on these two objectives. Experimentally, we compare the performances of \framework~with the state-of-the-art stability-seeking algorithms, MDRR (Mixed Delayed Repeated Retraining,~\citet{rank2024performative}) and RPO-FS~\citep{mandal2023performativereinforcementlearning}. The results validate that \framework~yields significantly higher average return than the baselines. 
\vspace*{-0.5em}\begin{tcolorbox}[top=2pt,bottom=2pt,left=1pt,right=1pt]
    \cite{mandal2023performativereinforcementlearning} poses the question of developing performative policy gradient algorithms as an open problem. \textit{We \emph{affirmatively} solve an extension of this open problem to compute the performatively optimal policies for PeMDPs with discrete state-actions.} 
\end{tcolorbox}
\vspace*{-.5em}\section{Preliminaries: From RL to Performative RL}\vspace*{-.2em}
Now, we formalise the RL and performative RL problems, and provide the basics of policy gradient algorithms in RL.

\vspace{-.5em}\subsection{ Infinite-horizon Discounted MDP}\vspace*{-.5em}

In RL, we commonly study Markov Decision Processes (MDPs) defined as the tuple $(\cS,\cA,\transition,\reward,\gamma)$. Here, $\cS \subseteq \reals^d$ is the state space and $\cA \subseteq \reals^d$ is the action space. Both the spaces are assumed to be compact. At any time $t\in\mathbb{N}$, an agent plays an action $a_t \in \cA$ at a state $s_t\in \cS$. It transits the MDP environment to a state $s_{t+1}$ according to a transition function $\transition(\cdot\mid s_t,a_t) \in \Delta(\cS)$. The agent further receives a reward $\reward(s_t,a_t) \in \reals$ quantifying the goodness of taking the action $a_t$ at $s_t$. The strategy to take an action is represented by a stochastic map, called \textit{policy}, i.e., $\bpi : \cS \rightarrow \Delta(\cA)$.
Given an initial state distribution $\brho \in \Delta(\cS)$, \textit{the goal is to find the optimal policy $\bpi^{\star}$ that maximises} the expected discounted sum of rewards, i.e., the \textit{value function}:
$
V^{\bpi}(\brho) \defn \expectation_{s_0 \sim \brho, s_{t+1} \sim \transition(\cdot \mid s_t, \bpi(s_t))}\left[ \sum_{t=0}^{\infty} \gamma^t \reward(s_t, \bpi(s_t))\right]\,,
$ where $\gamma \in (0,1)$ is called the \textit{discount factor}. $\gamma$ indicates how much a previous reward matters in the next step, and bounds the effective horizon of a policy to $(1-\gamma)^{-1}$. 

\ifsinglecol
\else
\setlength{\textfloatsep}{6pt}
\begin{algorithm}[t!]
\caption{Vanilla Policy Gradient}\label{alg:vanilla_pg}
\begin{algorithmic}[1]
\STATE \textbf{Input:} Learning rate $\eta>0$.
\STATE \textbf{Initialize:} Policy parameter $\btheta_0(s,a) \forall s\in\cS,a\in\cA$.
\FOR{$t=1$ to T}
\STATE Estimate the gradient ${\nabla_{\btheta} V^\bpi(\brho)}\mid_{\btheta=\btheta_t}$
\STATE \textbf{Gradient ascent}:
    $\btheta_{t+1} \leftarrow \btheta_t+\eta {\nabla_{\btheta} V^\bpi(\brho)}\mid_{\btheta=\btheta_t} $
\ENDFOR 
\end{algorithmic}
\end{algorithm}
\fi
\textbf{Policy Gradient (PG) Algorithms.} PG algorithms maximise the value function by updating the policy with a gradient of value function~\citep{Williams1992}. To compute the gradient, we choose a parametric family of policies $\bpi_{\btheta}$ for some $\btheta \in \reals^d$ (e.g. direct~\citep{PGtheory,wang2022policy}, softmax~\citep{PGtheory,mei2020global}, Gaussian~\citep{ciosek2020expected}). 
Specifically, vanilla PG (Algorithm~\ref{alg:vanilla_pg}) performs a gradient ascent on the policy parameter at each step $t$. As the goal is to maximise $V^{\bpi}(\brho)$, we update $\btheta$ towards $\nabla_{\btheta}V^{\bpi}(\brho)$, i.e., the direction improving the value $V^{\bpi}(\brho)$ with a fixed learning rate $\eta>0$. For vanilla PG, the policy gradient takes the convenient form leading to estimators computable only with policy rollouts.\vspace*{-.2em}
\begin{theorem}[Policy Gradient Theorem~\citep{Sutton1999}]
    Given a differentiable parametrisation $\btheta \mapsto \bpi_\btheta$, the Q-value and advantage functions are $Q^{\bpi_\btheta}(s, a) \triangleq \expectation_{s_{t+1} \sim \transition(\cdot \mid s_t, \bpi(s_t))}\left[ \sum_{t=0}^{\infty} \gamma^t \reward(s_t, \bpi(s_t))\mid s_0=s,a_0=a\right]$ and $A^{\bpi_\btheta}(s, a) \triangleq Q^{\bpi_\btheta}(s, a) - V^{\bpi_\btheta}(s)$. Then, the gradient of value function is $(1-\gamma)\nabla_{\btheta}V^{\bpi_\btheta}(\brho)= \expectation\nolimits_{\tau \sim \prob^{\bpi_\btheta}}\left[\sum_{t=0}^\infty { \gamma^tA^{\pi_\btheta}(s, a)\nabla_{\btheta}\log \bpi_\btheta(a \mid s)} \right]$.
\vspace*{-.8em}
\end{theorem}
Since the value function is not concave in the policy parameters, achieving optimality with PG becomes a challenge. But scalability and efficiency of PG motivated a rich line of research deriving minimum conditions and parametric forms of policies for convergence to optimal policy~\citep{PGtheory,mei2020global,wang2022policy,yuan2022general}. \textit{Our work extends these algorithmic techniques and theoretical insights to performative RL.}

\vspace*{-.6em}\subsection{Infinite-horizon Discounted Performative MDP}\vspace*{-.2em}

Given a set of policies $\Pi$, the Performative Markov Decision Process (PeMDP) is defined as the set of MDPs $\cM(\Pi) \triangleq \{ \cM(\bpi) \mid \bpi \in \Pi\}$, where each MDP is a tuple $ \cM(\bpi) \triangleq (\cS, \cA, {\color{blue}\transition_{\bpi}}, {\color{blue}\reward_{\bpi}}, \gamma)$. Note that the transition function and rewards are now functions of the deployed policy $\bpi \in \Delta(\cA)$~\citep{mandal2023performativereinforcementlearning}. 
In this setting, the probability of generating a trajectory \( \tau \defn (s_t, a_t)_{t=0}^{\infty} \) with a policy \( \bpi \) and underlying MDP $\cM(\bpi')$ is given by\footnote{Hereafter, for relevant quantities, $\bpi$ in superscript denotes the deployed policy, and $\bpi'$ in subscript denotes the transition and reward functions shifting policy that the algorithm interacts with.}
$
   \prob_{\bpi'}^{\bpi}(\tau\mid \brho) \defn \brho(s_0) \prod_{t=0}^{\infty} \bpi(a_t\mid s_t) {\color{blue}\transition_{\bpi'}}(s_{t+1} \mid s_t, a_t)\,.
$ 
The state-action occupancy measure for the deployed policy $\bpi$ and the environment-inducing policy $\bpi'$ is $\occupancy{\bpi', \brho}{\bpi} \triangleq {(1-\gamma)}~\mathbb{E}_{\tau \sim \prob_{\bpi'}^{\bpi}} \left[ \sum_{t=0}^{\infty} \gamma^t \mathds{1}(s_t =s, a_t=a) \mid s_0 \sim \brho \right]$. 
Now, we define the total expected return in PeMDPs, i.e., the \textit{performative value function}, that we aim to maximise.

\begin{definition}[Performative Value Function]\label{def:performative_V}   
Given a policy \( \bpi \in \Pi \) and an initial state distribution \( \brho \in \Delta(\cS) \), the performative value function \( V_{\bpi}^{\bpi}(\brho) \) is\vspace*{-.5em}
\begin{align}\label{eq:performative_V}
V_{\bpi}^{\bpi}(\brho) \triangleq \mathbb{E}_{\tau \sim \prob_{\bpi}^{\bpi}} \left[ \sum_{t=0}^{\infty} \gamma^t {\color{blue}\reward_{\bpi}}(s_t, \bpi(s_t)) \mid s_0 \sim \brho \right]\,.
\end{align}\vspace*{-1em}
\end{definition}

Equation~\eqref{eq:performative_V} captures the performativity aspect in PeMDPs as the dynamics changes with a deployed policy $\bpi(\cdot\mid s)$.  On a similar note, we define the performative Q-value function (or action-value function) of a policy $\bpi$.

\begin{definition}[Performative Q-value]\label{def:performative_Q}
   Given a policy \( \bpi \in \Pi \) and an initial state-action pair $(s,a) \in (\cS,\cA)$, the performative Q-value function \( Q_{\bpi}^{\bpi}(s,a) \) is\vspace*{-.5em} 
   \begin{align}\label{eq:performative_V}
       Q^{\bpi}_{\bpi}(s,a) \defn \expectation_{\tau \sim \prob^{\bpi}_{\bpi}} \left[\sum_{t=0}^{\infty} \gamma^t \reward_\bpi(s_t,a_t)\Big| s_0 =s,a_0=a\right]
   \end{align}
\end{definition}

Performative Q-value satisfies $Q^{\bpi}_{\bpi}(s,a) = \reward_{\bpi}(s,a) + \gamma \expectation_{s' \sim \transition_\bpi(\cdot|s,a)} \left[ V^{\bpi}_{\bpi}(s')\right]$. 
We can maximise performative value function in two ways: (i) considering $\bpi$ as both the environment-inducing policy and the policy of the RL agent, or (ii) agent plays another policy $\bpi'$, while fixing $\bpi$ as the environment-inducing policy. At this point, we introduce the notions of optimal and stable policies in PeMDPs.
\ifsinglecol
\else
\begin{definition}[Performative Optimality]\label{def:perf_opt}
   A policy \( \opt_o \) is performatively optimal if it maximizes the performative value function, i.e., $\bpi^{\star}_o \in \argmax_{\bpi \in \Delta(\cA)} V_{\bpi}^{\bpi}(\brho)$.
\end{definition}
\fi
This implies that if we play the policy $\bpi$ in the environment induced by policy $\bpi$ to maximise the expected return, we land on the performatively optimal policy.
\ifsinglecol
\else 
\begin{definition}[Performative Stability]\label{def:perf_stable}
    A policy $\opt_s$ is performatively stable if there is no gain in performative value function due to deploying any other policy than $\opt_s$ in the environment induced by $\opt_s$ i.e., $\bpi^{\star}_s \in \argmax_{\bpi \in \Delta(\cA)} V_{\opt_s}^{\bpi}(\brho)$.
\end{definition}
\fi
A performatively optimal policy may not be stable, i.e., $\opt_{o}$ may not be optimal for an environment $\cM(\opt_o)$, when it is deployed~\citep{mandal2023performativereinforcementlearning}. In general, the performative value function of $\opt_{o}$ might be equal to or higher than that of $\opt_{s}$. In this work, \textit{we design PG algorithms computing a performatively optimal policy for a given PeMDP}, and reinstate their differences with performatively stable policies.

The existing literature on PeMDPs~\citep{mandal2023performativereinforcementlearning,mandal2024performative,rank2024performative,chen2024practical,pollatos2025corruption} focused primarily on finding a performatively stable policy (Definition~\ref{def:perf_stable}). In practice, while the stable policies matter for certain applications, they might show very sub-optimal performance, which are not desired in many real-life tasks. \textit{We bridge this gap by proposing the first provably converging and computationally-efficient PG algorithm for PeMDPs.} We also empirically show the deficiency of the existing stability-seeking algorithms if we aim for optimality (Section~\ref{sec:exp}).

\textbf{Entropy Regularised PeMDPs.} Entropy regularisation has emerged as a simple but powerful technique in classical RL to design smooth and efficient RL algorithms with sufficient exploration. Thus, we study a variant of the performative value function that is regularised using discounted entropy~\citep{neu2017unified,liu2019policy,zhao2019maximum}. In this approach, the original value function in Definition~\ref{def:performative_V} is regularised with the discounted entropy $H_{\bpi}(\brho) \defn \expectation_{\tau \sim \prob_{\bpi}^{\bpi}} \left[-\sum_{t=0}^{\infty} \gamma^t \log\bpi(a_t\mid s_t)\right]$. This is equivalent to maximising the expected reward with a shifted reward function $\softreward_{\bpi}(\bpi(s_t),s_t) \triangleq \reward_{\bpi}(\bpi(s_t),s_t) - \lambda\log(\bpi(a_t\mid s_t))$ for some $\lambda\geq 0$. $\softreward_{\bpi}$ is referred as the ``soft-reward'' in literature~\citep{wang2024reward,herman2016inverse,shi2019soft}. Now, we define the \textit{soft performative value function}.

\ifsinglecol
\else
\begin{definition}[Entropy Regularised (or \textit{Soft}) Performative Value Function]\label{def:soft_perf_value}
Given a policy \( \bpi \in \Pi\), a starting state distribution \( \brho \in \Delta(S) \), and a regularisation parameter $\lambda\geq0$, the soft performative value function \vspace*{-.5em}
\begin{align}\label{eq:performative_V_regularised} 
    &\softV_{\bpi}^{\bpi}(\brho) \defn \mathbb{E}_{\substack{\tau \sim \prob_{\bpi}^{\bpi}\\s_0 \sim \brho }}\left[ \sum_{t=0}^{\infty} \gamma^t \softreward_{\bpi}(s_t, \bpi(s_t)) \right]
\end{align}\vspace*{-1.8em}
\end{definition}
\fi
Since policies belong to the probability simplex, the entropy regularisation naturally lends to smoother and stable PG algorithms. Later, we show that the discounted entropy is a smooth function of the policy parameters for PeMDPs extending the optimization-wise benefits of entropy regularisation to PeMDPs. 
Additionally, using the notion of soft rewards, we can similarly define soft performatively optimal and stable policies for entropy regularised PeMDPs. Here, \textit{we unifiedly design PG algorithms for both the unregularised and the entropy regularised PeMDPs}.
\vspace*{-.5em}\section{Policy Gradient in Performative RL}
We first study the impact of policy updates in PeMDPs. Then, we leverage it to derive the performative policy gradient theorem, and design Performative PG (\framework) algorithm for any differentiable parametric policy class.

\vspace*{-.5em}\subsection{Impact of Policy Updates on PeMDPs}\label{sec:analysis}
In RL, performance difference lemma quantifies the impact of changing policies on the value functions~\citep{kakade2002approximately}. It has been central to analysing and developing PG algorithms~\citep{PGtheory,silver2014deterministic,kallel2024augmented}. 
Here, we derive the performative version of the performance difference lemma quantifying the shift in the performative value function due to change in the deployed and environment-inducing policies.

\ifsinglecol
\else
\begin{tcolorbox}[top=2pt,bottom=2pt,left=1pt,right=1pt]
\begin{lemma}[Performative Performance Difference Lemma]\label{lemma:perf_perf_diff}
  The difference in performative value functions induced by $\bpi$ and $\bpi' \in \Pi$ while starting from the initial state distribution $\brho$ is $V^{\bpi}_{\bpi}(\brho) - V^{\bpi'}_{\bpi'}(\brho)~=$
  \begin{align}
   \frac{1}{1-\gamma}\Big(&\expectation_{(s,a)\sim \occupancy{\bpi',\brho}{\bpi}}[A_{\bpi'}^{\bpi'}(s,a) + {\color{blue}(\reward_{\bpi}(s,a) - \reward_{\bpi'}(s,a))}\notag \\
    & {\color{blue} + \gamma (\transition_{\bpi}(\cdot|s,a) - \transition_{\bpi'}(\cdot|s,a))^\top V^{\bpi}_{\bpi}(\cdot)} \Big]\Big)
  \end{align}
  where $A_{\bpi'}^{\bpi'}(s,a) \defn  Q_{\bpi'}^{\bpi'}(s,a) - V_{\bpi'}^{\bpi'}(s)$ is the performative advantage function for any $s\in\cS$ and $a\in\cA $. 
\end{lemma}
\end{tcolorbox}
\fi
The crux is to decompose the performative value among environment-inducing and deployed policies. Specifically, we observe that the suboptimality gap
\begin{align*}
     \subopt(\bpi') &\triangleq V_{\opt}^{\opt}(\brho)- V_{\bpi'}^{\bpi'}(\brho)\\
     &= \underbrace{V_{\opt}^{\opt}(\brho) - V_{\bpi'}^{\opt}(\brho)}_{\text{performative shift term}}+  \underbrace{V_{\bpi'}^{\opt}(\brho)-V_{\bpi'}^{\bpi'}(\brho)}_{\text{performance difference term}}\vspace*{-2em}
\end{align*}

(1) \textit{Connection to Classical RL.} In classical RL, the performance difference lemma yields $V^{\bpi}(\brho) - V^{\bpi'}(\brho)  = \frac{1}{1-\gamma} \expectation_{(s,a)\sim \occupancy{\rho}{\bpi}}[A^{\bpi'}(s,a)]$. The first term in Lemma~\ref{lemma:perf_perf_diff} is equivalent to the classical result in the environment induced by $\bpi'$. But due to environment shift, two more terms appear in the performative performance difference incorporating the impacts of reward shifts and transition shifts.

(2) \textit{Connection to Performative Stability.} The performance difference term, i.e., $V_{\bpi'}^{\bpi}(\brho)-V_{\bpi'}^{\bpi'}(\brho)$, quantifies the impact of changing the deployed policy from $\bpi'$ to $\bpi$ in an environment induced by $\bpi'$. Thus, a stability-seeking algorithm tries to minimise this term, while an optimality-seeking algorithm incorporates both the terms.
    
Hence, we ask: 
\begin{tcolorbox}[left=1pt,right=1pt,top=2pt,bottom=2pt]
    \textit{How much does the performative shift term influence the performative performance difference?}
\end{tcolorbox}

Hereafter, we focus on PeMDPs with bounded rewards, which is a common assumption in RL~\citep{agarwal2019reinforcement}. 
\begin{assumption}[Bounded Rewards]\label{ass:bounded_r}
    We assume that the rewards of PeMDPs are bounded in $[-R_{\max}, R_{\max}]$. 
\end{assumption}\vspace{-.5em}


\begin{algorithm}[t!]
\caption{\framework : \textbf{Pe}rformative \textbf{P}olicy \textbf{G}radient}\label{alg:pepg}
\begin{algorithmic}[1]
\STATE \textbf{Input:} Reward bound $R_{\max}>0$, discount factor $\gamma \in(0,1)$, initial state distribution $\brho$, learning rate $\eta>0$ 
\STATE \textbf{Initialize:} Initial policy parameters $\theta_0$, initial value function parameters $\phi_0$.
\FOR{$k = 1, 2, \dots$}
    \STATE \textbf{Collect trajectories:} $\mathcal{D}_k = \{\tau_i\}_{i=1}^I$, where each  $\tau_i \defn \{(s_{i,t},a_{i,t},s_{i,t+1},r_{i,t})\}_{t=0}^{T-1}$ of length $T$ by playing $\bpi_{\btheta_k}$
    
    \STATE Compute returns ${R}_{k} \defn \{{R}_{k,i} \}_{i=1}^I$, where $R_{k,i} = \{R_{k,i,t}\}_{t=0}^{T-1}$
    \STATE Compute advantage estimates $\hat{A}_{k}(\tau_i)$ using value function $\hat V_{\phi_k}(\tau_i)$ for each $\tau_i \in \cD_k$, i.e., estimate of $V^{\bpi_{\btheta_k}}_{\bpi_{\btheta_k}}(\tau_i)$ obtained from fitted value network with parameters $\phi_k$
    \STATE \textbf{Gradient estimation:} Estimate gradient using \eqref{eq:gradient_estimate_for_pseudocode}
    \STATE \textbf{Ascent step:} Use parameter update Equation \eqref{eqn:ascent_step} 
    \STATE Fit value function $V_{\phi_{k+1}}$
    \ifsinglecol
\else
    $$\hspace*{-1em}\phi_{k+1} \leftarrow \argmin_{\phi} \frac{1}{I T}\sum_{i =1}^I \sum_{t=0}^{T-1} \Big(\hat V_{\phi_k}(s_t \in \tau_i) - {R}_{k,i,t}\Big)^2$$
    \fi
\ENDFOR
\end{algorithmic}
\end{algorithm}

Now, we bound the effect of performative shift for gradually shifting PeMDPs with Lipschitz transitions and rewards with respect to the deployed policies~\citep{rank2024performative}.

\begin{tcolorbox}[top=1pt,bottom=1pt,left=1pt,right=1pt]
\begin{lemma}[Bounding the Performative Performance Difference for Gradually Shifting Environments]\label{lemma:perf_grad_upperbound}
  Let us assume that the rewards and transitions are Lipschitz functions of policy, i.e., $\norm{\reward_{\bpi} - \reward_{\bpi'}}_1 \leq \lipreward \norm{\bpi-\bpi'}_\infty~~\text{and}~~\norm{\transition_{\bpi} - \transition_{\bpi'}}_1 \leq \liptransition \norm{\bpi-\bpi'}_\infty\,$, for $\lipreward,\liptransition \geq 0$. Under Assumption~\ref{ass:bounded_r}, the performative shift in the sub-optimality gap of a policy $\bpi_\btheta$ satisfies 
  \begin{align*}
    &\Big| \subopt(\bpi_{\btheta}) -\frac{1}{1-\gamma}\expectation_{(s,a)\sim d_{\bpi_\btheta, \brho}^{\opt_o}}[A_{\bpi_\btheta}^{\bpi_\btheta}(s,a)] \Big| \leq\notag\\ 
    &\frac{2\sqrt{2}}{1-\gamma} \Big(\lipreward + \frac{\gamma\liptransition R_{\max}}{1-\gamma} \Big) \expectation_{s_0\sim \brho}\hellinger{\opt_o(\cdot|s_0)}{\bpi_\btheta(\cdot|s_0)}.
  \end{align*}
  $\hellinger{\cdot}{\cdot}$ is the Hellinger distance between distributions.
\end{lemma}
\end{tcolorbox}
\textbf{Implications.} Lemma~\ref{lemma:perf_grad_upperbound} characterises the \textit{extra cost} to adapt to performativity of the environment in terms of Hellinger distance between the true performatively optimal policy $\opt_o$ and any other policy $\bpi_\btheta$. This implies that the gap between the optimal performative value function and that of any stability-seeking algorithm is $\bigO\left((1-\gamma)^{-1}\right)$. This gap is significantly less than the sub-optimality gap achieved by the existing algorithms. For example, repeated policy optimisation shows sub-optimality gap $\bigO\left(\max \lbrace \frac{|\cS|^{5/3} |\cA|^{1/3} \epsilon^{2/3}}{ (1-\gamma)^{14/3}}, \frac{\epsilon |\cS|}{(1-\gamma)^4} \rbrace \right)$~\citep{mandal2023performativereinforcementlearning}. 
Thus, we seek to improve on the existing works and design algorithms achieving sub-optimality gap $\bigO\left((1-\gamma)^{-1}\right)$. 

We note that an optimality-seeking algorithm tries to minimise both the advantage function and the performative shifts in the environment quantified by $\hellinger{\opt_o(\cdot|s_0)}{\bpi_\btheta(\cdot|s_0)}$. A stability-seeking algorithm tries to minimise the advantage function, and thus, RHS of Lemma~\ref{lemma:perf_grad_upperbound} cannot go lower than $\bigO((1-\gamma)^{-2})$. Thus, the optimality-seeking algorithms can yield a lower sub-optimality gap than the stability-seeking algorithms if they also incorporate the performative shifts.

\subsection{Algorithm: Performative Policy Gradient (\framework)}\label{sec:algo}
For optimality-seeking algorithms, the goal is to maximise the performative value function. Gradient ascent is a standard first-order optimisation method to find maxima of a function. The ascent step of performative policy gradient is
\ifsinglecol
\else 
\begin{align}\label{eqn:ascent_step}
\hspace*{-1em}\btheta_{t+1} \gets 
\begin{cases}
    \btheta_t+\eta_t \nabla_{\btheta}V^{\bpi_\btheta}_{\bpi_\btheta}(\tau)\mid_{\btheta = \btheta_t} \text{ (unregularised)}\\
    \btheta_t+\eta_t \nabla_{\btheta}\softV^{\bpi_\btheta}_{\bpi_\btheta}(\tau)\mid_{\btheta = \btheta_t} \text{ (entropy reg.)}
\end{cases}
\end{align}
\fi

Given this ascent step, we evaluate the gradient at each time from the roll-outs of the present policy. In classical PG, the policy gradient theorem supports this computation~\citep{Williams1992,Sutton1999,silver2014deterministic}. Now, we derive the \textit{performative policy gradient theorem}.
\ifsinglecol
\else
\begin{tcolorbox}[top=2pt,bottom=2pt,left=1pt,right=1pt]
\begin{theorem}[Performative Policy Gradient Theorem]\label{thm:perf_pg_theorem}
(a) For the unregularised case, the gradient of the performative value function with respect to $\btheta$ is $\nabla_\btheta V_{\bpi_\btheta}^{\bpi_\btheta}(\tau) =$\vspace*{-.5em}
    \begin{align} 
         &\expectation_{\tau \sim \prob_{\bpi_\btheta}^{\bpi_\btheta}} \Big[ \sum_{t=0}^{\infty} \gamma^t A_{\bpi_\btheta}^{\bpi_\btheta}(s_t, a_t) \Big(\nabla_\btheta \log \bpi_\btheta(a_t \mid s_t) \notag\\&+   \nabla_\btheta \log P_{\bpi_\btheta}(s_{t+1}|s_t,a_t)\Big) + \gamma^t \nabla_\btheta r_{\bpi_\btheta}(s_t, a_t)\Big].\label{eq:perf_grad_unregularised}\vspace*{-1em}
    \end{align}
(b) For the entropy-regularised case, we define the soft advantage, soft Q-value, and soft value functions with respect to the soft rewards $\tilde r_{\bpi_\btheta}$ satisfying $\tilde A^{\bpi_\btheta}_{\bpi_\btheta}(s,a) = \tilde Q^{\bpi_\btheta}_{\bpi_\btheta}(s,a) - \tilde V^{\bpi_\btheta}_{\bpi_\btheta}(s)$ that further yields $\nabla_\btheta \tilde V_{\bpi_\btheta}^{\bpi_\btheta}(\tau) =$\vspace*{-.5em}
\begin{align}
    \hspace{-4em}&\expectation_{\tau \sim \prob_{\bpi_\btheta}^{\bpi_\btheta}} \Big[
    \sum_{t=0}^{\infty} \gamma^t \tilde A_{\bpi_\btheta}^{\bpi_\btheta}(s_t, a_t) \Big( \nabla_\btheta \log \bpi_\btheta(a_t \mid s_t) \notag\\
    &+  \nabla_\btheta \log P_{\bpi_\btheta}(s_{t+1}|s_t,a_t)\Big) + \gamma^t \nabla_\btheta  \tilde{r}_{\bpi_\btheta}(s_t, a_t) \Big].\label{eq:perf_grad_regularised}
\end{align}
\end{theorem}
\end{tcolorbox}
\fi

\textbf{\framework:} 
With the appropriate parameter choices, and initialisation of the policy parameter $\btheta$ and value function parameter $\phi$, for each episode $k$, \framework~first collects $I$ trajectories to calculate the return ${R}^i$ and estimates the advantage function $\hat{A}_k$ (Line 4-6). 
For a particular trajectory $\tau_i$, the estimated advantage for a given state-action is $   \widehat{A^{\bpi_{\btheta_k}}_{\bpi_{\btheta_k}}}(s_t^i,a_t^i) =  R_{t,k}^i -  V_{\phi_k}(s_t^i)$,
where $R^i= \sum_{t=0}^{T-1} \gamma^t \reward_{\bpi_{\btheta_k}}(s_t^i,a_t^i)$.

\textbf{Gradient Estimation (Line 7).} Using all the $I$ trajectories, \framework~computes a gradient estimate as $\widehat{\nabla_{{\btheta_k}}} V^{\bpi_{\btheta_k}}_{\bpi_{\btheta_k}}(\tau) =$ 
\ifsinglecol
\else
\begin{align}\label{eq:gradient_estimate_for_pseudocode}
    \hspace{-1em}&\frac{1}{I}\sum_{i=1}^{I}
    \sum_{t=0}^{T} \gamma^t \Big(\widehat{A_{\bpi_{\btheta_k}}^{\bpi_{\btheta_k}}}(s_t^i, a_t^i) \Big(\nabla_{\btheta_k} \log \bpi_{\btheta_k}(a_t^i \mid s_t^i)+\notag\\
    &  \nabla_{\btheta_k} \log P_{\bpi_{\btheta_k}}(s_{t+1}^i|s_t^i,a_t^i)\Big)+ \nabla_{\btheta_k} r_{\bpi_{\btheta_k}}(s_t^i, a_t^i|\btheta_k)\Big).
\end{align}
\fi
Further, in Line 8, \framework~updates the policy parameter for the next episode using a gradient ascent step leveraging the estimated average gradient over all $I$ trajectories. Specifically, we plug in $\widehat{\nabla_{{\btheta_k}}} V^{\bpi_{\btheta_k}}_{\bpi_{\btheta_k}}$ to both the unregularised and entropy-regularised update rules as given in Equation \eqref{eqn:ascent_step}. 
\section{Convergence Analysis of \framework}\label{sec:softmax}
In this section, we first derive the minimal condition on the PeMDPs and derive the convergnce analysis of \framework{} for them. Then, we specialise the analysis for exponential family PeMDPs. Hereafter, we use softmax parametrisation of policies~\citep{mei2020global,agarwal2019reinforcement,PGtheory}, defined as   $\bpi_\btheta(a|s) = \frac{e^{\btheta_{s,a}}}{\sum_{a'} e^{\btheta_{s,a'}} }$ for all $a\in \cA$ and $s\in\cS$.  

\textbf{Generic Convergence Analysis.} To maximise a given objective, any first-order ascent method uses gradients to compute the direction for improvement. Following that, smoothness of the objective plays a critical role to stitch the ascent steps over iterations (Equation~\eqref{eqn:ascent_step}), and thus, influences the convergence rate, step size selection, and overall efficiency.

Specifically, for PeMDPs the value function depends on reward and transition functions. Thus, we start the convergence analysis of \framework{} with the necessary smoothness assumption on reward and transitions, and their gradients. 
\begin{assumption}[Smooth/Bounded Sensitivity PeMDPs]\label{ass:generic}
The transition and reward functions of the PeMDP are (a) Lipschitz functions of policy, i.e. $\norm{\reward_{\bpi} - \reward_{\bpi'}}_1 \leq \lipreward \norm{\bpi-\bpi'}_{\infty}$ and $\norm{\transition_{\bpi} - \transition_{\bpi'}}_1 \leq \liptransition \norm{\bpi-\bpi'}_{\infty}\,,$ with $\lipreward,\liptransition \geq 0$, and
(b) smooth functions of policy, i.e. $\norm{\nabla_{\bpi}\reward_{\bpi} - \nabla_{\bpi'}\reward_{\bpi'}}_1 \leq R_2 \norm{\bpi-\bpi'}_\infty$ and $\norm{\nabla_{\bpi} \transition_{\bpi} - \nabla_{\bpi'} \transition_{\bpi'}}_1 \leq T_2 \norm{\bpi-\bpi'}_\infty$ with $R_2, T_2 \geq 0$.\vspace*{-0.5em}
\end{assumption}
Assumption \ref{ass:generic}-(a) is essential to ensure bounded gradients for a first-order optimiser. Assumption \ref{ass:generic}-(b) is equivalent to the bounded sensitivity condition of the environmental dynamics in control theory~\citep{he2025decision} and smoothness in optimisation~\citep{nesterov2018smooth,mahdavi2013mixed}. 

\textbf{Challenges and Three Step Analysis.} 
The main challenge to prove convergence of \framework{} arises due to non-concavity of the performative value function in the policy parametrisation $\btheta$. A similar issue occurs while proving convergence of PG-type algorithms in classical RL, which has been overcome by leveraging smoothness properties of the value functions and by deriving the local Polyak-\L ojasiewicz (PL)-type conditions, known as \textit{gradient domination}~\citep{agarwal2019reinforcement,yuan2022general}.  
Extending these insights, we devise a three step convergence analysis for \framework{}.

\textbf{Step 1: Performative Gradient Domination.} First, we connect the performative performance difference (Lemma~\ref{lemma:perf_grad_upperbound}) with the norm of the performative policy gradient (Theorem~\ref{thm:perf_pg_theorem}). This allows us to connect the per-iteration improvement in the performative value function in \framework{} with the performative gradient ascent at that step.  
\begin{tcolorbox}[top=2pt,bottom=2pt,left=2pt,right=2pt]
   \begin{lemma}[Performative Gradient Domination Lemma] \label{lemma:grad_domm_generic}
         Let us define $\Cov \triangleq \max_{\btheta, \bnu} \Big\|\frac{\occupancy{\bpi_\btheta,\brho}{\opt_o}}{\occupancy{\bpi_\btheta,\bnu}{\bpi_\btheta}}\Big\|_{\infty}$. Then, under Assumption \ref{ass:bounded_r} and \ref{ass:generic}-(a), we get 
        (a) $\subopt(\bpi_{\btheta}) \leq {\sqrt{|\cS||\cA|}} \Cov \|\nabla_\btheta V^{\bpi_\btheta}_{\bpi_\btheta}(\bnu)\|_2 + \frac{1+\Cov}{(1-\gamma)^2}\Big(\lipreward + \liptransition R_{\max}\Big)$ for unregularised value function, and (b) $\subopt(\bpi_{\btheta}\mid\blambda)\leq {\sqrt{|\cS||\cA|}} \Cov \|\nabla_\btheta\softV^{\bpi_\btheta}_{\bpi_\btheta}(\bnu)\|_2 + \frac{2+\Cov}{(1-\gamma)^2}\Big(\lipreward + \liptransition (R_{\max}+\lambda \log |\cA|)\Big) $ for entropy-regularisation.
\end{lemma} 
\end{tcolorbox}

\textbf{Step 2: Smoothness of Performative Value Functions.} 
Now, to properly stitch the gradient ascent steps over iterations, we prove that the unregularised performative value function is $\bigO(\frac{|\cA|}{(1-\gamma)^2})$-smooth (Appendix \ref{app:smoothness}, Lemma~\ref{lemm:perf_smooth}). Further, we show that the entropy regularised performative value function is also $\bigO(\frac{|\cA|}{(1-\gamma)^2})$-smooth as entropy is a $\bigO\left( \frac{\log |\cA|}{(1-\gamma)^3}\right)$-smooth function for PeMDPs (Lemma~\ref{lemma:regularizer_smoothness}). 

\textbf{Step 3: Iterative Application of Gradient Domination for Smooth Objectives.} Finally, we apply gradient domination (Lemma~\ref{lemma:grad_domm_generic}) along with the iterative convergence proof of gradient ascent for smooth functions. The intuition is that since the per-step sub-optimality is dominated by the gradient and the smooth functions are bounded by quadratic envelopes of parameters, applying gradient ascent iteratively would bring the sub-optimality down to small error levels after enough iterations. We formally prove this in Theorem~\ref{thm:conv}.
\begin{table*}[t!]
\resizebox{\textwidth}{!}{
\begin{tabular}{c c c c}
\toprule
 Algorithms & Regulariser $\lambda$ & Min. \#samples & Environment\\ 
\midrule
RPO-FS~\citep{mandal2023performativereinforcementlearning} & $ \bigO \left( \frac{|\cS|+\gamma|\cS|^{5/2}}{(1-\omega)(1-\gamma)^4} \right)$  & $\bigO\left(\frac{|\cA|^2|\cS|^3}{\epsilon^4(1-\gamma)^6 \lambda^2} \ln \left(\# \mathrm{iter}\right)\right)$ &Direct PeMDPs $+$ quadratic-regul. on occupancy\\ 
& & & $\omega$-dependence between two envs.\\
MDRR~\citep{rank2024performative} & $\bigO \left( \frac{|\cS|+\gamma|\cS|^{5/2}}{(1-\omega)(1-\gamma)^4} \right)$ & $\bigO\left(\frac{|\cA|^2|\cS|^3}{\epsilon^4(1-\gamma)^6 \lambda^2} \ln \left(\#\mathrm{iter}\right)\right)$ & Direct PeMDPs $+$ quadratic-regul. on occupancy\\
& & & $\omega$-dependence between two envs.\\
\framework~(This work, Theorem~\ref{thm:conv_softmax_b}) & $\frac{R_{\max}(1-\gamma)}{1+2\log(|\cA|)}
$ & $\bigO\left(\frac{|\cS||\cA|^2}{\epsilon^2(1-\gamma)^3}\right)$ & Exponential family PeMDPs $+$ entropy regul. on policy \\ 
\framework~(This work, Theorem~\ref{thm:conv_softmax_a}) & $0
$ & $\bigO\left(\frac{|\cS||\cA|}{\epsilon^2} \max\Big\lbrace\frac{\gamma R_{\max}\mid\cA\mid}{(1-\gamma)^3}, \frac{\gamma^2}{(1-\gamma)^4} \Big\rbrace\right)$ & Exponential family PeMDPs $+$ no regularisation \\
\bottomrule
\end{tabular}}\vspace*{.5em}
\caption{Comparison of theoretical performance of SOTA stability-seeking algorithms and \framework.}\label{tab:comparison}\vspace*{-2.1em}
\end{table*}

\begin{tcolorbox}[top=2pt,bottom=2pt,left=1pt,right=1pt]
    \begin{theorem}[Convergence of \framework{} for Smooth PeMDPs]\label{thm:conv}
We set learning rate $\eta = \bigO\left(\frac{(1-\gamma)^2}{|\cA|}\right)$. Then,
(a) For unregularised objective , we get
$\min_{t < T} \subopt(\bpi_{\btheta_t}) \leq \epsilon + \bigO\left(\frac{\Cov}{(1-\gamma)^2 }\right)$  when $T = \Omega\left( \frac{|\cS||\cA|^2{\rm Cov}^2}{\epsilon^2 (1-\gamma)^3}\right)$. 
(b) For entropy-regularised objective with $\lambda = \frac{(1-\gamma)R_{\max}}{1+2\log|\cA|}$, we get
$
    \min_{t < T} \subopt(\bpi_{\btheta_t}|\lambda) \;\; \leq \;\; \epsilon + \bigO\left(\frac{\Cov}{(1-\gamma)^2}\right)$   when  $T = \Omega\left( \frac{|\cS||\cA|^2{\rm Cov}^2}{\epsilon^2 (1-\gamma)^3}\right)$.
\end{theorem}
\end{tcolorbox}
\textbf{Implications.}  
\textit{1. Efficiency Gain.} \framework{} converges to an $\epsilon$-optimal policy in $\bigO\left(\frac{|\cS||\cA|^2}{\epsilon^2(1-\gamma)^3}\right)$ iterations. \textit{This reduces the sample complexity by at least} $\bigO\left(\frac{|\cS|^2}{\epsilon^2 (1-\gamma)^3}\right)$ in comparison to the existing stability-seeking algorithms that directly optimise the occupancy measures~\citep{mandal2023performativereinforcementlearning,mandal2024performative,rank2024performative}. 

\textit{2. Value of $\lambda$.} Regularisation parameters for the existing algorithms must be bigger than $\bigO\left(\frac{|\cS|}{(1-\gamma)^4}\right)$. This is counter-intuitive in practice. We show that setting $\lambda = \bigO\left( \frac{1}{\log|\cA|}\right)$ suffices for proving convergence to $\epsilon$-optimality. 

3. \textit{Coverage Condition.} Minimum number of samples required to achieve convergence is proportional to $\Cov^2$ for smooth PeMDPs. This is a ubiquitous quantity dictating convergence of PG-methods in classical RL~\citep{PGtheory,mei2020global}, and retraining methods in performative RL~\citep{mandal2023performativereinforcementlearning,rank2024performative} as we have to pay for the data coverage of the optimal policy with respect to that of the policy played at any iteration.

4. \textit{Bias \& Suboptimality Gap.} The $\bigO\left(\frac{\Cov}{(1-\gamma)^2}\right)$ bias appearing in Theorem \ref{thm:conv} is analogous to the effect of using relaxed weak gradient domination result \citep[Corollary 3.7]{yuan2022general}. It argues that if the policy gradient in classical MDPs satisfies  $\epsilon' + \|\nabla_\theta V(\theta)\| \ge 2\sqrt{\mu}\,\subopt(\bpi_\btheta)$  
for some $\mu>0$ and $\epsilon' > 0$, then the corresponding PG algorithm guarantees  
$\min_{t < T} \subopt(\bpi_{\btheta_t}) \le \bigO(\epsilon+\epsilon') $ for big enough $T$. Lemma~\ref{lemma:grad_domm_generic} constructs the performative counterpart of this relaxed weak gradient domination with  $\epsilon' = \bigO\left(\frac{\Cov}{(1-\gamma)^2}\right)$. Existence of such irreducible bias at convergence for both stability and optimality seeking algorithms indicate that this might be inherent to performative RL~\citep{sahitaj2025independent,mandal2023performativereinforcementlearning}. We further observe that this effect vanishes for \framework{} if the Lipschitz constants $\lipreward$ and $\liptransition$ in Lemma~\ref{lemma:perf_grad_upperbound} decay to $0$, i.e., if the performative effect on environment stabilises with time. 


\textbf{Convergence of \framework{} for Exponential Family PeMDPs.} 
Though for our proposed convergence analysis, it is enough to assume smooth parametrisation of transitions and rewards without fixing a specific parametric family, we further specialise our analysis for \textit{exponential family PeMDPs} to (a) show exact gradient computations, and (b) establish an explicit relation between parametrisation of PeMDP and smoothness of performative value functions.

The exponential family PeMDPs have an exponential family distribution as transition functions with non-negative feature map $\psi(\cdot):\cS \rightarrow \reals$, and linear reward functions with respect to the policy parameters. 
Specifically, $\{ \cM(\btheta) = \cM(\bpi_{\btheta})\mid \btheta \in \reals^{|\cS|\times |\cA|}\}$ such that, $\forall s\in\cS$ and $a\in\cA$, $\reward_{\bpi_\btheta}(s,a) = \mathcal{P}_{[-R_{\max}, R_{\max}]}[\xi \btheta_{s,a}]$, and $\transition_{\bpi_\btheta} (s'|s,a) = \exp\left( \btheta_{s',a} \psi(s') - \log(\sum_{s''}e^{\btheta_{s,a} \psi(s'')})\right)$,
where $\psi(\cdot)\leq\psi_{\max}$, and $\xi \in [0,R_{\max}]$ to align with Assumption~\ref{ass:bounded_r}. 
Note that modelling the Markovian transition function as exponential families is common in bandits and RL theory~\citep{moulos2019,al2021navigating,ouhamma2023bilinear,karthik2024optimal}.
Instantiating the three step analysis of smooth PeMDPs, we prove convergence of \framework{} for exponential family PeMDPs. We tabulate the sample complexity of \framework{} for this parametrisation in Table~\ref{tab:comparison}. Corollary~\ref{cor:smoothness_constants_softmax}-\ref{cor:smoothness_constants_softmax_reg} and Theorem~\ref{thm:conv_softmax_a}-\ref{thm:conv_softmax_b} show that smoothness guaranties of smooth PeMDP are retained here, though we observe a gain in terms of per-step improvement (gradient domination), and thus, in the achievable sub-optimality gap.  

\textbf{Discussions.} \textit{1. Reduction of Bias in Performative Gradient Domination.} For exponential family PeMDPs with softmax policy class, we observe an improvement of $\bigO\left(\frac{\Cov}{(1-\gamma)}\right)$ in the per-step bias term compared to Lemma~\ref{lemma:grad_domm_generic}.

\vspace*{-1.5em}\begin{tcolorbox}[top=2pt,bottom=2pt,left=2pt,right=2pt]
\begin{lemma}[Performative Gradient Domination for Exponential Family PeMDPs with Softmax Policy Class] \label{lemma:grad_domm_softmax} 
\ifsinglecol
\else 
(a) For unregularised value function, $\subopt(\bpi_\btheta) \leq 
    \sqrt{|\cS||\cA|} \Cov \|\nabla_\btheta V^{\bpi_\btheta}_{\bpi_\btheta}(\bnu)\|_2  + \frac{R_{\max}}{1-\gamma} $.\\
(b) For entropy-regularisation, $\subopt(\bpi_\btheta|\lambda) \leq
    \sqrt{|\cS||\cA|} \Cov \|\nabla_\btheta V^{\bpi_\btheta}_{\bpi_\btheta}(\bnu)\|_2 +\frac{R_{\max}+\lambda\log|\cA|}{1-\gamma}$.
\fi
\end{lemma}
\end{tcolorbox}\vspace*{-1em}

\begin{figure*}[t!]
    \centering    
    \includegraphics[width=0.7\linewidth]{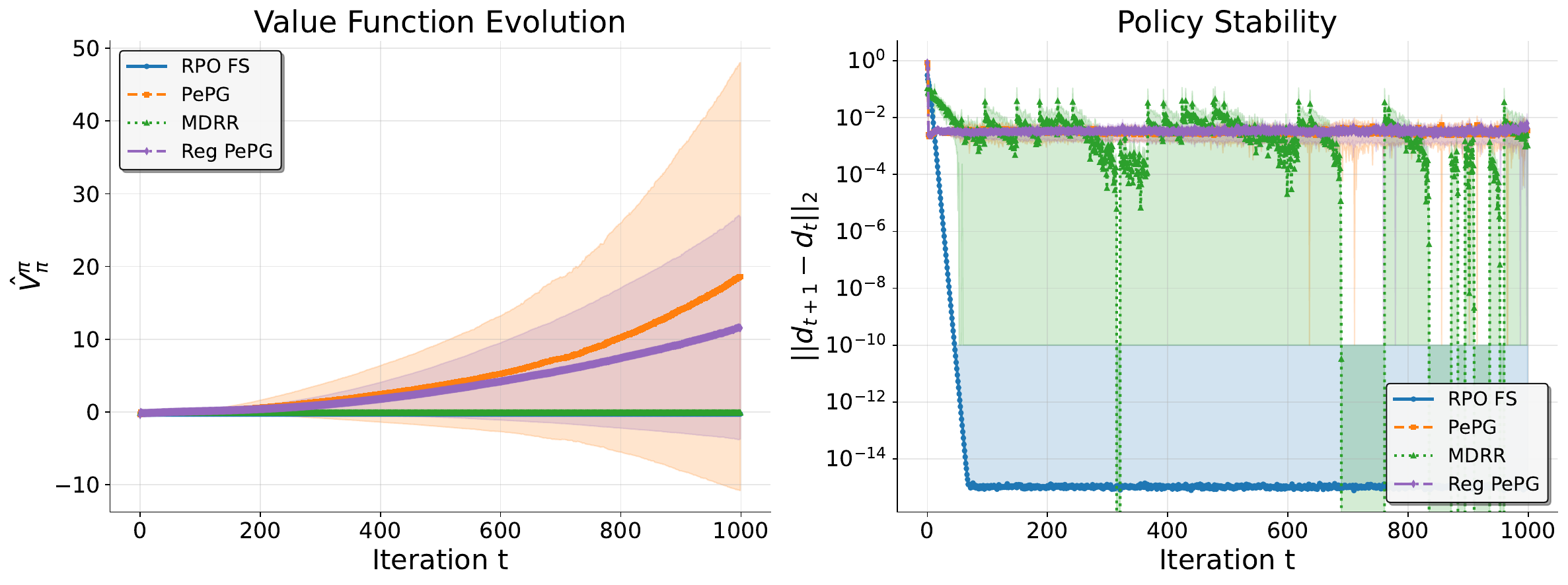}\vspace*{-.2em}
    \caption{Comparison of evolution in expected average return (both regularised and unregularised) and stability of \framework~with SOTA stability-achieving methods. Each algorithm is run for 20 random seeds and 1000 iterations.}\label{fig:overview}\vspace*{-1.2em}
\end{figure*}

\textit{2. Improvement over the SOTA in Performative RL.} Theorem~\ref{thm:conv}, \ref{thm:conv_softmax_a} and \ref{thm:conv_softmax_b} indicate that \framework{} reaches closer to optimality whereas a stability seeking algorithm can yield policies which are $\bigO\left(\Cov/(1-\gamma)^4 \right)$ far from the optimal value function~\citep[Theorem 7]{mandal2024performative}. We find the existing algorithms to achieve $\bigO\left(1/(1-\gamma)^6\right)$~\citep{mandal2023performativereinforcementlearning} and $\bigO\left(1/(1-\gamma)^8\right)$~\citep{sahitaj2025independent} sub-optimality in different settings of performative RL.\footnote{We omit coverage term as its expression varies with settings.} This emphasises on the lacunae of stability seeking algorithms and effectiveness of \framework{} to achieve optimality.    
\vspace*{-.5em}\section{Experimental Analysis}\label{sec:exp}\vspace*{-.2em}
In this section, we empirically compare the performance of \framework~and analyse its behaviour against the state-of-the-art stability-seeking algorithms.\footnote{Code of \framework~ is available in~\href{https://anonymous.4open.science/r/PePG-0C27/README.md}{Link}. Further ablation studies with respect to hyperparameters are in Appendix~\ref{app:ablation}.} 

\textbf{Baselines and Experimental Setup.} We evaluate \framework{} (with and without entropy regularisation) alongside MDRR~\citep{rank2024performative} and RPO-FS~\citep{mandal2023performativereinforcementlearning}, which represent the current state-of-the-art in performative RL under gradually shifting environments. MDRR has demonstrated significant improvements over traditional repeated retraining methods like RPO-FS, by leveraging historical data from multiple deployments. 
We refer to Appendix \ref{app:exp_details} for details on the setup and the environment.

\textbf{Results and Observations.}
Experimental evaluation across 1000 iterations reveals fundamental differences between \framework~, MDRR and RPO-FS in the performative setting. 

\textbf{I. Optimality.} Left panel of  (Figure~\ref{fig:overview}) reveals  a clear performance 
hierarchy among the three methods. \framework~achieves the highest value function performance, reaching approximately $18$ for standard \framework{} and $11$ for regularized \framework{} (Reg PePG), both showing consistent improvement from initial values around 0. This sustained upward progression over iterations highlights \framework's effectiveness in discovering better performative equilibria rather than settling for the first stable solution encountered like MDRR or RPO-FS. 


\textbf{II. Comparison of Optimality- and Stability-seeking Algorithms.} Right panel of Figure~\ref{fig:overview} exposes a critical limitation of algorithms designed primarily for stability rather than optimality. RPO-FS remains most stable with successive occupancy differences below $10^{-14}$ throughout training, achieving its design goal of rapid convergence. However, this stability comes at the cost of solution quality, as RPO-FS remains at the lowest performance level near 0.
MDRR shows intermediate behaviour with stability fluctuations around $10^{-2}$ to $10^{-4}$, punctuated by occasional sharp drops to near $10^{-14}$ (approximately every 200-300 iterations). Despite these periodic stabilization events, MDRR does not achieve notable performance improvement, demonstrating that stability-focused methods can get trapped at suboptimal equilibria. 
In contrast, both \framework{}~variants maintain consistent moderate variability 
around $10^{-2}$ throughout the $1000$ iterations, indicating persistent exploration without convergence. These results demonstrate that optimality-seeking algorithms benefit from maintaining policy variability rather than prematurely stabilizing, even though this comes with higher occupancy measure distances between iterations. Also, \framework~exhibits marginally better performance compared to its regularised variant albeit with larger variance across its runs. Note, we measure stability of \framework{} by $\|\occupancy{t+1}{}-\occupancy{t}{}\|_2$ to align with stability-notion of the baselines, as they optimise the occupancy measure per step.

\textbf{III. Computational Gain.} In Appendix~\ref{app:exp_details}, we further show that \framework{} incurs $\sim 1.25-2.5\times$ less computational time per iteration in comparison with the baselines. 


\vspace*{-.6em}\section{Discussions, Limitations, and Future Works}
We study the problem of Performative Reinforcement learning in tabular MDPs, where the agent's actions cause potential shift in underlying reward and transition dynamics. First, we derive the novel performative counterpart of classic Performance Difference Lemma and Policy Gradient Theorem that successfully capture the performative dynamics of the underlying environment. Next, we propose the first PG algorithm, \framework, which attains $\epsilon$-performative optimality unlike the existing stability-seeking algorithms, affirmatively solving an extended open problem in \citep{mandal2023performativereinforcementlearning}.

An interesting future direction is to scale \framework~for continuous state-action space while incorporating variance-reduction techniques~\citep{wu2018variance,papini2018stochastic}. Moreover, the gridworld environment is the only benchmark currently available for testing performative RL algorithms. In this direction, constructing a performative test-bed or simulator for both discrete and continuous state-action spaces, is an important future work.


\section*{Acknowledgements}
 The authors would like to acknowledge the Inria-ISI Kolkata associate team SeRAI for supporting the collaboration. DB and UD would also like to acknowledge ANR JCJC project REPUBLIC (ANR-22-CE23-0003-01) and PEPR project FOUNDRY (ANR23-PEIA-0003). BD is also partially supported by the project ANR JCJC project NeuRL (ANR-23-CE23-0006).
\bibliography{ref,iclr2026_conference}
\bibliographystyle{icml2026}

\newpage
\onecolumn
\appendix
\part{Appendix}
\parttoc
\clearpage
\section{Notations}
\renewcommand{\arraystretch}{1.5}
%
\begin{longtable}{ll}
\hline
\textbf{Notation} & \textbf{Description}  \\
\hline  
$\cS$ & state space \\
$\cA$ &  action space\\
$\gamma$ & discount factor\\
$\bpi_{\btheta}$ & policy parametrized by $\btheta$\\
$\Pi(\Theta)$ & policy space\\
$\transition_{\bpi}$ & transition under the environment induced by policy $\bpi$\\
$\reward_{\bpi}$ & reward under the environment induced by  policy $\bpi$\\
$\opt_s$ & performatively stable policy\\
$\opt_o$ & performatively optimal policy\\
$\transition_{\opt_o}$ & reward under the environment induced by performatively optimal policy\\
$\reward_{\opt_o}$ & reward under the environment induced by performatively optimal policy\\
$\occupancy{\opt_o}{\opt_o}$ & state-action occupancy of optimal policy\\
$V^{\opt_o}_{\opt_o}$ & value function of optimal policy\\
$\occupancy{\bpi_2}{\bpi_1}$ & state-action occupancy of playing policy $\bpi_2$ in the environment induced by policy $\bpi_1$\\
$V^{\bpi_2}_{\bpi_1}$ & value function for playing policy $\bpi_2$ in the environment induced by policy $\bpi_1$\\
$Q^{\bpi_2}_{\bpi_1}$ &  Q-value function for playing policy $\bpi_2$ in the environment induced by policy $\bpi_1$\\
$A^{\bpi_2}_{\bpi_1}$ & advantage function for playing policy $\bpi_2$ in the environment induced by policy $\bpi_1$\\
$\tilde Q^{\bpi_2}_{\bpi_1}$ &  entropy regularised Q-value function for playing policy $\bpi_2$ in the environment induced by policy $\bpi_1$\\
$\softA^{\bpi_2}_{\bpi_1}$ & entropy regularised advantage function for playing policy $\bpi_2$ in the environment induced by policy $\bpi_1$\\
$\simplex$& $K$-dimensional simplex\\
$\brho$ & Initial state distribution $\in \Delta_{\cS}$\\
$\Cov$ &  Coverage $\defn \max_{\btheta, \nu} \Bigg\|\frac{\occupancy{\bpi_\btheta,\brho}{\opt_o}}{\occupancy{\bpi_\btheta,\nu}{\bpi_\btheta}}\Bigg\|_{\infty}$\\
$\softV^{\bpi_2}_{\bpi_1}$ & entropy regularised value function for playing policy $\bpi_2$ in the environment induced by policy $\bpi_1$\\
\hline
\end{longtable}\newpage

\section{Details of the Toy Example: Loan Approvement Problem}\label{app:example} 
\textbf{Environment.} We consider a simple setup, where a population of loan applicants are represented by a scalar feature $x \in \mathbb{R}$, such that 
$x \sim \mathcal{N}(\mu, \sigma^2)$,
where $\mu$ is the mean and variance $\sigma^2 > 0$ is fixed and known.

\textbf{Bank's Policy.} The bank chooses a \emph{threshold policy} parametrised by $\theta \in \mathbb{R}$. A loan is granted to an applicant with feature $x$, if $x \geq \theta$.  
For a smoothed analysis, we use a differentiable policy class:
$\pi_\theta(x) = \sigma\big(k (x - \theta)\big)$, where $\sigma(z) = \frac{1}{1+e^{-z}}$ is the logistic sigmoid and $k>0$ controls smoothness.  

\begin{wrapfigure}{r}{8cm}
\centering
\includegraphics[width=0.45\textwidth]{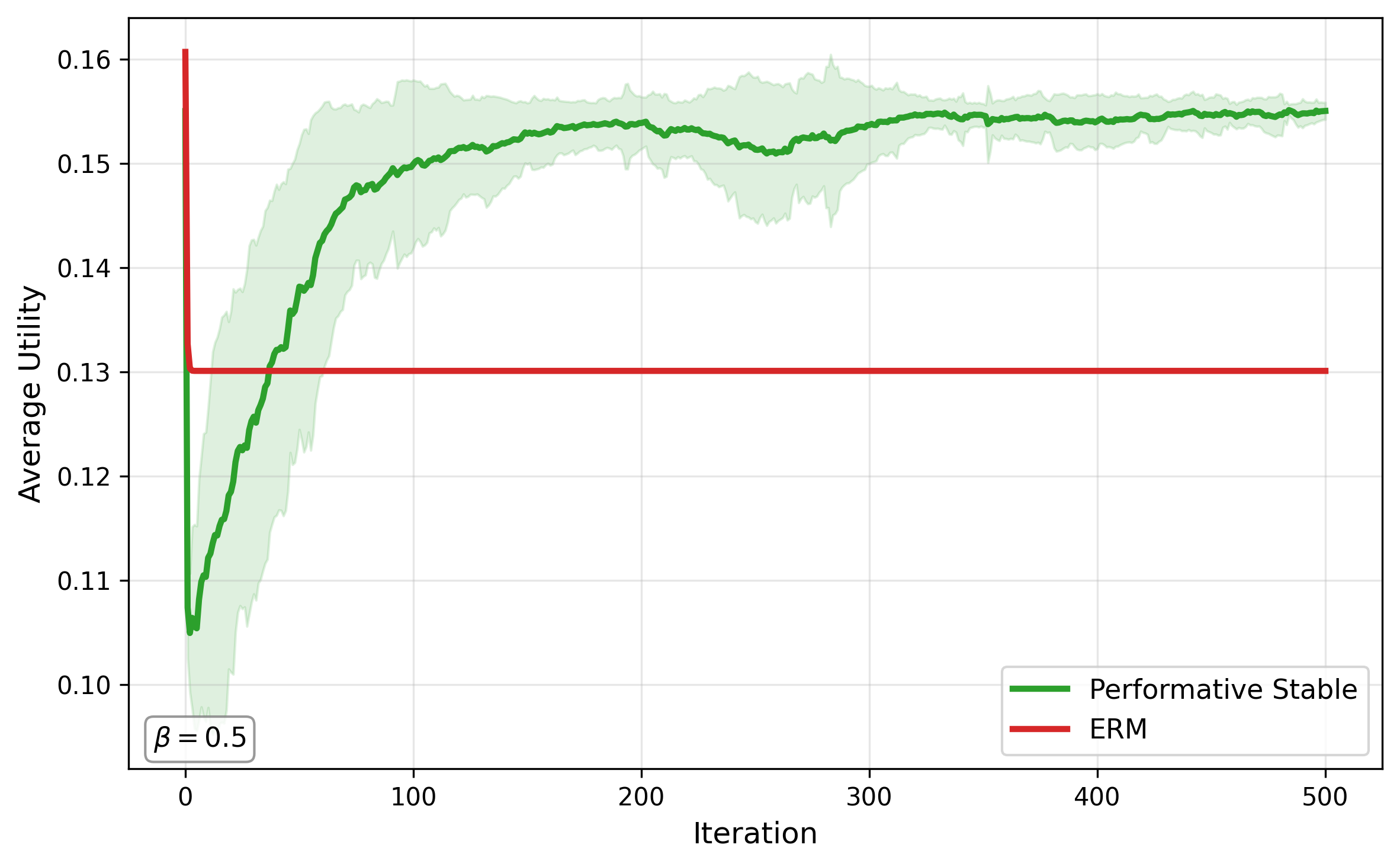}
\caption{Average reward (over 10 runs) obtained by ERM optimal policies across performative strength $\beta$.}
\end{wrapfigure}

\textbf{Rewards.} If a loan is granted to applicant $x$, the bank receives a random payoff:
$$
r(x) = \begin{cases}
+R & \text{if applicant repays}, \\\\
-L & \text{if applicant defaults},
\end{cases}
$$
with repayment probability
$\mathbb{P}(\text{repay} \mid x) = \sigma(\gamma x - c)$, where $\gamma>0$ controls sensitivity and $c$ is a calibration constant.  
Thus, the expected reward from granting to $x$ is
$$
u(x) = \sigma(\gamma x - c) \cdot R - \big(1 - \sigma(\gamma x - c)\big) \cdot L.
$$

\textbf{Expected Utility.} Given feature distribution $x \sim \mathcal{N}(\mu,\sigma^2)$, the bank’s expected utility for policy $\theta$ is
$$
U(\theta, \mu) = \mathbb{E}_{x \sim \mathcal{N}(\mu,\sigma^2)}\big[ \pi_\theta(x) \cdot u(x) \big].
$$

\textbf{Performative Feedback.} We define a grant rate:
$g(\theta, \mu) = \mathbb{E}_{x \sim \mathcal{N}(\mu,\sigma^2)}\big[ \pi_\theta(x) \big]$. We assume that dependence of $\mu$ on policy is tailored by a bounded performative update rule:
$\mu_{t+1} = (1 - \beta)\mu_t + \beta \cdot f\big(g(\theta, \mu_t)\big)$,
where $\beta \in [0,1]$ is called the \textit{Performative Strength} and
$f(g) \in [-M, M]$
projects grant rate to a bounded domain.  

Hence, at equilibrium, the induced feature distribution satisfies the fixed point condition:
\[
\mu^*(\theta) = (1 - \beta)\mu^*(\theta) + \beta f\big(g(\theta, \mu^*(\theta))\big).
\]

\subsection{Optimisation Problems}
\textbf{ERM Optimum.}  
Ignoring performative effects (i.e. assuming $\mu = \mu_0$ is fixed), the ERM-optimal policy solves
$
\theta^{\text{ERM}} = \argmax_\theta U(\theta, \mu_0).
$

\textbf{Performatively Optimal Optimisation.}  
Accounting for performative dynamics of the underlying environment, the performatively optimal policy solves
$
\theta^{\text{Perf}} = \argmax_\theta U\big(\theta, \mu^*(\theta)\big).
$

In Algorithm~\ref{alg:rl_protocol}, we present a simple  agent-environment interaction/learning protocol under performative feedbacks. 
\begin{algorithm}[h!]
\caption{Learning Protocol of the Toy Example with Reinforcement Learning}\label{alg:rl_protocol}
\begin{algorithmic}[1]
\STATE \textbf{Input:} Performative strength $\beta$, $R,L,c,\gamma, M, k$, variance $\sigma^2 > 0$, .
\STATE \textbf{Initialize:} Policy parameter $\theta$ and initial value of mean $\mu_0$.
\FOR{$t=1$ to T}
\STATE Sample $x_t \sim \mathcal{N}(\mu_t, \sigma^2)$.
\STATE Grant loan with probability $\pi_{\theta_t}(x_t)$.
\STATE Observe reward $r_t$.
\STATE Update $\theta_{t}$ using Policy Gradient (REINFORCE~\citep{Sutton1999,agarwal2019reinforcement}).
\STATE Update population mean via performative dynamics:
    $\mu_{t+1} = (1 - \beta)\mu_t + \beta f\big(g(\theta_t, \mu_t)\big)$.
\ENDFOR 
\end{algorithmic}
\end{algorithm}
\section{Extended Related Works}\label{sec:related_works}
\textbf{Performative Prediction.} The study of performative prediction started with the pioneering work of~\citep{perdomo2020performative}, where they leveraged repeated retraining with the aim to converge towards a performatively stable point. We see extension of this work trying to achieve performative optimality~\citep{izzo2021learn,izzo2022learn,miller2021outside}. This further opened a plethora of works in various other domains such as Multi-agent systems~\citep{narang2023multiplayer,li2022multi,piliouras2023multi}, control systems~\citep{cai2024performative,barakat2025multi}, stochastic optimisation~\citep{li2022state,mendler2020stochastic}, games~\citep{wang2023network,gois2024performative} etc. There has been several attempt towards achieving performative optimality or stability for real-life tasks like recommendation~\citep{eilat2023performative}, to measure the power of firms~\citep{hardt2022performative,mofakhami2023performative}, in healthcare~\citep{zhang2022shifting} etc. Another interesting setting is the \textit{stateful} performative prediction i.e. prediction under gradual shifts in the distribution~\citep{brown2022performative,izzo2022learn,ray2022decision}, that paved the way for incorporating performative prediction in 
Reinforcement Learning.

\textbf{Performative Reinforcement Learning.} \cite{bell2021reinforcement} were the first to propose a setting where the transition and reward of an underlying
MDP depend non-deterministically on the deployed policy, thus capturing the essence of performativity to some extent. However, \cite{mandal2023performativereinforcementlearning} can be considered the pioneer in introducing the notion of ``\textit{Performative Reinforcement Learning}" and its solution concepts, performatively stable and optimal policy. They propose direct optimization and ascent based techniques which manage to attain performative stability upon repeated retraining. Extensions to this work, \cite{rank2024performative} and \cite{mandal2024performative} manage to solve the same problem with delayed retraining for linear MDPs. However, there exists no literature that proposes a performative RL algorithm that converges to the performative optimal policy. 


Specifically, \cite{mandal2023performativereinforcementlearning} frames the question of using policy gradient to find stable policies as an open problem. The authors further contemplate, as PG objective functions are non-concave in the policy space, whether it is possible to converge towards a stable policy. Thus, in this paper, we affirmatively solve an extension of this open problem for tabular PeMDPs.    

\textbf{Policy Gradient Algorithms.} Policy gradient algorithms build a central paradigm in reinforcement learning, directly optimizing parametrised policies by estimating the gradient of expected return. The foundational policy gradient theorem \citep{Sutton1999} established an expression for this gradient in terms of the score and action-value function, while \citet{Williams1992} introduced the REINFORCE algorithm, providing an unbiased likelihood-ratio estimator. Convergence properties of stochastic gradient ascent in policy space were analysed in these early works. Subsequently, \citet{Konda2000} formalized actor–critic methods via two-timescale stochastic approximation, and \citet{Kakade2002} proposed the natural policy gradient, leveraging the Fisher information geometry to accelerate learning. Extensions to trust region methods \citep{Schulman2015TRPO}, proximal policy optimization \citep{Schulman2017PPO}, and entropy-regularized objectives \citep{Mnih2016A3C} have made policy gradient methods widely practical in high-dimensional settings. Recent theoretical advances provide finite-sample convergence guarantees and complexity analyses \citep{PGtheory,yuan2022general}, as well as robustness to distributional shift and adversarial perturbations \citep{zhang2020global,xu2020improved}. Collectively, this body of work establishes policy gradient methods as both practically effective and theoretically grounded method for solving MDP.

 \newpage
\section{Impact of Policy Updates on PeMDPs (Section~\ref{sec:analysis})}
\begin{replemma}{lemma:perf_perf_diff}[Peformative Performance Difference Lemma]
  The difference in performative value functions induced by $\bpi$ and $\bpi' \in \Pi$ while starting from the initial state distribution $\brho$ is
  \begin{align}
  (1)\quad  V^{\bpi}_{\bpi}(\brho) - V^{\bpi'}_{\bpi'}(\brho)
    = &~\frac{1}{1-\gamma} \expectation_{(s,a)\sim \occupancy{\bpi',\brho}{\bpi}}[A_{\bpi'}^{\bpi'}(s,a)]\notag \\ &~+ \frac{1}{1-\gamma} \expectation_{(s,a)\sim \occupancy{\bpi',\brho}{\bpi}}\Big[(\reward_{\bpi}(s,a) - \reward_{\bpi'}(s,a)) + \gamma (\transition_{\bpi}(\cdot|s,a) - \transition_{\bpi'}(\cdot|s,a))^\top V^{\bpi}_{\bpi}(\cdot) \Big] \,.
  \end{align}
  where $A_{\bpi'}^{\bpi'}(s,a) \defn  Q_{\bpi'}^{\bpi'}(s,a) - V_{\bpi'}^{\bpi'}(s)$ is the performative advantage function for any state $s\in\cS$ and action $a\in\cA $. 
  \begin{align}
  (2)\quad  V^{\bpi}_{\bpi}(\brho) - V^{\bpi'}_{\bpi'}(\brho)
    = &~\frac{1}{1-\gamma} \expectation_{(s,a)\sim \occupancy{\bpi',\brho}{\bpi}}\Big[A_{\bpi'}^{\bpi'}(s,a)\Big]\notag \\
    &+ \frac{1}{1-\gamma} \expectation_{(s,a)\sim \occupancy{\bpi,\brho}{\bpi}}\Big[(\reward_{\bpi}(s,a) - \reward_{\bpi'}(s,a))+ \gamma (\transition_{\bpi}(\cdot|s,a) - \transition_{\bpi'}(\cdot|s,a))^\top V^{\bpi}_{\bpi'}(\cdot) \Big] \,.
  \end{align}
  where $A_{\bpi'}^{\bpi'}(s,a) \defn  Q_{\bpi'}^{\bpi'}(s,a) - V_{\bpi'}^{\bpi'}(s)$ is the performative advantage function for any state $s\in\cS$ and action $a\in\cA $. 
  \begin{align}
  (3)\quad  V^{\bpi}_{\bpi}(\brho) - V^{\bpi'}_{\bpi'}(\brho)
    = &~\frac{1}{1-\gamma} \expectation_{(s,a)\sim \occupancy{\bpi,\brho}{\bpi}}\Big[A_{\bpi}^{\bpi'}(s,a)\Big]\notag\\
    &+ \frac{1}{1-\gamma} \expectation_{(s,a)\sim \occupancy{\bpi',\brho}{\bpi'}}\Big[(\reward_{\bpi}(s,a) - \reward_{\bpi'}(s,a))+ \gamma (\transition_{\bpi}(\cdot|s,a) - \transition_{\bpi'}(\cdot|s,a))^\top V^{\bpi'}_{\bpi}(\cdot) \Big] \,.
  \end{align}
  where $A_{\bpi}^{\bpi'}(s,a) \defn  Q_{\bpi}^{\bpi'}(s,a) - V_{\bpi}^{\bpi'}(s)$ is the performative advantage function for any state $s\in\cS$ and action $a\in\cA $. 
\end{replemma}
We only use the first version of this lemma in the main draft, and also hereafter, for the proofs.
\begin{proof}[Proof of Lemma~\ref{lemma:perf_perf_diff}]\, We do this proof in two steps. First step involves a decomposition of the difference in value function into two terms :  (i) difference in value function after deploying the same policy while agent plays two different policies i.e. the difference that explains stability of the deployed policy, and (ii) difference in value function for deploying two different policies i.e. performance difference for changing the deployed policy. While the second term can be bounded using classic performance difference lemma, in the next and final step, we control the stability inducing term (i). \\

\noindent \textit{Part(1) --} \textbf{Step 1: Decomposition.} We start by decomposing the performative performance difference to get a stability and a performance difference terms separately.

 \begin{align}
     V_{\bpi}^{\bpi}(\brho)- V_{\bpi'}^{\bpi'}(\brho) =& \underbrace{V_{\bpi}^{\bpi}(\brho) - V_{\bpi'}^{\bpi}(\brho)}_{\text{performative shift term}}+  \underbrace{V_{\bpi'}^{\bpi}(\brho)-V_{\bpi'}^{\bpi'}(\brho)}_{\text{performance difference term}}\notag\\
     =& V_{\bpi}^{\bpi}(\brho) - V_{\bpi'}^{\bpi}(\brho) + \frac{1}{1-\gamma} \expectation_{(s,a)\sim \occupancy{\bpi',\brho}{\bpi}}[A_{\bpi'}^{\bpi'}(s,a)]\label{eq:decomposition}
\end{align}
The last equality is a consequence of the classical performance difference lemma~\citep{Kakade2002ApproximatelyOA}.\\
\noindent \textbf{Step 2: Controlling the performative shift term.} First, let us define $\transition_{\bpi}^{\bpi} (s',s) \triangleq \sum_{a\in\cA} \transition_{\bpi} (s'|s,a) \bpi(a|s)$, and $\langle \transition_{\bpi}^\bpi(\cdot,s_0), V^{\bpi}_{\bpi}(\cdot)\rangle \triangleq \sum_{s\in\cS} V^{\bpi}_{\bpi}(s) \transition_{\bpi}^\bpi(s,s_0)$, for some $s_0 \sim \brho$.

We first observe that
\begin{align*}
V^{\bpi}_{\bpi}(s_0)  - V^{\bpi}_{\bpi'}(s_0) 
&= \expectation_{a \sim \bpi(\cdot|s_0)}
\Big[ \reward_{\bpi}(s_0,a) - \reward_{\bpi'}(s_0,a) \Big]
+ \gamma \expectation_{s \sim \transition_{\bpi}^{\bpi} (\cdot,s_0)} [V^{\bpi}_{\bpi}(s)] -  \gamma \expectation_{s \sim \transition_{\bpi'}^{\bpi} (\cdot,s_0)} [V^{\bpi}_{\bpi'}(s)]\\
&= \expectation_{a \sim \bpi(\cdot|s_0)}
\Big[ \reward_{\bpi}(s_0,a) - \reward_{\bpi'}(s_0,a) \Big]\\
&\qquad\qquad+ \gamma\sum_{s} \left(\transition_{\bpi}^{\bpi} (s,s_0) - \transition_{\bpi'}^{\bpi} (s,s_0)\right)V^{\bpi}_{\bpi}(s) + \gamma\sum_{s} \transition_{\bpi'}^{\bpi} (s,s_0)\left(V^{\bpi}_{\bpi}(s)  -  V^{\bpi}_{\bpi'}(s)\right)\\
&= \frac{1}{1-\gamma}\expectation_{(s,a) \sim \occupancy{\bpi',s_0}{\bpi}}
\Big[ \reward_{\bpi}(s,a) - \reward_{\bpi'}(s,a) + \gamma (\transition_{\bpi}(\cdot|s,a) - \transition_{\bpi'}(\cdot|s,a))^\top V^{\bpi}_{\bpi}(\cdot) \Big] 
\end{align*}
The last equality is obtained by recurring the preceding step iteratively.

We combine \textbf{Steps 1} and \textbf{2} and take expectation over $s_0 \sim  \brho$, we get
\begin{align*}
    V^{\bpi}_{\bpi}(\brho) - V^{\bpi'}_{\bpi'}(\brho)
    = \frac{1}{1-\gamma} \expectation_{(s,a)\sim \occupancy{\bpi',\brho}{\bpi}}\Big[A_{\bpi'}^{\bpi'}(s,a) + (\reward_{\bpi}(s,a) - \reward_{\bpi'}(s,a)) + \gamma (\transition_{\bpi}(\cdot|s,a) - \transition_{\bpi'}(\cdot|s,a))^\top V^{\bpi}_{\bpi}(\cdot) \Big] \,.
\end{align*}
\textit{Part(2) --} The second equality is obtained by changing the Step 2 as follows:
\begin{align*}
    V^{\bpi}_{\bpi}(\brho)  - V^{\bpi}_{\bpi'}(\brho) 
&= \expectation_{a \sim \bpi(\cdot|\brho)}
\Big[ \reward_{\bpi}(s_0,a) - \reward_{\bpi'}(s_0,a) \Big]
+ \gamma \expectation_{s \sim \transition_{\bpi}^{\bpi} (\cdot,s_0)} [V^{\bpi}_{\bpi}(s)] -  \gamma \expectation_{s \sim \transition_{\bpi'}^{\bpi} (\cdot,s_0)} [V^{\bpi}_{\bpi'}(s)]\\
&= \expectation_{a \sim \bpi(\cdot|\brho)}
\Big[ \reward_{\bpi}(s_0,a) - \reward_{\bpi'}(s_0,a) \Big]\\
&\qquad\qquad+ \gamma\sum_{s} \left(\transition_{\bpi}^{\bpi} (s,s_0) - \transition_{\bpi'}^{\bpi} (s,s_0)\right)V^{\bpi}_{\bpi'}(s) + \gamma\sum_{s} \transition_{\bpi}^{\bpi} (s,s_0)\left(V^{\bpi}_{\bpi}(s)  -  V^{\bpi}_{\bpi'}(s)\right)\\
\implies  V_{\bpi}^{\bpi}(\brho)- V_{\bpi'}^{\bpi'}(\brho)&= ~\frac{1}{1-\gamma} \expectation_{(s,a)\sim \occupancy{\bpi',\brho}{\bpi}}\Big[A_{\bpi'}^{\bpi'}(s,a)\Big]\\
&\qquad\qquad+ \frac{1}{1-\gamma} \expectation_{(s,a)\sim \occupancy{\bpi,\brho}{\bpi}}\Big[(\reward_{\bpi}(s,a) - \reward_{\bpi'}(s,a)) 
+ \gamma (\transition_{\bpi}(\cdot|s,a) - \transition_{\bpi'}(\cdot|s,a))^\top V^{\bpi}_{\bpi'}(\cdot) \Big] \,.
\end{align*}
The last equality is obtained by recurring the preceding step iteratively.

\textit{Part(3) --} The third equality is obtained through the following steps.
\begin{align*}
    V^{\bpi}_{\bpi}(\brho) - V^{\bpi'}_{\bpi'}(\brho)
    &= {V_{\bpi}^{\bpi}(\brho) - V_{\bpi}^{\bpi'}(\brho)}+  {V_{\bpi}^{\bpi'}(\brho)-V_{\bpi'}^{\bpi'}(\brho)}\notag\\
    &= \frac{1}{1-\gamma} \expectation_{(s,a)\sim \occupancy{\bpi,\brho}{\bpi}}[A_{\bpi}^{\bpi'}(s,a)] + V_{\bpi}^{\bpi'}(\brho)-V_{\bpi'}^{\bpi'}(\brho)\\
    &= \frac{1}{1-\gamma} \expectation_{(s,a)\sim \occupancy{\bpi,\brho}{\bpi}}[A_{\bpi}^{\bpi'}(s,a)] + \expectation_{a \sim \bpi'(\cdot|\brho)}\big[\reward_{\bpi}(s_0,a) - \reward_{\bpi'}(s_0,a)\big] \\
    &\qquad\qquad+ \gamma\sum_{s} \left(\transition_{\bpi}^{\bpi'} (s,s_0) - \transition_{\bpi'}^{\bpi'} (s,s_0)\right)V^{\bpi'}_{\bpi}(s) + \gamma\sum_{s} \transition_{\bpi'}^{\bpi'} (s,s_0)\left(V^{\bpi'}_{\bpi}(s)  -  V^{\bpi'}_{\bpi'}(s)\right) \\
    &= \frac{1}{1-\gamma} \expectation_{(s,a)\sim \occupancy{\bpi,\brho}{\bpi}}\Big[A_{\bpi}^{\bpi'}(s,a)\Big]\\
    &\qquad\qquad+ \frac{1}{1-\gamma} \expectation_{(s,a)\sim \occupancy{\bpi',\brho}{\bpi'}}\Big[(\reward_{\bpi}(s,a) - \reward_{\bpi'}(s,a)) 
    + \gamma (\transition_{\bpi}(\cdot|s,a) - \transition_{\bpi'}(\cdot|s,a))^\top V^{\bpi'}_{\bpi}(\cdot) \Big] \,. 
\end{align*}
\end{proof}

\begin{replemma}{lemma:perf_grad_upperbound}[Bounding Performative Performance Difference for Gradually Shifting Environments]
  Let us assume that both rewards and transitions are Lipschitz functions of policy, i.e. $\norm{\reward_{\bpi} - \reward_{\bpi'}}_1 \leq \lipreward \norm{\bpi-\bpi'}_\infty~~\text{and}~~\norm{\transition_{\bpi} - \transition_{\bpi'}}_1 \leq \liptransition \norm{\bpi-\bpi'}_\infty\,$, for some $\lipreward,\liptransition \geq 0$. Then, under Assumption~\ref{ass:bounded_r}, the performative shift in the sub-optimality gap of a policy $\bpi_\btheta$ satisfies 
  \begin{align}
    \Big| \subopt(\bpi_{\btheta}) -\frac{1}{1-\gamma}\expectation_{(s,a)\sim d_{\bpi_\btheta, \brho}^{\opt_o}}[A_{\bpi_\btheta}^{\bpi_\btheta}(s,a)] \Big|
    \leq \frac{2\sqrt{2}}{1-\gamma} (\lipreward + \frac{\gamma}{1-\gamma} \liptransition R_{\max}) \expectation_{s_0\sim \brho}\hellinger{\opt_o(\cdot|s_0)}{\bpi_\btheta(\cdot|s_0)} \,,
  \end{align}
  where $\hellinger{\bx}{\by}$ denotes the Hellinger distance between $\bx$ and $\by$.
\end{replemma}

\begin{proof}[Proof of Lemma~\ref{lemma:perf_grad_upperbound}]\,
We do this proof in three steps. We start from the final expression in Lemma \ref{lemma:perf_perf_diff}, then in step 2 we impose bounds on reward and transition differences leveraging the Lipschitz assumption. Lastly, we bound the policy difference in first order norm using relation between Total Variation (TV) and Hellinger distance. \\

\noindent \textbf{Step 1:} From Lemma~\ref{lemma:perf_perf_diff}, we get

\begin{align*}
\subopt(\bpi_{\btheta}\mid \brho) 
    =\frac{1}{1-\gamma} \expectation_{(s,a)\sim \occupancy{\bpi_\btheta,\brho}{\opt_o}}\Big[A_{\bpi_\btheta}^{\bpi_\btheta}(s,a) + (\reward_{\opt_o}(s,a) - \reward_{\bpi_\btheta}(s,a)) + \gamma (\transition_{\opt_o}(\cdot|s,a) - \transition_{\bpi_\btheta}(\cdot|s,a))^\top V^{\opt_o}_{\opt_o}(\cdot) \Big]\,.
\end{align*}
Thus,
\begin{align}
&~~~\Big| \subopt(\bpi_{\btheta}) -\frac{1}{1-\gamma}\expectation_{(s,a)\sim d_{\bpi_\btheta, \brho}^{\opt_o}}[A_{\bpi_\btheta}^{\bpi_\btheta}(s,a)] \Big| \notag \\
 &= \frac{1}{1-\gamma}\Big| \expectation_{(s,a)\sim \occupancy{\bpi_\btheta,\brho}{\opt_o}}(\reward_{\opt_o}(s,a) - \reward_{\bpi_\btheta}(s,a)) + \gamma (\transition_{\opt_o}(\cdot|s,a) - \transition_{\bpi_\btheta}(\cdot|s,a))^\top V^{\opt_o}_{\opt_o}(\cdot)\Big| \label{eq:upper_bound_main}
\end{align}


\noindent\textbf{Step 2: } Using Jensen's inequality together with the fact that $\occupancy{\bpi_\btheta}{\opt_o}(s,a|\brho)\leq 1$, for rewards, we get 
\begin{align*}
    \Bigg|\expectation_{(s,a) \sim \occupancy{\bpi_\btheta,\brho}{\opt_o}} \Big[\reward_{\opt_o}(s, a) - \reward_{\bpi_\btheta}(s,a)\Big]\Bigg| \leq \expectation_{(s,a) \sim \occupancy{\bpi_\btheta,\brho}{\opt_o}}\Big| \reward_{\opt_o}(s, a) - \reward_{\bpi_\btheta}(s,a) \Big| \leq \|\reward_{\opt_o} - \reward_{\bpi_\btheta}\|_1 
\end{align*}

Similarly for transitions, we get 
\begin{align*}
    \Bigg|\expectation_{(s,a) \sim \occupancy{\bpi_\btheta,\brho}{\opt_o}} \Big[ (\transition_{\opt_o} - \transition_{\bpi_\btheta})^\top V^{\bpi}_{\bpi} \Big]\Bigg|
    &\leq \expectation_{(s,a) \sim \occupancy{\bpi_\btheta,\brho}{\opt_o}} \Bigg| (\transition_{\opt_o} - \transition_{\bpi_\btheta})^\top V^{\bpi}_{\bpi} \Bigg|\\
    &\overset{(a)}{\leq} \expectation_{(s,a) \sim \occupancy{\bpi_\btheta,\brho}{\opt_o}} \Big[ \|\transition_{\opt_o} - \transition_{\bpi_\btheta}\|_1 \cdot \|V^{\opt_o}_{\opt_o}\|_\infty\Big] \\
    &= \|\transition_{\opt_o} - \transition_{\bpi_\btheta}\|_1 \cdot \|V^{\opt_o}_{\opt_o}\|_\infty\,,
\end{align*}
(a) holds due to H\"older's inequality. 

Now, leveraging the triangle inequality and Lipschitzness assumption on reward and transitions, we further get
\begin{align*}
    \Bigg|\expectation_{(s,a) \sim \occupancy{\bpi_\btheta,\brho}{\opt_o}} \Big[\reward_{\opt_o}(s, a) - \reward_{\bpi_\btheta}(s,a)  + \gamma (\transition_{\opt_o} - \transition_{\bpi_\btheta})^\top V^{\bpi}_{\bpi} \Big]\Bigg|\leq \lipreward \norm{\opt_o -\bpi_\btheta}_\infty + \gamma \liptransition \norm{\bV_{\opt_o}^{\opt_o}}_{\infty}\norm{\opt_o -\bpi_\btheta}_{\infty}
\end{align*}
Finally, due to Assumption~\ref{ass:bounded_r}, we get $ \norm{\bV_{\opt_o}^{\opt_o}}_{\infty} \leq \frac{R_{\max}}{1-\gamma}$, 
and thus,
\begin{align*}
    \Bigg|\expectation_{(s,a) \sim \occupancy{\bpi_\btheta,\brho}{\opt_o}} \Big[\reward_{\opt_o}(s, a) - \reward_{\bpi_\btheta}(s,a)  + \gamma (\transition_{\opt_o} - \transition_{\bpi_\btheta})^\top V^{\opt_o}_{\opt_o} \Big]\Bigg| \leq \lipreward \norm{\opt_o -\bpi_\btheta}_{\infty} +  \frac{\gamma}{1-\gamma} \liptransition R_{\max}\norm{\opt_o -\bpi_\btheta}_{\infty}
\end{align*}

\noindent \textbf{Step 3:} We know $\|\opt_o - \bpi_{\btheta}\|_{\infty} \leq\|\opt_o - \bpi_{\btheta}\|_1 = 2\TV{\opt_o}{\bpi_\btheta} \leq 2\sqrt{2}\hellinger{\opt_o}{\bpi_\btheta}$. Thus,
\begin{align}\label{eq:hellinger_upper_bound_second_term}
    &\Bigg|\expectation_{(s,a) \sim \occupancy{\bpi_\btheta,\brho}{\opt_o}} \Big[\reward_{\opt_o}(s, a) - \reward_{\bpi_\btheta}(s,a)  + \gamma (\transition_{\opt_o} - \transition_{\bpi_\btheta})^\top V^{\opt_o}_{\opt_o} \Big]\Bigg| \notag \\
    &\leq 2\sqrt{2}\left(\lipreward  +   \frac{\gamma}{1-\gamma} \liptransition R_{\max}\right)\hellinger{\opt_o(\cdot\mid s_0)}{\bpi_\btheta(\cdot\mid s_0)}
\end{align}

We conclude this proof by putting the upper bound in Equation \eqref{eq:hellinger_upper_bound_second_term} in Equation \eqref{eq:upper_bound_main} and taking expectation over $s_0 \sim \brho$ to get the desired expression.






\end{proof}

\section{Smoothness of Performative Value Function and Entropy Regulariser} \label{app:smoothness}
\begin{lemma}[Performative Smoothness Lemma]\label{lemm:perf_smooth}
    Let \( \bpi_\alpha \defn \bpi_{\theta + \alpha u} \), and let \( V_{\alpha}^{\alpha}(s_0) \) be the corresponding value at a fixed state \( s_0 \), i.e.,  $V_{\alpha}^{\alpha}(s_0) \defn V^{\bpi_\alpha}_{\bpi_\alpha}(s_0)\,.$
If the following conditions hold true,
\begin{align*}
& \sum_{a \in \mathcal{A}} \left. \left| \frac{\dd \bpi_\alpha(a \mid s_0)}{\dd \alpha} \right|_{\alpha = 0}\right| \leq C_1,
\quad
 \sum_{a \in \mathcal{A}}\left. \left| \frac{\dd^2 \bpi_\alpha(a \mid s_0)}{\dd\alpha^2} \right|_{\alpha = 0}\right| \leq C_2\,, \sum_{s \in \mathcal{S}} \left.\left| \frac{\dd \transition_\alpha(s \mid s_0,a_0)}{\dd\alpha} \right|_{\alpha = 0}\right| \leq T_1,\\
& \sum_{s \in \mathcal{S}} \left. \left| \frac{\dd^2 \transition_\alpha(s \mid s_0,a_0)}{\dd\alpha^2} \right|_{\alpha = 0}\right| \leq T_2\,,
 \sum_{a \in \mathcal{A}} \left. \left| \frac{\dd \reward_{\alpha}(s_0,a)}{\dd\alpha} \right|_{\alpha = 0}\right| \leq R_1,
\quad
 \sum_{a \in \mathcal{A}} \left. \left| \frac{\dd^2 \reward_{\alpha}(s_0,a)}{\dd\alpha^2} \right|_{\alpha = 0}\right| \leq R_2\,,
\end{align*}
we get
\[
\max_{\|u\|_2 = 1}
\left\| \left. \frac{\dd^2 V_{\alpha}^{\alpha}(s_0)}{\dd\alpha^2} \right|_{\alpha = 0} \right\|
\leq
\frac{C_2}{1-\gamma} + 2C_1\beta_1 + C_2\beta_2 \triangleq L
\,,
\]
\noindent where $ \beta_1 = \frac{\gamma}{(1 - \gamma)^2}(C_1 + T_1) + \frac{R_1}{1 - \gamma}$ and $\beta_2 =
\frac{2\gamma^2}{(1 - \gamma)^3} (C_1 + T_1)^2
+ \frac{\gamma}{(1 - \gamma)^2} (C_2 + 2C_1 T_1 + T_2)+ \frac{2\gamma R_1}{(1 - \gamma)^2} (C_2 + 2C_1 T_1 + T_2)
+ \frac{R_2}{1 - \gamma}
+ \frac{\gamma C_1 R_1}{(1 - \gamma)^2}$.
\end{lemma}

\begin{proof}

\noindent 
\textbf{Step 1:} To prove the second order smoothness of the value function we start by taking its second derivative. Consider the expected return under policy \( \bpi_\alpha \):
\[
V_{\alpha}^{\alpha}(s_0)  = \sum_a \bpi_\alpha(a \mid s_0) Q^{\alpha}_{\alpha}(s_0, a)
\]

\noindent Differentiating twice with respect to \( \alpha \), we obtain:
\[
\frac{\dd^2 V_{\alpha}^{\alpha}(s_0)}{\dd\alpha^2}
= \sum_a \frac{\dd^2 \bpi_\alpha(a \mid s_0)}{\dd\alpha^2} Q^\alpha_{\alpha}(s_0, a)
+ 2 \sum_a \frac{\dd \bpi_\alpha(a \mid s_0)}{\dd\alpha} \frac{\dd Q^{\alpha}_{\alpha}(s_0, a)}{\dd\alpha}
+ \sum_a \bpi_\alpha(a \mid s_0) \frac{\dd^2 Q^\alpha_{\alpha}(s_0, a)}{\dd\alpha^2}
\]

\noindent \( Q_\alpha^{\alpha}(s_0, a_0) \) is the Q-function corresponding to the policy \( \bpi_\alpha \) at state \( s_0 \) and action \( a_0 \). Observe that \( Q_\alpha^{\alpha}(s_0, a_0) \) can further be written as:
\[
Q_\alpha^{\alpha}(s_0, a_0) = e^{\top}_{(s_0, a_0)} (I - \gamma \tilde\transition(\alpha))^{-1} \reward_\alpha = e^{\top}_{(s_0, a_0)} M(\alpha) \reward_{\alpha}
\]

\noindent where $M(\alpha) \defn (I - \gamma \transition(\alpha))^{-1}$ and \( \tilde{\transition}(\alpha) \) is the state-action transition matrix under policy \( \bpi_\alpha \), defined as:
\[
[\tilde{\transition}(\alpha)](s', a'\mid s,a) \defn \bpi_\alpha(a' \mid s') \transition_\alpha(s' \mid s,a)
\]

\noindent Differentiating \( Q_\alpha^{\alpha}(s,a) \) with respect to \( \alpha \) gives:

\[
\frac{\dd Q^\alpha_{\alpha}(s_0, a_0)}{\dd\alpha}
= \gamma e_{(s_0, a_0)}^\top M(\alpha) \frac{\dd \tilde{\transition}(\alpha)}{\dd\alpha} M(\alpha) \reward_{\alpha} 
+ e_{(s_0, a_0)}^\top M(\alpha) \frac{\dd \reward_{\alpha}}{\dd\alpha}
\]
\noindent And correspondingly,

\begin{align} \label{eq:Q_derivative}
\frac{\dd^2 Q^\alpha_{\alpha}(s_0, a_0)}{\dd\alpha^2}
&= 2\gamma^2 e_{(s_0, a_0)}^\top M(\alpha) \frac{\dd \tilde{\transition}(\alpha)}{\dd\alpha} M(\alpha) \frac{\dd \tilde{\transition}(\alpha)}{\dd\alpha} M(\alpha) \reward_{\alpha}
+ \gamma e_{(s_0, a_0)}^\top M(\alpha) \frac{\dd^2 \tilde{\transition}(\alpha)}{\dd\alpha^2} M(\alpha) \reward_{\alpha} \nonumber\\
&+ \gamma e_{(s_0, a_0)}^\top M(\alpha) \frac{\dd \tilde{\transition}(\alpha)}{\dd\alpha} M(\alpha) \frac{\dd \reward_{\alpha}}{\dd\alpha}
+ e_{(s_0, a_0)}^\top M(\alpha) \frac{\dd^2 \reward_{\alpha}}{\dd\alpha^2}\nonumber\\
&+ \gamma e_{(s_0, a_0)}^\top M(\alpha) \frac{\dd \tilde{\transition}(\alpha)}{\dd\alpha} M(\alpha) \frac{\dd \reward_{\alpha}}{\dd\alpha}
\end{align}

\noindent 
\textbf{Step 2:} Now we need to find the derivative of $\tilde{\transition}(\alpha)$ w.r.t $\alpha$ in order to substitute in \eqref{eq:Q_derivative}.
Hence, we can differentiate \( \tilde\transition(\alpha) \) with respect to \( \alpha \) to obtain:
\begin{align*}
\left.\frac{\dd \tilde\transition(\alpha)}{\dd\alpha} \right|_{\alpha=0}(s', a'\mid s,a)   = \left.\frac{\dd \bpi_\alpha(a' \mid s')}{\dd\alpha} \right|_{\alpha=0} \transition_\alpha(s' \mid s,a) + \left.\frac{\dd \transition_\alpha(s'\mid s,a)}{\dd\alpha} \right|_{\alpha=0} \bpi_\alpha(a' \mid s')
\end{align*}

\noindent Now, for an arbitrary vector \( \bx \), we have:
\begin{align*}
\left[ \left. \frac{\dd \tilde\transition(\alpha)}{\dd\alpha} \right|_{\alpha=0} \bx \right]_{(s,a)} 
&= \sum_{s', a'} \left. \frac{\dd \bpi_\alpha(a' \mid s')}{\dd\alpha} \right|_{\alpha=0} \transition_\alpha(s' \mid s,a) \bx_{s',a'} \notag\\
&+ \sum_{s', a'}\left.\frac{\dd \transition_\alpha(s'\mid s,a)}{\dd\alpha} \right|_{\alpha=0} \bpi_\alpha(a' \mid s')\bx_{s',a'}
\end{align*}

\noindent Taking the maximum over unit vectors \( \bu \) in \( \ell_2 \)-norm:
\begin{align} 
\max_{\|\bu\|_2 = 1} \left\| \left. \frac{\dd \tilde\transition(\alpha)}{\dd\alpha} \right|_{\alpha=0} \bx \right\|_\infty 
&\leq \max_{s,a} \max_{\|\bu\|_2 = 1} \left| \sum_{s',a'} \left. \frac{\dd \bpi_\alpha(a' \mid s')}{\dd\alpha} \right|_{\alpha=0} \transition_\alpha(s' \mid s,a) \bx_{s',a'} \right| \nonumber\\
&\qquad+  \max_{s,a}  \max_{\|\bu\|_2 = 1} \left| \sum_{s',a'} \left. \frac{\dd \transition_\alpha(s' \mid s,a)}{\dd\alpha} \right|_{\alpha=0} \bpi_\alpha(a' \mid s')\bx_{s',a'} \right| \nonumber\\
&\leq \max_{s,a} \sum_{s'} \transition_\alpha(s' \mid s,a) \sum_{a'} \left| \left. \frac{\dd \bpi_\alpha(a' \mid s')}{\dd\alpha} \right|_{\alpha=0} \right| \cdot \|\bx\|_\infty \nonumber\\
&\qquad\qquad+ \max_{s,a} \sum_{a'} \bpi_\alpha(a' \mid s') \sum_{s'} \left| \left. \frac{\dd \transition_\alpha(s' \mid s,a)}{\dd\alpha} \right|_{\alpha=0} \right| \cdot \|\bx\|_\infty \nonumber\\
&\leq (C_1 + T_1) \|\bx\|_\infty \nonumber\\
&\leq C_1 + T_1 \label{eq:transition_alpha_bound}
\end{align}


\noindent Similarly, differentiating $\tilde{\transition}(\alpha)$ twice w.r.t. $\alpha$, we get

\begin{align*}
 \bigg[\frac{\dd^2 \tilde{\transition}(\alpha)}{\dd\alpha^2} \big|_{\alpha=0} \bigg]_{(s,a) \rightarrow (s',a')} &= 
\left. \frac{\dd^2 \bpi_\alpha(a' \mid s')}{(\dd\alpha)^2} \right|_{\alpha=0} \transition_\alpha(s' \mid s,a)
+ \left. \frac{\dd^2 \transition_\alpha(s' \mid s,a)}{\dd\alpha^2} \right|_{\alpha=0} \bpi_\alpha(a' \mid s') \\
&+  2 \left. \frac{\dd \bpi_\alpha(a' \mid s')}{\dd\alpha} \right|_{\alpha=0} 
\left. \frac{\dd \transition_\alpha(s' \mid s,a)}{\dd\alpha} \right|_{\alpha=0}
\end{align*}

\vspace{1em}

\noindent 
Hence, we can consider the following norm bound:
\begin{align} \label{eq:transition_alpha_double_bound}
\max_{\|\bu\|_2 = 1} \left\| \left. \frac{\dd^2 \tilde{\transition}(\alpha)}{\dd\alpha^2} \right|_{\alpha=0} \bx \right\|_1
\leq C_2 \|\bx\|_\infty + 2 C_1 T_1 \|\bx\|_\infty + T_2 \|\bx\|_\infty
\leq C_2 + 2 C_1 T_1 + T_2
\end{align}

\textbf{Step 3:} 
\noindent Now we need to put the pieces back together in order to calculate the second derivative of $V^{\alpha}_{\alpha}$ w.r.t $\alpha$. Let us recall \( M(\alpha) \). Using the power series expansion of the matrix inverse, we can write \( M(\alpha) \) as:
\[
M(\alpha) = (I - \gamma \tilde\transition(\alpha))^{-1} = \sum_{n=0}^{\infty} \gamma^n \tilde\transition(\alpha)^n
\]
\noindent which implies that \( M(\alpha) \geq 0 \) (component-wise), and
\[
M(\alpha)\mathbf{1} = \frac{1}{1 - \gamma} \mathbf{1},
\]
i.e., each row of \( M(\alpha) \) is positive and sums to \( \frac{1}{1 - \gamma} \).

\noindent This implies:
\[
\max_{\|u\|_2 = 1} \|M(\alpha)\bx\|_\infty \leq \frac{1}{1 - \gamma} \|\bx\|_\infty.
\]

\noindent This gives, using the expressions for \( \frac{\dd^2 Q^\alpha_{\alpha}(s_0, a_0)}{\dd\alpha^2} \) and \( \frac{\dd Q^\alpha_{\alpha}(s_0, a_0)}{\dd\alpha} \), an upper bound on their magnitudes based on \( \|\bx\|_\infty \) and constants arising from bounds on the derivatives of \( \tilde{\transition}(\alpha) \) and \( \reward_{\alpha} \).

\begin{align*}
&\max_{\|\bu\|_2 = 1} \left\| \frac{\dd^2 Q^\alpha_{\alpha}(s_0, a_0)}{\dd\alpha^2} \right\|_{\infty}\\
\leq& 
2\gamma^2 \left\| M(\alpha) \frac{\dd\tilde{\transition}(\alpha)}{\dd\alpha} M(\alpha) \frac{\dd\tilde{\transition}(\alpha)}{\dd\alpha} M(\alpha) \reward_{\alpha} \right\|_{\infty}
+ \gamma\left\| M(\alpha) \frac{\dd^2 \tilde{\transition}(\alpha)}{\dd\alpha^2} M(\alpha) \reward_{\alpha} \right\|_{\infty}\\
&+ \gamma \left\| M(\alpha) \frac{\dd^2 \tilde{\transition}(\alpha)}{\dd\alpha^2} M(\alpha) \frac{\dd \reward_{\alpha}}{\dd\alpha} \right\|_{\infty}+\left\| M(\alpha) \frac{\dd^2 \reward_{\alpha}}{\dd\alpha^2} \right\|_{\infty}
+ 2\gamma \left\| M(\alpha) \frac{\dd\tilde{\transition}(\alpha)}{\dd\alpha} M(\alpha) \frac{\dd \reward_{\alpha}}{\dd\alpha} \right\|_{\infty}
\end{align*}

\noindent Bounding using known bounds on transitions and rewards:

\begin{align*}
\max_{\|\bu\|_2 = 1} \left\| \frac{\dd^2 Q^\alpha_{\alpha}(s_0, a_0)}{\dd\alpha^2} \right\|_{\infty}
&\leq 
\frac{2\gamma^2}{(1 - \gamma)^3} (C_1 + T_1)^2
+ \frac{\gamma}{(1 - \gamma)^2} (C_2 + 2C_1 T_1 + T_2)\\
&+ \frac{2\gamma R_1}{(1 - \gamma)^2} (C_2 + 2C_1 T_1 + T_2)
+ \frac{R_2}{1 - \gamma}
+ \frac{\gamma C_1 R_1}{(1 - \gamma)^2} = \beta_2
\end{align*}

\noindent Corresponding bound on the first derivative is:

\begin{align*}
\max_{\|\bu\|_2 = 1} \left\| \frac{\dd Q^\alpha_{\alpha}(s_0, a_0)}{\dd\alpha} \right\|_\infty
&\leq 
\gamma \left\| M(\alpha) \frac{\dd\tilde{\transition}(\alpha)}{\dd\alpha} M(\alpha) \frac{\dd \reward_{\alpha}}{\dd\alpha} \right\|_{\infty}
+ \left\| M(\alpha) \frac{\dd \reward_{\alpha}}{\dd\alpha} \right\|_\infty\\
&\leq \frac{\gamma}{(1 - \gamma)^2}(C_1 + T_1) + \frac{R_1}{1 - \gamma} = \beta_1
\end{align*}

\textbf{Step 4:}
\noindent Finally, putting all the bounds together to evaluate the upper bound of the desired quantity, we get,

\begin{align}
\max_{\|\mathbf{u}\|_2 = 1} \left\| \frac{\dd^2 V_{\alpha}^{\alpha}(s_0)}{\dd\alpha^2} \right\|_\infty \leq \frac{C_2}{1-\gamma} + 2C_1\beta_1 + \beta_2
\end{align}

\end{proof}

\begin{corollary}[Smoothness guaranty for Exponential PeMDPs]\label{cor:smoothness_constants_softmax}
    For exponential PeMDPs, we characterise
    \begin{align*}
    &C_1 = 2, \quad  C_2 = 6,\quad
    T_1 = \max_{s} |\psi(s)| \triangleq \psi_{\max}, \quad T_2 =  \max_{s} |\psi(s)|^2, \quad R_1 = \xi|\cA|, \quad R_2 = 0
\end{align*}
Thus, 
\begin{align}\label{eq:smoothness_constant_unregularised}
\max_{\|u\|_2 = 1} \left\| \left. \frac{\dd^2 V_{\alpha}^{\alpha}(s_0)}{\dd\alpha^2} \right|_{\alpha = 0} \right\|
\leq \bigO\left( \max\Bigg\lbrace\frac{\gamma R_{\max}\mid\cA\mid}{(1-\gamma)^2}, \frac{\gamma^2}{(1-\gamma)^3} \Bigg\rbrace \right) \triangleq \bigO\left( L\right)\,.
\end{align}
\end{corollary}
\begin{proof}
We use the expressions already found in \eqref{eq:individual_grads} to state the following: 

\begin{align*}
\sum_{a \in \mathcal{A}} 
    \left. \left|\frac{\dd}{\dd\alpha}\, \bpi_{\btheta+\alpha \bu}(a \mid s) \right|_{\alpha=0}
    \right|\le
    \sum_{a \in \mathcal{A}} 
    \bpi_\btheta(a \mid s)\, \left|\bu_s^\top \big(\be_a - \bpi(\cdot \mid s)\big)\right|
    \le
    \max_{a \in \mathcal{A}} \Big( \bu_s^\top \be_a + \bu_s^\top \bpi(\cdot \mid s) \Big)
    \le 2.
\end{align*}

Similarly, differentiating once again w.r.t.\ $\alpha$, we get
\begin{align*}
\sum_{a \in \mathcal{A}} 
    \left| \frac{\dd^2}{\dd\alpha^2}\, \bpi_{\btheta+\alpha \bu}(a \mid s) \Big|_{\alpha=0}\right|
    \le
    \max_{a \in \mathcal{A}} \Big(&
        \bu_s^\top \be_a \be_a^\top \bu_s
        + \bu_s^\top \be_a \bpi(\cdot \mid s)^\top \bu_s
        + \bu_s^\top \bpi(\cdot \mid s) \be_a^\top \bu_s \\
        &+ 2\, \bu_s^\top \bpi(\cdot \mid s)\bpi(\cdot \mid s)^\top \bu_s
        + \bu_s^\top \mathrm{diag}(\bpi(\cdot \mid s)) \bu_s
    \Big)
    \le 6.
\end{align*}

And hence for transition we get,
\begin{align*}
    \sum_{s' \in \cS} 
    \left. \left|\frac{\dd}{\dd\alpha}\, \transition_{\bpi_{\btheta+\alpha \bu}}(\cdot \mid s,a) \right|_{\alpha=0}
    \right| &\le
    \sum_{s' \in \cS} \Bigg(\Bigg|\psi(s')
    \transition_{\bpi_\btheta}(s' \mid s,a)\, \bu_{s',a}\Bigg| + \left|\transition_{\bpi_\btheta}(s'|s,a) \sum_{s''}\transition_{\bpi_\btheta}(s''|s,a)) \psi(s'')\bu_{s'',a}\right|\Bigg) \notag\\
    &\le
      2\max_{s} |\psi(s)| = \bigO\left(\psi_{\max}\right)
\end{align*}

And similarly, it can be shown that:

\begin{align*}
    \sum_{s' \in \mathcal{S}} 
    \left. \left|\frac{\dd^2}{\dd\alpha^2}\, \transition_{\bpi_{\btheta+\alpha \bu}}(\cdot \mid s,a) \right|_{\alpha=0}
    \right| \le \bigO\left(\max_{s} |\psi(s)|^2 \right)
\end{align*}

Similarly for rewards we get:

\begin{align*}
\sum_{a \in \mathcal{A}} 
    \left. \left|\frac{\dd}{\dd\alpha}\, \reward_{\bpi_{\btheta+\alpha \bu}}(s,a) \right|_{\alpha=0}
    \right|\le \xi |\cA| \qquad, \qquad \sum_{a \in \mathcal{A}} 
    \left. \left|\frac{\dd^2}{\dd\alpha^2}\, \reward_{\bpi_{\btheta+\alpha \bu}}(s,a) \right|_{\alpha=0}
    \right| = 0
\end{align*}

Hence, we can use the following choice of constants for softmax parametrization, 
\begin{align*}
    &C_1 = 2 \quad , \quad C_2 = 6\\
    &T_1 = \bigO(\max_{s} |\psi(s)|) \quad , \quad T_2 =  \bigO(\max_{s} |\psi(s)|^2)\\
    &R_1 = \xi|\cA| \quad , \quad R_2 = 0
\end{align*}
to get the desired order of 
$\max_{\|u\|_2 = 1}
\left\| \left. \frac{\dd^2 V_{\alpha}^{\alpha}(s_0)}{\dd\alpha^2} \right|_{\alpha = 0} \right\|$.

\end{proof}

\begin{lemma}[Smoothness of Entropy Regularizer]\label{lemma:regularizer_smoothness}
Define the discounted entropy regularizer as:
\begin{align*}   \entropy^{\bpi_{\btheta_\alpha}}_{\bpi_{\btheta_\alpha}}(s) 
= \mathbb{E}_{\tau \sim \transition^\bpi_\bpi}
\left[ \sum_{t=0}^{\infty} -\gamma^t \log \bpi_{\btheta \alpha}(a_t \mid s_t) \right]
\end{align*}
Under the same assumptions as Lemma \ref{lemm:perf_smooth}, the following holds:
\begin{align*}
    \max_{\|u\|_2=1}\left\|\frac{\partial ^2 \entropy^{\bpi_{\btheta_ \alpha}}_{\bpi_{\btheta_ \alpha}}(s)}{\partial \alpha^2}\Bigg|_{\alpha=0}\right\|_\infty \leq \beta_\lambda 
\end{align*}

where 
\begin{align*}
    \beta_\lambda =& 2\gamma^2  \frac{3  (1 + \log |\cA|)}{1-\gamma} + \gamma  \frac{2  \log |\cA|}{(1-\gamma)^2}(C_1+T_1)
    + 2\gamma   \frac{\log |\cA|}{(1-\gamma)^2} (C_2+2C_1 T_1+T_2) + \frac{\log |\cA|}{(1-\gamma)^3} (C_1+T_1)^2\,.
\end{align*}
    
\end{lemma}

\begin{proof}
    \textbf{Step 1:} Define the state-wise entropy term:
    \begin{align*}      
h_{\btheta_\alpha}(s) 
= - \sum_{a} \bpi_{\btheta_\alpha}(a \mid s) \, \log \bpi_{\btheta_\alpha}(a \mid s).
    \end{align*}
    From \cite{mei2020global} (Lemma 7) we report that,
    \begin{align}
        \left\|\frac{\partial h_{\btheta_\alpha}}{\partial \alpha}\right\|_{\infty} 
    \leq 2 \cdot \log |\cA| \cdot \|u\|_2,
\qquad
\left\|\frac{\partial^2 h_{\btheta_\alpha}}{\partial \alpha^2}\right\|_{\infty} 
    \leq 3 \cdot (1 + \log |\cA|) \cdot \|\bu\|_2^2.
    \end{align}
    Additionally, \cite{mei2020global} also presents a second result expressing the second derivative of the entropy w.r.t $\alpha$,
    \begin{align*}      
\frac{\partial^2 \entropy^{\bpi_{\btheta_ \alpha}}_{\bpi_{\btheta_ \alpha}}(s)}{\partial \alpha^2}
=& 2 \gamma^2 \, \be_s^\top M(\alpha) \frac{\partial \transition(\alpha)}{\partial \alpha} 
    M(\alpha) \frac{\partial \transition(\alpha)}{\partial \alpha} M(\alpha) h_{\btheta_\alpha} \\
&+ \gamma \, \be_s^\top M(\alpha) \frac{\partial^2 \transition(\alpha)}{\partial \alpha^2} 
    M(\alpha) h_{\btheta_\alpha}
+ 2 \gamma \, \be_s^\top M(\alpha) \frac{\partial \transition(\alpha)}{\partial \alpha} 
    M(\alpha) \frac{\partial h_{\btheta_\alpha}}{\partial \alpha}
+ \be_s^\top M(\alpha) \frac{\partial^2 h_{\btheta_\alpha}}{\partial \alpha^2}.
\end{align*}

\textbf{Step 2:} Now we proceed with bounding the absolute value of each term which will contribute towards bounding the overall second derivative of the regulariser.

For the last term,

\begin{align*}
\left|\be_s^\top M(\alpha) \frac{\partial^2 h_{\btheta_\alpha}}{\partial \alpha^2}\Big|_{\alpha=0}\right|
&\leq \|\be_s^\top\|_1 \cdot 
    \Big\|M(\alpha) \frac{\partial^2 h_{\btheta_\alpha}}{\partial \alpha^2}\Big|_{\alpha=0}\Big\|_\infty \\
&\leq \frac{1}{1-\gamma} \cdot 
    \left\|\frac{\partial^2 h_{\btheta_\alpha}}{\partial \alpha^2}\Big|_{\alpha=0}\right\|_\infty \\
&\leq \frac{3 \cdot (1 + \log |\cA|)}{1-\gamma} \cdot \|\bu\|_2^2.
\end{align*}

For the second last term,

\begin{align*}
\left|\be_s^\top M(\alpha) \frac{\partial \transition(\alpha)}{\partial \alpha} 
    M(\alpha) \frac{\partial h_{\btheta_\alpha}}{\partial \alpha}\Big|_{\alpha=0}\right|
&\leq 
\Big\| M(\alpha) \frac{\partial \transition(\alpha)}{\partial \alpha} 
    M(\alpha) \frac{\partial h_{\btheta_\alpha}}{\partial \alpha}\Big|_{\alpha=0} \Big\|_\infty \\
&\leq \frac{1}{1-\gamma} \cdot 
    \Big\|\frac{\partial \transition(\alpha)}{\partial \alpha} 
    M(\alpha) \frac{\partial h_{\btheta_\alpha}}{\partial \alpha}\Big|_{\alpha=0}\Big\|_\infty \\
&\leq \frac{(C_1+T_1) \cdot \|u\|_2}{1-\gamma} \cdot 
    \Big\|M(\alpha) \frac{\partial h_{\btheta_\alpha}}{\partial \alpha}\Big|_{\alpha=0}\Big\|_\infty \\
&\leq \frac{(C_1+T_1) \cdot \|\bu\|_2}{(1-\gamma)^2} \cdot 
    \Big\|\frac{\partial h_{\btheta_\alpha}}{\partial \alpha}\Big|_{\alpha=0}\Big\|_\infty \\
&\leq \frac{2 \cdot \log |\cA|}{(1-\gamma)^2} (C_1+T_1)\cdot \|\bu\|_2^2.
\end{align*}

For the second term,

\begin{align*}
\left|\be_s^\top M(\alpha) \frac{\partial^2 \transition(\alpha)}{\partial \alpha^2} 
    M(\alpha) h_{\btheta_\alpha}\Big|_{\alpha=0}\right|
&\leq 
\Big\| M(\alpha) \frac{\partial^2 \transition(\alpha)}{\partial \alpha^2} 
    M(\alpha) h_{\btheta_\alpha}\Big|_{\alpha=0} \Big\|_\infty \\
&\leq \frac{1}{1-\gamma} \cdot 
    \Big\|\frac{\partial^2 \transition(\alpha)}{\partial \alpha^2} 
    M(\alpha) h_{\btheta_\alpha}\Big|_{\alpha=0}\Big\|_\infty \\
&\leq \frac{\|\bu\|_2^2}{1-\gamma} \cdot 
    \Big\|M(\alpha) h_{\btheta_\alpha}\Big|_{\alpha=0}\Big\|_\infty (C_2+2C_1 T_1+T_2)\\
&\leq \frac{ \|\bu\|_2^2}{(1-\gamma)^2} \cdot 
    \Big\|h_{\btheta_\alpha}\Big|_{\alpha=0}\Big\|_\infty (C_2+2C_1 T_1+T_2)\\
&\leq \frac{\log |\cA|}{(1-\gamma)^2} (C_2+2C_1 T_1+T_2) \cdot \|\bu\|_2^2.
\end{align*}

For the first term,

\begin{align*}
\left|\be_s^\top M(\alpha) \frac{\partial \transition(\alpha)}{\partial \alpha} 
    M(\alpha) \frac{\partial \transition(\alpha)}{\partial \alpha} 
    M(\alpha) h_{\btheta_\alpha}\Big|_{\alpha=0}\right|
&\leq 
\Big\| M(\alpha) \frac{\partial \transition(\alpha)}{\partial \alpha} 
    M(\alpha) \frac{\partial \transition(\alpha)}{\partial \alpha} 
    M(\alpha) h_{\btheta_\alpha}\Big|_{\alpha=0}\Big\|_\infty \\
&\leq \frac{1}{1-\gamma} \cdot \|\bu\|_2 \cdot 
       \frac{1}{1-\gamma} \cdot \|\bu\|_2 \cdot 
       \frac{1}{1-\gamma} \cdot \log |\cA| \\
       &~~~~\cdot (C_1+T_1)^2\\
&= \frac{\log |\cA|}{(1-\gamma)^3} (C_1+T_1)^2\cdot \|\bu\|_2^2.
\end{align*}

\textbf{Step 3:} Now combining all the above equations, we get the final expression,

\begin{align*}
    \max_{\|\bu\|_2=1}\left\|\frac{\partial ^2 \entropy^{\bpi_{\btheta_ \alpha}}_{\bpi_{\btheta_ \alpha}}(s)}{\partial \alpha^2}\Bigg|_{\alpha=0}\right\|_\infty \leq \beta_\lambda 
\end{align*}

where 
\begin{align*}
    \beta_\lambda =& 2\gamma^2 \cdot \frac{3 \cdot (1 + \log |\cA|)}{1-\gamma} + \gamma \cdot \frac{2 \cdot \log |\cA|}{(1-\gamma)^2} (C_1+T_1)\\
    &+ 2\gamma \cdot  \frac{\log |\cA|}{(1-\gamma)^2} (C_2+2C_1 T_1+T_2) + \frac{\log |\cA|}{(1-\gamma)^3} (C_1+T_1)^2
\end{align*}
\end{proof}

\begin{corollary}\label{cor:smoothness_constants_softmax_reg}
    For PeMDPs with Softmax policy class and exponential transitions, the following holds:
    \begin{align*}
\max_{\|u\|_2 = 1} \left\| \left. \frac{\dd^2 \softV_{\alpha}^{\alpha}(s_0)}{\dd\alpha^2} \right|_{\alpha = 0} \right\|
\leq\bigO\left( \max\Bigg\lbrace\frac{\gamma R_{\max}\mid\cA\mid}{(1-\gamma)^2}, \frac{\lambda \log|\cA| \psi_{\max}^2}{(1-\gamma)^3} \Bigg\rbrace \right)\,.
\end{align*}
\end{corollary}

\begin{proof}
By definition of smoothness, the ``soft performative value function" $\softV^{\bpi}_{\bpi}$ is Lipschitz smooth with Lipschitz constant $L_\lambda$ where
$L_\lambda \defn L + \lambda\beta_\lambda$. Once again, we can choose $C_1,C_2,T_1,T_2$ according to Corollary \ref{cor:smoothness_constants_softmax} for simplification to get the order $\beta_{\lambda} = \bigO\left( \frac{\log|\cA|}{(1-\gamma)^3} \psi_{\max}^2 \right)$. Thus, the final bound for $L_{\lambda}$ as 
\begin{align}\label{eq:smoothness_constant_regularised}
    L_{\lambda} = \bigO\left( \max\left\{ L, \lambda \beta_{\lambda} \right\}\right) = \bigO\left( \max\Bigg\lbrace\frac{\gamma R_{\max}\mid\cA\mid}{(1-\gamma)^2}, \frac{\lambda \log|\cA| \psi_{\max}^2}{(1-\gamma)^3} \Bigg\rbrace \right)\,.
\end{align}
\end{proof}

\newpage
\section{Derivation of Performative Policy Gradients}
\begin{reptheorem}{thm:perf_pg_theorem}[Performative Policy Gradient Theorem]

The gradient of the performative value function w.r.t $\btheta$ is as follows:

    (a) For the unregularised objective, 
    \begin{align*} 
         \nabla_\btheta V_{\bpi_\btheta}^{\bpi_\btheta}(\tau) = \expectation_{\tau \sim \prob_{\bpi_\btheta}^{\bpi_\btheta}} \left[
\sum_{t=0}^{\infty} \gamma^t \Big(A_{\bpi_\btheta}^{\bpi_\btheta}(s_t, a_t) \left(\nabla_\btheta \log \bpi_\btheta(a_t \mid s_t) +   \nabla_\btheta \log P_{\bpi_\btheta}(s_{t+1}|s_t,a_t)\right) + \nabla_\btheta r_{\bpi_\btheta}(s_t, a_t)\Big)\right]\,.
\end{align*}

(b) For the entropy-regularised objective, we define the soft advantage, soft Q, and soft value functions with respect to the soft rewards $\tilde r_{\bpi_\btheta}$ satisfying $\tilde A^{\bpi_\btheta}_{\bpi_\btheta}(s,a) = \tilde Q^{\bpi_\btheta}_{\bpi_\btheta}(s,a) - \tilde V^{\bpi_\btheta}_{\bpi_\btheta}(s)$ that further yields
\begin{align*}
    \nabla_\btheta \tilde V_{\bpi_\btheta}^{\bpi_\btheta}(\tau) = \expectation_{\tau \sim \prob_{\bpi_\btheta}^{\bpi_\btheta}} \Bigg[
    \sum_{t=0}^{\infty} \gamma^t \Big(\tilde A_{\bpi_\btheta}^{\bpi_\btheta}(s_t, a_t) \left( \nabla_\btheta \log \bpi_\btheta(a_t \mid s_t) +  \nabla_\btheta \log P_{\bpi_\btheta}(s_{t+1}|s_t,a_t)\right) + \nabla_\btheta \tilde{r}_{\bpi_\btheta}(s_t, a_t|\btheta) \Big)\Bigg]\,.
\end{align*}
\end{reptheorem}

\begin{proof}[Proof of Theorem~\ref{thm:perf_pg_theorem}]
We prove each part of this theorem separately.

\textit{Proof of part (a).} First, we derive explicit closed form gradient for unregularised performative value function.

\textbf{Step 1.} Given a trajectory $\tau =\lbrace s_0, a_0, \ldots, s_t, a_t, \ldots\rbrace$, let us denote the unregularised objective function as
\[
f_\btheta(\tau) = \sum_{t=0}^{\infty} \gamma^t \reward_{\bpi_\btheta}(s_t, a_t)
\]

Thus, 
\begin{align*}
\nabla_\btheta V_{\bpi_\btheta}^{\bpi_\btheta}(\tau) = \nabla_\btheta \expectation_{\tau \sim \prob_{\bpi_\btheta}^{\bpi_\btheta}} [f_\btheta(\tau)] 
&= \nabla_\btheta \sum_{\tau} \prob_{\bpi_\btheta}^{\bpi_\btheta}(\tau) f_\btheta(\tau)  \\
&= \sum_{\tau} \nabla_\btheta (\prob_{\bpi_\btheta}^{\bpi_\btheta}(\tau) f_\btheta(\tau))   \\
&= \sum_{\tau} (\nabla_\btheta \prob_{\bpi_\btheta}^{\bpi_\btheta}(\tau)) f_\btheta(\tau)  + \sum_{\tau} \prob_{\bpi_\btheta}^{\bpi_\btheta}(\tau)(\nabla_\btheta f_\btheta(\tau))\quad\\
&\overset{\text{(a)}}{=} \sum_{\tau} \prob_{\bpi_\btheta}^{\bpi_\btheta}(\tau) (\nabla_\btheta \log \prob_{\bpi_\btheta}^{\bpi_\btheta}(\tau)) f_\btheta(\tau)  + \expectation_{\tau \sim \prob_{\bpi_\btheta}^{\bpi_\btheta}} [\nabla_\btheta f_\btheta(\tau)] \quad \\
&= \expectation_{\tau \sim \prob_{\bpi_\btheta}^{\bpi_\btheta}} \left[ (\nabla_\btheta \log \prob_{\bpi_\btheta}^{\bpi_\btheta}(\tau)) f_\btheta(\tau) \right] + \expectation_{\tau \sim \prob_{\bpi_\btheta}^{\bpi_\btheta}} [\nabla_\btheta f_\btheta(\tau)]\,.
\end{align*}

\noindent $(a)$ holds since  $\nabla_{\btheta} \log \prob_{\bpi_\btheta}^{\bpi_\btheta}(\tau) = \frac{\nabla_{\btheta} \prob_{\bpi_\btheta}^{\bpi_\btheta}(\tau)}{\prob_{\bpi_\btheta}^{\bpi_\btheta}(\tau)}$.

\textbf{Step 2.} Given the initial state distribution $\brho$, we further have
\begin{align*}
\log \prob_{\bpi_\btheta}^{\bpi_\btheta}(\tau) 
&= \log \brho(s_0) + \sum_{t=0}^{\infty} \log \bpi_\btheta(a_t \mid s_t) + \sum_{t=0}^{\infty} \log \transition_{\bpi_\btheta}(s_{t+1}|s_t, a_t) 
\end{align*}

\noindent Taking the gradient with respect to $\btheta$, we obtain
\begin{align*}
\nabla_\btheta \log \prob_{\bpi_\btheta}^{\bpi_\btheta}(\tau) 
&= \sum_{t=0}^{\infty} \nabla_\btheta \log \bpi_\btheta(a_t \mid s_t) + \sum_{t=0}^{\infty} \nabla_\btheta \log \transition_{\bpi_\btheta}(s_{t+1}|s_t, a_t)
\end{align*}

\textbf{Step 3.} Now, by substituting the value of $\nabla_{\btheta}\log(\transition_{\bpi_\btheta}^{\bpi_\btheta})$ in $\nabla_\btheta V_{\bpi_\btheta}^{\bpi_\btheta}(\tau)$, we get,
\begin{align*}
\nabla_\btheta V_{\bpi_\btheta}^{\bpi_\btheta}(\tau)&= \nabla_\btheta \expectation_{\tau \sim \prob_{\bpi_\btheta}^{\bpi_\btheta}} [f_\btheta(\tau)] \notag\\
&= 
\expectation_{\tau \sim \prob_{\bpi_\btheta}^{\bpi_\btheta}} \left[
\left( \sum_{t=0}^{\infty} \nabla_\btheta \log \bpi_\btheta(a_t \mid s_t) \right)
\cdot
\left( \sum_{t=0}^{\infty} \gamma^t r_{\bpi_\btheta}(s_t, a_t) \right)
\right]\\
&\qquad\qquad\qquad + \expectation_{\tau \sim \prob_{\bpi_\btheta}^{\bpi_\btheta}} \left[
\left( \sum_{t=0}^{\infty} \nabla_\btheta \log \transition_{\bpi_\btheta}(s_{t+1}|s_{t},a_{t}) \right)
\cdot
\left( \sum_{t=0}^{\infty} \gamma^t r_{\bpi_\btheta}(s_t, a_t) \right)\right] \\
&\qquad\qquad\qquad+ \expectation_{\tau \sim \prob_{\bpi_\btheta}^{\bpi_\btheta}} \left[ \sum_{t=0}^{\infty} \gamma^t \nabla_\btheta r_{{\bpi_\btheta}}(s_t, a_t)\right]\\
&=\expectation_{\tau \sim \prob_{\bpi_\btheta}^{\bpi_\btheta}} \left[
\sum_{t=0}^{\infty}\gamma^t A_{\bpi_\btheta}^{\bpi_\btheta}(s_t, a_t) \nabla_\btheta \log \bpi_\btheta(a_t \mid s_t) 
\right] + \expectation_{\tau \sim \prob_{\bpi_\btheta}^{\bpi_\btheta}} \left[
\sum_{t=0}^{\infty} \gamma^t A_{\bpi_\btheta}^{\bpi_\btheta}(s_t, a_t)  \nabla_\btheta \log \transition_{\bpi_\btheta}(s_{t+1}|s_{t},a_{t}) \right]\\& \qquad\qquad\qquad+ \expectation_{\tau \sim \prob_{\bpi_\btheta}^{\bpi_\btheta}} \left[ \sum_{t=0}^{\infty} \gamma^t \nabla_\btheta \reward_{\bpi_\btheta}(s_t, a_t)\right]\,.
\end{align*}
The last equality is due to the definition of advantage function \begin{align*}
A_{\bpi_\btheta}^{\bpi_\btheta}(s_t, a_t) &\triangleq \sum_{i=t+1}^{\infty} \gamma^{t-i} \reward_{\bpi_\btheta}(s_i, \bpi_\btheta(s_i)) - \expectation_{\substack{s_{t'+1}\sim \transition_{\bpi_\btheta}(\cdot|s_{t'},a_{t'})\\\forall t' \in [t, \infty)}}\left[\sum_{i=t+1}^{\infty} \gamma^{t-i} {\reward}_{\bpi_\btheta}(s_i, \bpi_\btheta(s_i))|(s_t, a_t)\right] \notag\\
&\triangleq Q_{\bpi_\btheta}^{\bpi_\btheta}(s_t, a_t) - V_{\bpi_\btheta}^{\bpi_\btheta}(s_t) 
\end{align*}
as in classical policy gradient theorem. Hence, we conclude the proof for part (a) of the theorem. 

\textit{Proof of part (b).} Now, we derive explicit gradient form for entropy-regularised value function.

Let us define the soft reward as $\softreward_{\bpi_\btheta}(s_t, a_t)  \triangleq r_{\bpi_\btheta}(s_t, a_t) - \lambda \log \bpi_\btheta (a_t|s_t)$. Again, we start by defining regularised objective function
\[
\tilde{f}_\btheta(\tau) = \sum_{t=0}^{\infty} \gamma^t \softreward_{\bpi_\btheta}(s_t, a_t)
\]

\noindent Following the same steps as that of \textit{Part (a)}, we get
\begin{align*}
\nabla_\btheta \softV_{\bpi_\btheta}^{\bpi_\btheta}(\tau) &= \nabla_\btheta \expectation_{\tau \sim \prob_{\bpi_\btheta}^{\bpi_\btheta}} [\tilde{f}_\btheta(\tau)] \notag\\ &=\expectation_{\tau \sim \prob_{\bpi_\btheta}^{\bpi_\btheta}} \left[
\sum_{t=0}^{\infty}\gamma^t \tilde A_{\bpi_\btheta}^{\bpi_\btheta}(s_t, a_t) \nabla_\btheta \log \bpi_\btheta(a_t \mid s_t) 
\right] + \expectation_{\tau \sim \prob_{\bpi_\btheta}^{\bpi_\btheta}} \left[
\sum_{t=0}^{\infty} \gamma^t \tilde A_{\bpi_\btheta}^{\bpi_\btheta}(s_t, a_t)  \nabla_\btheta \log \transition_{\bpi_\btheta}(s_{t+1}|s_{t},a_{t}) \right]\\
&\qquad\qquad\qquad+ \expectation_{\tau \sim \prob_{\bpi_\btheta}^{\bpi_\btheta}} \left[ \sum_{t=0}^{\infty} \gamma^t \nabla_\btheta \softreward_{\bpi_\btheta}(s_t, a_t)\right]\,. \\
& = \expectation_{\tau \sim \prob_{\bpi_\btheta}^{\bpi_\btheta}} \left[
\sum_{t=0}^{\infty}\gamma^t \tilde A_{\bpi_\btheta}^{\bpi_\btheta}(s_t, a_t) \nabla_\btheta \log \bpi_\btheta(a_t \mid s_t) 
\right] + \expectation_{\tau \sim \prob_{\bpi_\btheta}^{\bpi_\btheta}} \left[
\sum_{t=0}^{\infty} \gamma^t \tilde A_{\bpi_\btheta}^{\bpi_\btheta}(s_t, a_t)  \nabla_\btheta \log \transition_{\bpi_\btheta}(s_{t+1}|s_{t},a_{t}) \right] \\
& +  \expectation_{\tau \sim \prob_{\bpi_\btheta}^{\bpi_\btheta}} \left[ \sum_{t=0}^{\infty} \gamma^t \nabla_\btheta \reward_{\bpi_\btheta}(s_t, a_t)\right] 
- \lambda \expectation_{\tau \sim \prob_{\bpi_\btheta}^{\bpi_\btheta}} \left[ \sum_{t=0}^{\infty} \gamma^t \nabla_\btheta \log \bpi_\btheta (a_t|s_t) \right]
\end{align*}
Here, 
\begin{align*}
\tilde{A}_{\bpi_\btheta}^{\bpi_\btheta}(s_t, a_t) &\triangleq \sum_{i=t+1}^{\infty} \gamma^{t-i} \tilde{\reward}_{\bpi_\btheta}(s_i, \bpi_\btheta(s_i)) - \expectation_{\substack{s_{t'+1}\sim \transition_{\bpi_\btheta}^{\bpi_\btheta}(\cdot|s_{t'},a_{t'})\\\forall t' \in [t, \infty)}}\left[\sum_{i=t}^{\infty} \gamma^{t-i} \tilde{\reward}_{\bpi_\btheta}(s_i, \bpi_\btheta(s_i))|(s_t,a_t)\right] \notag\\
&\triangleq \tilde{Q}_{\bpi_\btheta}^{\bpi_\btheta}(s_t,a_t) - \tilde{V}_{\bpi_\btheta}^{\bpi_\btheta}(s_t)
\end{align*}
denotes the advantage function with soft rewards, or in brief, the soft advantage function. 
Hence, we conclude proof of part (b). 
\end{proof}\clearpage
%
\section{Convergence results under Assumption~\ref{ass:generic}}\label{app:generic}

\begin{definition}
    The discounted state occupancy measure $\occupancy{\bpi'}{\bpi}(s|s_0)$ induced by a policy $\bpi$ and an MDP environment defined by $\bpi'$ is defined as 
    \begin{align*}
        \occupancy{\bpi'}{\bpi}(s|\brho) \defn \sum_{a \in \cA}\occupancy{\bpi'}{\bpi}(s,a|\brho) = (1-\gamma)\,\sum_{a \in \cA}\expectation_{\tau\sim\prob_{\bpi'}^{\bpi}}\Big[\sum_{t=0}^{\infty}\gamma^t\indicator\{s_t=s, a_t =a\}\Big]\,.
    \end{align*}
\end{definition}

   \begin{replemma}{lemma:grad_domm_generic}
         We define $\Cov \triangleq \max_{\btheta, \bnu} \Big\|\frac{\occupancy{\bpi_\btheta,\brho}{\opt_o}}{\occupancy{\bpi_\btheta,\bnu}{\bpi_\btheta}}\Big\|_{\infty}$, then
        (a) for unregularised value function, $\subopt(\bpi_{\btheta}) \leq {\sqrt{|\cS||\cA|}} \Cov \|\nabla_\btheta V^{\bpi_\btheta}_{\bpi_\btheta}(\bnu)\|_2 + \frac{1+\Cov}{(1-\gamma)^2}\Big(\lipreward + \liptransition R_{\max}\Big)$, (b) for cross-entropy regularised value function $\subopt(\bpi_{\btheta}\mid\blambda)\leq {\sqrt{|\cS||\cA|}} \Cov \|\nabla_\btheta\softV^{\bpi_\btheta}_{\bpi_\btheta}(\bnu)\|_2 + \frac{2+\Cov}{(1-\gamma)^2}\Big(\lipreward + \liptransition (R_{\max}+\lambda \log |\cA|)\Big) $ 
\end{replemma}

\begin{proof}[Proof of part - (a)]
This proof is divided into two parts. In the first part we bound the expected advantage term from Lemma~\ref{lemma:perf_grad_upperbound} with the norm of the gradient of value function. During this step, we need to express the expected advantage as a linear combination of the advantage itself and the occupancy measure over all states and actions. The expectation however is taken w.r.t the occupancy measure $\occupancy{\bpi_\btheta,\brho}{\opt_o}$, thus we need to perform a change of measure which introduces a coverage term as shown below. In the second step we directly use the bound of rewards and transitions obtained from their Lipchitzness in lemma \ref{lemma:perf_grad_upperbound}. We know by Lemma~\ref{lemma:perf_perf_diff} that
\begin{align*}
    \subopt(\bpi_{\btheta})
    = &~\frac{1}{1-\gamma} \expectation_{(s,a) \sim \occupancy{\bpi_\btheta,\brho}{\opt_o}}[A_{\bpi_\btheta}^{\bpi_\btheta}(s,a)] 
    +  \frac{1}{1-\gamma} \expectation_{(s,a)\sim \occupancy{\bpi_\btheta,\brho}{\opt_o}}[(\reward_{\opt_o}(s,a) - \reward_{\bpi_\btheta}(s,a)) \notag\\
    &\qquad \qquad \qquad + \gamma (\transition_{\opt_o}(\cdot|s,a) - \transition_{\bpi_\btheta}(\cdot|s,a))^\top V^{\opt_o}_{\opt_o}(\cdot) \Big] \,.
\end{align*}

\textbf{Step 1: Upper bounding Term 1.}  
Let us define $\text{Term} 1 \defn \frac{1}{1-\gamma}\expectation_{(s,a) \sim \occupancy{\bpi_\btheta,\brho}{\opt_o}(\cdot\mid s_0)}[A_{\bpi_\btheta}^{\bpi_\btheta}(s,a)] $

\begin{align} \label{eq:advantage_bound_generic}
        \expectation_{(s,a)\sim \occupancy{\bpi_\btheta,\brho}{\opt_o}}[A_{\bpi_\btheta}^{\bpi_\btheta}(s,a)] 
        = \sum_{s,a} \occupancy{\bpi_\btheta,\brho}{\opt_o}(s,a|\brho)A^{\bpi_\btheta}_{\bpi_\btheta}(s,a)
        &= \sum_{s,a} \frac{\occupancy{\bpi_\btheta,\brho}{\opt_o}(s,a|\brho)}{\occupancy{\bpi_\btheta}{\bpi_\btheta}(s,a|\bnu)}\occupancy{\bpi_\btheta}{\bpi_\btheta}(s,a|\bnu)A^{\bpi_\btheta}_{\bpi_\btheta}(s,a) \nonumber\\
        &\leq \Bigg\|\frac{\occupancy{\bpi_\btheta,\brho}{\opt_o}}{\occupancy{\bpi_\btheta, \bnu}{\bpi_\btheta}}\Bigg\|_{\infty} \sum_{s,a} \occupancy{\bpi_\btheta}{\bpi_\btheta}(s,a|\bnu)A^{\bpi_\btheta}_{\bpi_\btheta}(s,a)\notag\\
        &\leq \Cov \sum_{s,a} \occupancy{\bpi_\btheta}{\bpi_\btheta}(s,a|\bnu)A^{\bpi_\btheta}_{\bpi_\btheta}(s,a)
    \end{align}
\textbf{Step 2: Generic Gradient Calculation} Now, we leverage the gradient of generic performative MDPs,

    \begin{align*} 
        \frac{\partial}{\partial \btheta_{s,a}} V_{\bpi_\btheta}^{\bpi_\btheta}(\bnu) 
        =& \expectation_{\tau \sim \prob_{\bpi_\btheta,\bnu}^{\bpi_\btheta}} \Big[ \sum_{t=0}^{\infty} \gamma^t \Big(A_{\bpi_\btheta}^{\bpi_\btheta}(s_t, a_t) \frac{\partial}{\partial \btheta_{s,a}}\log \bpi_\btheta(a_t \mid s_t)+ A_{\bpi_\btheta}^{\bpi_\btheta}(s_t, a_t)  \frac{\partial}{\partial \btheta_{s,a}} \log P_{\bpi_\btheta}(s_{t+1}|s_t,a_t)\\
        &\qquad\qquad+ \frac{\partial}{\partial \btheta_{s,a}} \reward_{\bpi_\btheta}(s_t, a_t)\Big)\Big]\\
        \underset{(a)}{\geq}& \expectation_{\tau \sim \prob_{\bpi_\btheta,\bnu}^{\bpi_\btheta}} \Big[ \sum_{t=0}^{\infty} \gamma^t  A_{\bpi_\btheta}^{\bpi_\btheta}(s_t, a_t) \indicator[s_t=s,a_t=a]\Big] - \expectation_{\tau \sim \prob_{\bpi_\btheta,\bnu}^{\bpi_\btheta}} \Big[ \sum_{t=0}^{\infty} \gamma^t \bpi_\btheta(a|s) \indicator[s_t=s] A_{\bpi_\btheta}^{\bpi_\btheta}(s_t, a_t)\Big]\\
        &\qquad\qquad- \expectation_{\tau \sim \prob_{\bpi_\btheta,\bnu}^{\bpi_\btheta}} \Big[ \sum_{t=0}^{\infty} \gamma^t \lipreward \indicator[s_t=s, a_t=a]\Big] - \expectation_{\tau \sim \prob_{\bpi_\btheta,\bnu}^{\bpi_\btheta}} \Big[ \sum_{t=0}^{\infty} \gamma^t  A_{\bpi_\btheta}^{\bpi_\btheta}(s_t, a_t) \indicator[s_t=s, a_t=a] \liptransition \Big]  \\
        \underset{(b)}{\geq}& \frac{1}{1-\gamma}d_{\bpi_\btheta,\bnu}^{\bpi_\btheta}(s,a) A_{\bpi_\btheta}^{\bpi_\btheta}(s, a)
        - \frac{1}{(1-\gamma)^2}d_{\bpi_\btheta,\bnu}^{\bpi_\btheta}(s,a) {R_{\max}\liptransition}
        - \frac{1}{1-\gamma}\lipreward d_{\bpi_\btheta,\bnu}^{\bpi_\btheta}(s,a)
\end{align*}  
(a) holds due to Lipchitzness of rewards and transitions and (b) holds since $\expectation_{\tau \sim \prob_{\bpi_\btheta}^{\bpi_\btheta}} \Big[ \sum_{t=0}^{\infty} \gamma^t \bpi_\btheta(a|s)\indicator[s_t=s] A_{\bpi_\btheta}^{\bpi_\btheta}(s_t, a_t)\Big] = 0$. Hence, 
\begin{align*}
        \frac{1}{1-\gamma}\sum_{s,a} \occupancy{\bpi_\btheta}{\bpi_\btheta}(s,a|\bnu)A^{\bpi_\btheta}_{\bpi_\btheta}(s,a) 
        &\leq  \sum_{s,a} \frac{\partial V^{\bpi_\btheta}_{\bpi_\btheta}(\bnu)}{\partial \btheta_{s,a}} + \frac{\lipreward}{1-\gamma}  + \frac{R_{\max}\liptransition}{(1-\gamma)^2}\\
        &\leq  \sqrt{|\cS||\cA|} \|\nabla_\btheta V^{\bpi_\btheta}_{\bpi_\btheta}(\bnu)\|_2 + \frac{\lipreward}{1-\gamma}+ \frac{R_{\max}\liptransition}{(1-\gamma)^2}
    \end{align*}

    Now, substituting the above result back in Equation~\eqref{eq:advantage_bound_generic}, we get
    \begin{align}\label{eq:term1_lemma3_generic}
        \frac{1}{1-\gamma}\expectation_{(s,a) \sim \occupancy{\bpi_\btheta,\brho}{\opt_o}}[A_{\bpi_\btheta}^{\bpi_\btheta}(s,a)] &\leq \sqrt{|\cS||\cA|} \Cov \|\nabla_\btheta V^{\bpi_\btheta}_{\bpi_\btheta}(\bnu)\|_2 + \Big(\frac{\lipreward}{1-\gamma}+ \frac{R_{\max}\liptransition}{(1-\gamma)^2}\Big) \Cov
    \end{align}

    \textbf{Step 3: Upper bounding Term 2.} For lipchitz rewards and transitions, we further obtain from Lemma \ref{lemma:perf_grad_upperbound},
     \begin{align} \label{eq:generic_term2_upperbound}
        \text{Term }2 &\defn  \frac{1}{1-\gamma} \expectation_{(s,a)\sim \occupancy{\bpi_\btheta,\brho}{\opt_o}}\Big[(\reward_{\opt_o}(s,a) - \reward_{\bpi_\btheta}(s,a)) 
    + \gamma (\transition_{\opt_o}(\cdot|s,a) - \transition_{\bpi_\btheta}(\cdot|s,a))^\top V^{\opt_o}_{\opt_o}(\cdot) \Big] \,\notag\\
    &\leq \frac{1}{1-\gamma}\left(\lipreward  +   \frac{\gamma}{1-\gamma} \liptransition R_{\max}\right)
    \end{align}

    \textbf{Step 4:} Now, if we use Equation~\eqref{eq:term1_lemma3_generic} and~\eqref{eq:generic_term2_upperbound} together (and knowing $\gamma \leq1$), we get
    \begin{align*}
        \subopt(\bpi_{\btheta}) \leq  {\sqrt{|\cS||\cA|}} \Cov \|\nabla_\btheta V^{\bpi_\btheta}_{\bpi_\btheta}(\bnu)\|_2 + \frac{1+\Cov}{(1-\gamma)^2}\Bigg(\lipreward + \liptransition R_{\max}\Bigg) 
    \end{align*}
    \end{proof}

\begin{proof}[Proof of part - (b)]
We start by deriving a lower bound on the derivative of $\softV^{\bpi_\btheta}_{\bpi_\btheta}(\bnu)$
\begin{align*} 
        &\frac{\partial}{\partial \btheta_{s,a}} \softV_{\bpi_\btheta}^{\bpi_\btheta}(\bnu)\\ 
        &= \expectation_{\tau \sim \prob_{\bpi_\btheta,\bnu}^{\bpi_\btheta}} \Big[ \sum_{t=0}^{\infty} \gamma^t \Big(\softA_{\bpi_\btheta}^{\bpi_\btheta}(s_t, a_t) \frac{\partial}{\partial \btheta_{s,a}}\log \bpi_\btheta(a_t \mid s_t)+ \softA_{\bpi_\btheta}^{\bpi_\btheta}(s_t, a_t)  \frac{\partial}{\partial \btheta_{s,a}} \log P_{\bpi_\btheta}(s_{t+1}|s_t,a_t)\\
        &\qquad\qquad+ \frac{\partial}{\partial \btheta_{s,a}} \softreward_{\bpi_\btheta}(s_t, a_t)\Big)\Big]\\
        &\underset{(a)}{\geq} \expectation_{\tau \sim \prob_{\bpi_\btheta,\bnu}^{\bpi_\btheta}} \Big[ \sum_{t=0}^{\infty} \gamma^t  \softA_{\bpi_\btheta}^{\bpi_\btheta}(s_t, a_t) \indicator[s_t=s,a_t=a]\Big] 
        - \expectation_{\tau \sim \prob_{\bpi_\btheta,\bnu}^{\bpi_\btheta}} \Big[ \sum_{t=0}^{\infty} \gamma^t \bpi_\btheta(a|s) \indicator[s_t=s] \softA_{\bpi_\btheta}^{\bpi_\btheta}(s_t, a_t)\Big]\\
        &\qquad\qquad- \expectation_{\tau \sim \prob_{\bpi_\btheta,\bnu}^{\bpi_\btheta}} \Big[ \sum_{t=0}^{\infty} \gamma^t \lipreward \indicator[s_t=s, a_t=a]\Big] 
        - \expectation_{\tau \sim \prob_{\bpi_\btheta,\bnu}^{\bpi_\btheta}} \Big[ \sum_{t=0}^{\infty} \gamma^t  \softA_{\bpi_\btheta}^{\bpi_\btheta}(s_t, a_t) \indicator[s_t=s, a_t=a] \liptransition \Big]  \\
        &\underset{(b)}{=} \expectation_{\tau \sim \prob_{\bpi_\btheta,\bnu}^{\bpi_\btheta}} \Big[ \sum_{t=0}^{\infty} \gamma^t  \softA_{\bpi_\btheta}^{\bpi_\btheta}(s_t, a_t) \indicator[s_t=s,a_t=a]\Big] 
        + \bpi_\btheta(a|s)\expectation_{\tau \sim \prob_{\bpi_\btheta,\bnu}^{\bpi_\btheta}} \Big[ \sum_{t=0}^{\infty} \gamma^t \log \bpi_\btheta(a_t|s_t)\indicator[s_t=s] \Big]\\
        &\qquad\qquad- \expectation_{\tau \sim \prob_{\bpi_\btheta,\bnu}^{\bpi_\btheta}} \Big[ \sum_{t=0}^{\infty} \gamma^t \lipreward \indicator[s_t=s, a_t=a]\Big] 
        - \expectation_{\tau \sim \prob_{\bpi_\btheta,\bnu}^{\bpi_\btheta}} \Big[ \sum_{t=0}^{\infty} \gamma^t  \softA_{\bpi_\btheta}^{\bpi_\btheta}(s_t, a_t) \indicator[s_t=s, a_t=a] \liptransition \Big] \\
        &=\expectation_{\tau \sim \prob_{\bpi_\btheta,\bnu}^{\bpi_\btheta}} \Big[ \sum_{t=0}^{\infty} \gamma^t \softA_{\bpi_\btheta}^{\bpi_\btheta}(s_t, a_t) \indicator[s_t=s,a_t=a]\Big] 
        + \bpi_\btheta(a|s)\expectation_{\tau \sim \prob_{\bpi_\btheta,\bnu}^{\bpi_\btheta}} \Big[ \sum_{a}\log \bpi_\btheta(a|s)\sum_{t=0}^{\infty} \gamma^t \indicator[s_t=s,a_t=a] \Big]\\
        &\qquad\qquad- \expectation_{\tau \sim \prob_{\bpi_\btheta,\bnu}^{\bpi_\btheta}} \Big[ \sum_{t=0}^{\infty} \gamma^t \lipreward \indicator[s_t=s, a_t=a]\Big] 
        - \expectation_{\tau \sim \prob_{\bpi_\btheta,\bnu}^{\bpi_\btheta}} \Big[ \sum_{t=0}^{\infty} \gamma^t  \softA_{\bpi_\btheta}^{\bpi_\btheta}(s_t, a_t) \indicator[s_t=s, a_t=a] \liptransition \Big] \\   
        &= \frac{1}{1-\gamma}d_{\bpi_\btheta,\bnu}^{\bpi_\btheta}(s,a) \softA_{\bpi_\btheta}^{\bpi_\btheta}(s, a)
        - \frac{1}{(1-\gamma)^2}d_{\bpi_\btheta,\bnu}^{\bpi_\btheta}(s,a) {(R_{\max}+\lambda \log |\cA|)\liptransition}
        - \frac{1}{1-\gamma}\lipreward d_{\bpi_\btheta,\bnu}^{\bpi_\btheta}(s,a)\\
        &\qquad\qquad+ \frac{\lambda}{1-\gamma} \bpi_\btheta(a|s) \sum_{a} d_{\bpi_\btheta,\bnu}^{\bpi_\btheta}(s,a) \log \bpi_\btheta (a|s) \\
        &\underset{(c)}{\geq} \frac{1}{1-\gamma}d_{\bpi_\btheta,\bnu}^{\bpi_\btheta}(s,a) \softA_{\bpi_\btheta}^{\bpi_\btheta}(s, a)
        - \frac{1}{(1-\gamma)^2}d_{\bpi_\btheta,\bnu}^{\bpi_\btheta}(s,a) {(R_{\max} +\lambda \log |\cA|)\liptransition}
        - \frac{1}{1-\gamma}\lipreward d_{\bpi_\btheta,\bnu}^{\bpi_\btheta}(s,a)\\
        &\qquad\qquad-\frac{\lambda}{1-\gamma} d_{\bpi_\btheta,\bnu}^{\bpi_\btheta}(s,a) \log |\cA|
\end{align*}  
(a) holds due to Lipchitzness of rewards, transitions and also for the following:

\begin{align*}
    \expectation_{\tau \sim \prob^{\bpi_\btheta}_{\bpi_\btheta}}\left[\sum_{t=0}^\infty \gamma^t \frac{\partial}{\partial \theta_{s,a}} \log \bpi_\btheta(a_t|s_t) \right] &= \expectation_{\tau \sim \prob^{\bpi_\btheta}_{\bpi_\btheta}}\Big[ \sum_{t=0}^{\infty} \gamma^t \indicator[s_t=s,a_t=a] \Big] 
        - \expectation_{\tau \sim \prob^{\bpi_\btheta}_{\bpi_\btheta}}\Big[ \sum_{t=0}^{\infty} \gamma^t \bpi_\btheta(a|s) \indicator[s_t=s] \Big] \\
        &= d_{\bpi_\btheta,\brho}^{\bpi_\btheta}(s,a) - d_{\bpi_\btheta,\brho}^{\bpi_\btheta}(s)\bpi_\btheta (a|s) = 0
\end{align*}

(b) holds because:

\begin{align*}
    \expectation_{\tau \sim \prob_{\bpi_\btheta,\bnu}^{\bpi_\btheta}} \Big[ \sum_{t=0}^{\infty} \gamma^t \bpi_\btheta(a|s) \indicator[s_t=s] \softA_{\bpi_\btheta}^{\bpi_\btheta}(s_t, a_t)\Big] &= \expectation_{\tau \sim \prob_{\bpi_\btheta,\bnu}^{\bpi_\btheta}} \Big[ \sum_{t=0}^{\infty} \gamma^t \bpi_\btheta(a|s) \indicator[s_t=s] A_{\bpi_\btheta}^{\bpi_\btheta}(s_t, a_t)\Big] \\
    &- \expectation_{\tau \sim \prob_{\bpi_\btheta,\bnu}^{\bpi_\btheta}} \Big[ \sum_{t=0}^{\infty} \gamma^t \bpi_\btheta(a|s) \log \bpi_\btheta(a_t|s_t)\indicator[s_t=s] \Big] \\
    &=  - \bpi_\btheta(a|s) \Big[ \sum_{t=0}^{\infty} \gamma^t \log \bpi_\btheta(a_t|s_t)\indicator[s_t=s] \Big]
\end{align*}

And (c) holds since, 
\begin{align*}
    -\sum_{a} \occupancy{\bpi_\btheta}{\bpi_\btheta}(s,a|\bnu) \log {\bpi_\btheta(a|s)} &= \occupancy{\bpi_\btheta}{\bpi_\btheta}(s|\bnu)  \Big( - \sum_{a} \bpi_\btheta(a|s) \log \bpi_\btheta(a|s) \Big) \\
    &\underset{(d)}\leq \occupancy{\bpi_\btheta}{\bpi_\btheta}(s|\bnu)  \log |\cA| 
\end{align*}
while (d) holds as entropy is upper bounded by $\log |\cA|$~\citep[Theorem 2.6.4]{Cover2006}.

Hence, 
\begin{align}\label{eq:generic_reg_adv_upper_bound}
        \frac{1}{1-\gamma}\sum_{s,a} \occupancy{\bpi_\btheta}{\bpi_\btheta}(s,a|\bnu)A^{\bpi_\btheta}_{\bpi_\btheta}(s,a) 
        &\leq  \sum_{s,a} \frac{\partial V^{\bpi_\btheta}_{\bpi_\btheta}(\bnu)}{\partial \btheta_{s,a}} + \frac{\lipreward}{1-\gamma}  + \frac{(R_{\max}+\lambda \log |\cA|)\liptransition}{(1-\gamma)^2} + \frac{\lambda}{1-\gamma}\log |\cA|\notag\\
        &\leq  \sqrt{|\cS||\cA|} \|\nabla_\btheta V^{\bpi_\btheta}_{\bpi_\btheta}(\bnu)\|_2 + \frac{\lipreward}{1-\gamma}+ \frac{(R_{\max}+\lambda \log |\cA|)\liptransition}{(1-\gamma)^2}\notag\\
        &~+\frac{\lambda}{1-\gamma} \log |\cA|
    \end{align}

The last inequality holds from Cauchy-Schwartz. Now, upper-bounding term 2 we get from lemma \ref{lemma:reg_perf_grad_upperbound_generic},
\begin{align}\label{eq:generic_reg_subopt_upper_bound}
 \subopt(\bpi_{\btheta} \mid \lambda) &\leq \frac{1}{1-\gamma}\expectation_{(s,a)\sim \occupancy{\bpi_\btheta,\brho}{\opt_o}}[ \softA_{\bpi_\btheta}^{\bpi_\btheta}(s,a)] \nonumber\\
    &+\frac{1}{1-\gamma}   
    \Big( \lipreward + \liptransition \frac{\gamma (R_{\max} + \lambda \log |\cA|)}{1-\gamma} \Big)\,\nonumber\\ 
    &- \frac{\lambda}{1-\gamma} \sum_{s}  \occupancy{\bpi_\btheta,\brho}{\opt_o}(s) \KL{\opt_o(\cdot|s)}{\bpi_\btheta(\cdot|s)}
\end{align}

And we know,
\begin{align*} 
     -\KL{\opt_o(\cdot|s)}{\bpi_\btheta(\cdot|s)} \leq - \sum_{a\in\cA} \opt_o(a|s) \log \opt_o(a|s) \leq \log |\cA|
\end{align*}
 Hence, we get
 \begin{align}\label{eq:kl_bound_generic}
    - \sum_s \occupancy{\bpi_\btheta,\brho}{\opt_o}(s) \KL{\opt_o(\cdot|s)}{\bpi_\btheta(\cdot|s)} \leq \log |\cA|
\end{align}


Now, combining \eqref{eq:generic_reg_adv_upper_bound}, \eqref{eq:generic_reg_subopt_upper_bound}, \eqref{eq:kl_bound_generic}  and following the steps in the proof of Lemma \ref{lemma:grad_domm_generic}, we obtain the final gradient domination lemma.
\end{proof}




\begin{lemma}[Entropy-Regularized Performative Policy Difference: Generic Upper Bound]\label{lemma:reg_perf_grad_upperbound_generic}
  Under Assumption~\ref{ass:bounded_r} and ~\ref{ass:generic}~(a), the sub-optimality gap of a policy $\bpi_\btheta$ is 
  \begin{align} \label{eq:reg_perf_grad_upperbound}
    \subopt(\bpi_\btheta \mid \lambda)
    \leq &~\frac{1}{1-\gamma}\expectation_{(s,a)\sim \occupancy{\bpi_\btheta,\brho}{\opt_o}}[ \softA_{\bpi_\btheta}^{\bpi_\btheta}(s,a)] + \frac{1}{1-\gamma}  
    \Big( \lipreward + \liptransition \frac{\gamma (R_{\max} + \lambda \log |\cA|)}{1-\gamma} \Big)\, \notag\\
    &\qquad- \frac{\lambda}{1-\gamma} \sum_s \occupancy{\bpi_\btheta,\brho}{\opt_o}(s) \KL{\opt_o(\cdot|s)}{\bpi_\btheta(\cdot|s)}
  \end{align}
\end{lemma}

\begin{proof}
   This lemma follows the same sketch as Lemma \ref{lemma:perf_grad_upperbound} with an exception in the way the soft rewards are handled. The difference in the soft rewards equals the difference of the original rewards with a Lagrange dependent term. This term is the expected KL divergence over the state visitation distribution. Lemma \ref{lemma:perf_perf_diff} for regularized rewards reduces to,
  \begin{align} \label{eq:reg_perf_diff}
    \tilde V^{\bpi}_{\bpi}(\brho) - \tilde V^{\bpi'}_{\bpi'}(\brho)
    &= \frac{1}{1-\gamma} \expectation_{(s,a)\sim \occupancy{\bpi',\brho}{\bpi}}[\tilde A_{\bpi'}^{\bpi'}(s,a)] \notag\\
    &\qquad+\frac{1}{1-\gamma} \expectation_{(s,a) \sim \occupancy{\bpi',\brho}{\bpi}}  
    \Big( [\tilde \reward_{\bpi}(s, a) - \tilde \reward_{\bpi'}(s, a)] 
    + \gamma (\transition_{\bpi} - \transition_{\bpi'})^\top \tilde V^{\bpi}_{\bpi}(s_0) \Big) \,.
  \end{align}
  Therefore,
  \begin{align*}
       \tilde \reward_{\opt_o}(s,a) - \tilde\reward_{\bpi_\btheta}(s,a) &=  \reward_{\opt_o}(s,a) - \reward_{\bpi_\btheta}(s,a) + \lambda \Big(\log \bpi_\btheta(a|s) - \log \opt_o(a|s)\Big) \\
  \end{align*}
Therefore, we can write \eqref{eq:reg_perf_diff} in the following way,
  \begin{align*}
    \subopt(\bpi_\btheta \mid \lambda)
    =&~ \frac{1}{1-\gamma}\expectation_{(s,a)\sim \occupancy{\bpi_\btheta,\brho}{\opt_o}}[ \softA_{\bpi_\btheta}^{\bpi_\btheta}(s,a)] \nonumber\\
    &+\frac{1}{1-\gamma} \expectation_{(s,a) \sim \occupancy{\bpi_\btheta,\brho}{\opt_o}}  
    \Big( [\tilde \reward_{\opt_o}(s, a) - \tilde \reward_{\bpi_\btheta}(s, a)] 
    + \gamma (\transition_{\opt_o} - \transition_{\bpi_\btheta})^\top \tilde V^{\opt_o}_{\opt_o}(s_0) \Big)\,.\\ 
    &+ \frac{\lambda}{1-\gamma} \sum_{s,a} [\log \bpi_\btheta(a|s) - \log \opt_o(a|s)] \occupancy{\bpi_\btheta,\brho}{\opt_o}(s,a|s_0) \nonumber\\ 
    =&~ \frac{1}{1-\gamma}\expectation_{(s,a)\sim \occupancy{\bpi_\btheta,\brho}{\opt_o}}[ \softA_{\bpi_\btheta}^{\bpi_\btheta}(s,a)] \nonumber\\
    &+\frac{1}{1-\gamma} \expectation_{(s,a) \sim \occupancy{\bpi_\btheta,\brho}{\opt_o}}  
    \Big( [\tilde \reward_{\opt_o}(s, a) - \tilde \reward_{\bpi_\btheta}(s, a)] 
    + \gamma (\transition_{\opt_o} - \transition_{\bpi_\btheta})^\top \tilde V^{\opt_o}_{\opt_o}(s_0) \Big)\,\\ 
    &+ \frac{\lambda}{1-\gamma} \sum_{s,a}  \occupancy{\bpi_\btheta,\brho}{\opt_o}(s) \opt_o(a|s) [\log \bpi_\btheta(a|s) - \log \opt_o(a|s)] \nonumber\\ 
    \underset{(a)}{=}&~ \frac{1}{1-\gamma}\expectation_{(s,a)\sim \occupancy{\bpi_\btheta,\brho}{\opt_o}}[ \softA_{\bpi_\btheta}^{\bpi_\btheta}(s,a)] \nonumber\\
    &+\frac{1}{1-\gamma} \expectation_{(s,a) \sim \occupancy{\bpi_\btheta,\brho}{\opt_o}}  
    \Big( [\tilde \reward_{\opt_o}(s, a) - \tilde \reward_{\bpi_\btheta}(s, a)] 
    + \gamma (\transition_{\opt_o} - \transition_{\bpi_\btheta})^\top \tilde V^{\opt_o}_{\opt_o}(s_0) \Big)\,\\ 
    &- \frac{\lambda}{1-\gamma} \sum_{s}  \occupancy{\bpi_\btheta,\brho}{\opt_o}(s) \KL{\opt_o(\cdot|s)}{\bpi_\btheta(\cdot|s)}\nonumber\\ 
    \underset{\text{Holder's ineq.}}{\leq} &~ \frac{1}{1-\gamma}\expectation_{(s,a)\sim \occupancy{\bpi_\btheta,\brho}{\opt_o}}[ \softA_{\bpi_\btheta}^{\bpi_\btheta}(s,a)] \nonumber\\
    &+\frac{1}{1-\gamma} \expectation_{(s,a) \sim \occupancy{\bpi_\btheta,\brho}{\opt_o}}  
    \Big( [\tilde \reward_{\opt_o}(s, a) - \tilde \reward_{\bpi_\btheta}(s, a)] + \gamma \|\transition_{\opt_o} - \transition_{\bpi_\btheta}\|_1 \|\tilde V^{\opt_o}_{\opt_o}(s_0)\|_{\infty} \Big)\,\\ 
    &- \frac{\lambda}{1-\gamma} \sum_{s}  \occupancy{\bpi_\btheta,\brho}{\opt_o}(s) \KL{\opt_o(\cdot|s)}{\bpi_\btheta(\cdot|s)}\nonumber\\
    \underset{\text{(b)}}{\leq} &~ \frac{1}{1-\gamma}\expectation_{(s,a)\sim \occupancy{\bpi_\btheta,\brho}{\opt_o}}[ \softA_{\bpi_\btheta}^{\bpi_\btheta}(s,a)] \nonumber\\
    &+\frac{1}{1-\gamma} \expectation_{(s,a) \sim \occupancy{\bpi_\btheta,\brho}{\opt_o}}  
    \Big( [\tilde \reward_{\opt_o}(s, a) - \tilde \reward_{\bpi_\btheta}(s, a)] + \gamma \|\transition_{\opt_o} - \transition_{\bpi_\btheta}\|_1 \frac{R_{\max} + \lambda \log |\cA|}{1-\gamma} \Big)\,\\ 
    &- \frac{\lambda}{1-\gamma} \sum_{s}  \occupancy{\bpi_\btheta,\brho}{\opt_o}(s) \KL{\opt_o(\cdot|s)}{\bpi_\btheta(\cdot|s)}\nonumber\\
    \underset{\text{Lipschitz } \reward \text{ \& } \transition}{\leq} &~ \frac{1}{1-\gamma}\expectation_{(s,a)\sim \occupancy{\bpi_\btheta,\brho}{\opt_o}}[ \softA_{\bpi_\btheta}^{\bpi_\btheta}(s,a)]+\frac{1}{1-\gamma} \expectation_{(s,a) \sim \occupancy{\bpi_\btheta,\brho}{\opt_o}}  
    \Big( \lipreward + \liptransition \frac{\gamma (R_{\max} + \lambda \log |\cA|)}{1-\gamma} \Big) \|{\opt_o} - {\bpi_\btheta}\|_\infty\,\\ 
    &- \frac{\lambda}{1-\gamma} \sum_{s}  \occupancy{\bpi_\btheta,\brho}{\opt_o}(s) \KL{\opt_o(\cdot|s)}{\bpi_\btheta(\cdot|s)}\\
    \underset{(c)}{\leq} &~ \frac{1}{1-\gamma}\expectation_{(s,a)\sim \occupancy{\bpi_\btheta,\brho}{\opt_o}}[ \softA_{\bpi_\btheta}^{\bpi_\btheta}(s,a)] +\frac{1}{1-\gamma}  
    \Big( \lipreward + \liptransition \frac{\gamma (R_{\max} + \lambda \log |\cA|)}{1-\gamma} \Big)\,\\ 
    &- \frac{\lambda}{1-\gamma} \sum_{s}  \occupancy{\bpi_\btheta,\brho}{\opt_o}(s) \KL{\opt_o(\cdot|s)}{\bpi_\btheta(\cdot|s)}
    \end{align*}

The equality (a) holds since,
\begin{align*}
     \expectation_{a \sim \opt_o(\cdot \mid s)}[\log \bpi_\btheta(a|s) - \log \opt_o(a|s)] =- \KL{\opt_o(\cdot|s)}{\bpi_\btheta(\cdot|s)}
\end{align*}
The inequality (b) holds due to the result of~\cite{mei2020global}, i.e.
  \begin{align} \label{eq:reg_sup_value}
      \|\tilde V^{\opt_o}_{\opt_o}\|_{\infty} \leq \frac{R_{\max} + \lambda \log |\cA|}{1-\gamma}
  \end{align}
Finally, (c) is due to the fact that $\|{\opt_o}-{\bpi_\btheta}\|_\infty \leq 1$.
\end{proof}


Thus, as a consequence, we obtain convergence of \framework{}.
    \begin{reptheorem}{thm:conv}
We set learning rate $\eta = \bigO\left(\frac{(1-\gamma)^2}{|\cA|}\right)$. Then,
(a) For unregularised objective , we get
$\min_{t < T} \subopt(\bpi_{\btheta_t}) \leq \epsilon + \bigO\left(\frac{\Cov}{(1-\gamma)^2 }\right)$  when $T = \Omega\left( \frac{|\cS||\cA|^2{\rm Cov}^2}{\epsilon^2 (1-\gamma)^3}\right)$. 
(b) For entropy-regularised objective with $\lambda = \frac{(1-\gamma)R_{\max}}{1+2\log|\cA|}$, we get
$
    \min_{t < T} \subopt(\bpi_{\btheta_t}|\lambda) \;\; \leq \;\; \epsilon + \bigO\left(\frac{\Cov}{(1-\gamma)^2}\right)$   when  $T = \Omega\left( \frac{|\cS||\cA|^2{\rm Cov}^2}{\epsilon^2 (1-\gamma)^3}\right)$.
\end{reptheorem}

\begin{proof}[Proof of Theorem \ref{thm:conv}-- Part (a)]
We proceed with this proof by dividing it in four steps. In the first step, we use the smoothness of the value function to prove an upper bound for the minimum squared gradient norm of the value over time which is a constant times $1/T$ . In the second step, we derive a lower bound on the norm of gradient of value function using Lemma \ref{lemma:grad_domm_softmax}. In the final two steps, we combine the bounds obtained from the first two steps to derive lower bounds for $T$ and $\epsilon$, i.e. the error threshold. 

\textbf{Step 1:} As $V^{\bpi_\btheta}_{\bpi_\btheta}$ is $L$-smooth (Lemma~\ref{lemm:perf_smooth}), it satisfies  
    \begin{align*}
    \Big|V^{\bpi_\btheta}_{\bpi_\btheta}(\brho) - V^{\bpi_\btheta'}_{\bpi_\btheta'}(\brho )- \langle \nabla_\btheta V^{\bpi_\btheta}_{\bpi_\btheta}(\brho), \btheta - \btheta'\rangle \Big| \leq \frac{L}{2}\|\btheta- \btheta'\|^2
\end{align*}


Thus, taking $\btheta$ as $\btheta_{t+1}$ and $\btheta'$ as $\btheta_{t}$ and using the gradient ascent expression (Equation~\eqref{eqn:ascent_step}) yields
\begin{align*}
    \Big|V^{\bpi_{\btheta_{t+1}}}_{\bpi_{\btheta_{t+1}}}(\brho) - V^{\bpi_{\btheta_t}}_{\bpi_{\btheta_t}}(\brho) -\eta\|\nabla_{\btheta} V^{\bpi_{\btheta_t}}_{\bpi_{\btheta_t}}(\brho)\|^2\Big| &\leq \frac{L}{2}\|\btheta_{t+1}-\btheta_t\|^2 \\
    \implies\qquad\qquad\qquad\qquad\qquad V^{\bpi_{\btheta_{t+1}}}_{\bpi_{\btheta_{t+1}}}(\brho) - V^{\bpi_{\btheta_t}}_{\bpi_{\btheta_t}}(\brho) &\geq \eta\|\nabla V^{\bpi_{\btheta_t}}_{\bpi_{\btheta_t}}(\brho)\|^2 - \frac{L}{2}\|\btheta_{t+1}-\btheta_t\|^2
\end{align*}

This further implies that
\begin{align}\label{eq:grad_ineq_generic}
   V^{\bpi_{\btheta_{t+1}}}_{\bpi_{\btheta_{t+1}}}(\brho) - V^{\opt_o}_{\opt_o}(\brho) &\geq V^{\bpi_{\btheta_t}}_{\bpi_{\btheta_t}}(\brho) -V^{\opt_o}_{\opt_o}(\brho) + \eta\|\nabla_\btheta V^{\bpi_{\btheta_t}}_{\bpi_{\btheta_t}}(\brho)\|^2 - \frac{L}{2}\|\btheta_{t+1}-\btheta_t\|^2 \nonumber\\
    &= V^{\bpi_{\btheta_t}}_{\bpi_{\btheta_t}}(\brho) -V^{\opt_o}_{\opt_o}(\brho) + \eta(1-\frac{L\eta}{2})\|\nabla V^{\bpi_{\btheta_t}}_{\bpi_{\btheta_t}}(\brho)\|^2 
\end{align}
The last equality is due to Equation~\eqref{eqn:ascent_step}.

Now, telescoping Equation~\eqref{eq:grad_ineq_generic} leads to 
\begin{align}\label{eq:weighted_grad_bound_generic}
    \eta(1-\frac{L\eta}{2})\sum_{t=0}^{T-1} \|\nabla V^{\bpi_{\btheta_t}}_{\bpi_{\btheta_t}}(\brho)\|^2 &\leq \left(\subopt(\bpi_{\btheta_0}) \right) - \left(V^{\opt_o}_{\opt_o}(\brho) - V^{\bpi_{\btheta_T}}_{\bpi_{\btheta_T}}(\brho)  \right) \notag\\
    &\leq \left(\subopt(\bpi_{\btheta_0}) \right)
\end{align}
Since $\sum_{t=0}^{T-1} \|\nabla V^{\bpi_{\btheta_t}}_{\bpi_{\btheta_t}}(\brho)\|^2 \geq  T \min_{t\in [T-1]} \|\nabla V^{\bpi_{\btheta_t}}_{\bpi_{\btheta_t}}(\brho)\|^2$, we obtain 
\begin{align*}
    \min_{t\in [T-1]} \|\nabla V^{\bpi_{\btheta_t}}_{\bpi_{\btheta_t}}(\brho)\|^2 \leq \frac{1}{T\eta\left(1-\frac{L\eta}{2}\right)}\left(\subopt(\bpi_{\btheta_0}) \right) \leq \frac{R_{\max}}{T\eta\left(1-\frac{L\eta}{2}\right)(1-\gamma)}\,.
\end{align*}
The last inequality comes from $V^{\opt_o}_{\opt_o}(\brho) \leq \frac{R_{\max}}{1-\gamma}$ (Assumption~\ref{ass:bounded_r}).

\textbf{Step 2:} We derive from lemma \ref{lemma:grad_domm_generic} - part (a) that
\begin{align*}
 (\subopt(\bpi_{\btheta}))^2
&\leq \Bigg({\sqrt{|\cS||\cA|}} \Cov \|\nabla_\btheta V^{\bpi_\btheta}_{\bpi_\btheta}(\bnu)\|_2 + \frac{1+\Cov}{(1-\gamma)^2}\Bigg(\lipreward + \liptransition R_{\max}\Bigg)\Bigg)^2 \\
&\leq 2 {|\cS||\cA|} \Cov^2  \|\nabla_\btheta V^{\bpi_\btheta}_{\bpi_\btheta}(\bnu)\|_2^2 + 2 \Bigg( \frac{1+\Cov}{(1-\gamma)^2}\Bigg(\lipreward + \liptransition R_{\max}\Bigg)\Bigg)^2 \,.
\end{align*}

Thus, we further get
\begin{align*}
\min_{t \in [T-1]} (\subopt(\bpi_{\btheta_t}))^2 
&\leq 2 {|\cS||\cA|}{\rm Cov}^2 \min_{t \in [T-1]} \|\nabla_\btheta V^{\bpi_{\btheta_t}}_{\bpi_{\btheta_t}}(\bnu)\|_2^2 + 2 \Bigg( \frac{1+\Cov}{(1-\gamma)^2}\Bigg(\lipreward + \liptransition R_{\max}\Bigg)\Bigg)^2  \\
&\leq 2 {|\cS||\cA|}{\rm Cov}^2 \frac{R_{\max}}{T\eta\left(1-\frac{L\eta}{2}\right)(1-\gamma)} + 2 \Bigg( \frac{1+\Cov}{(1-\gamma)^2}\Bigg(\lipreward + \liptransition R_{\max}\Bigg)\Bigg)^2 \,.
\end{align*}
\textbf{Step 3:} Now, we set
\begin{align*}
    \min_{t
    \in [T-1]}(\subopt(\bpi_{\btheta_t}))^2 &\leq 2 {|\cS||\cA|}{\rm Cov}^2 \frac{R_{\max}}{T\eta\left(1-\frac{L\eta}{2}\right)(1-\gamma)} + 2 \Bigg( \frac{1+\Cov}{(1-\gamma)^2}\Bigg(\lipreward + \liptransition R_{\max}\Bigg)\Bigg)^2 \\
    &\leq \left(\sqrt{2 {|\cS||\cA|} \frac{R_{\max}}{T\eta\left(1-\frac{L\eta}{2}\right)(1-\gamma)}} {\rm Cov} + \frac{\sqrt{2}+\sqrt{2}\Cov}{(1-\gamma)^2}\Bigg(\lipreward + \liptransition R_{\max}\Bigg)\right)^2\\
    &\leq \left(\epsilon + \frac{\sqrt{2}+\sqrt{2}\Cov}{(1-\gamma)^2}\Bigg(\lipreward + \liptransition R_{\max}\Bigg)\right)^2\,,
\end{align*}
and solve for $T$ to get $T \ge \frac{2 {|\cS||\cA|}{\rm Cov}^2 R_{\max}}{\eta (1-\frac{L\eta}{2}) (1-\gamma) \epsilon^2} $

Choosing $\eta = \frac{1}{L}$, we get the final expression $T \ge \frac{4 L {|\cS||\cA|}{\rm Cov}^2 R_{\max}}{\epsilon^2(1-\gamma)} $,
for any $\epsilon > 0$ and the smoothness constant $L$ is same as in lemma \ref{lemm:perf_smooth}

Hence, we conclude that for $T = \Omega\left(\frac{|\cS||\cA| L~{\Cov}^2}{\epsilon^2 (1-\gamma)} \right)\,$, 
\begin{align*}
    \min_{t
    \in [T-1]}\subopt(\bpi_{\btheta_t}) 
    &\leq \epsilon + \bigO\left(\frac{\Cov}{(1-\gamma)^2}\right)\,.
\end{align*}
\end{proof}

\begin{proof}[Proof of Theorem \ref{thm:conv} - part (b)]

This proof follows similar steps as part (a) of Theorem~\ref{thm:conv} with two additional changes: (i) We have a $\lambda$, i.e. regularisation coefficient, dependent term due to the entropy regulariser. (ii) The maximum value of the soft value function is $\frac{R_{\max}+\lambda \log|\cA|}{1-\gamma}$ instead of $\frac{R_{\max}}{1-\gamma}$ for the unregularised value function. 

\textbf{Step 1:} From Equation~\eqref{eq:smoothness_constant_regularised}, we observe that the soft-value function $\softV^{\bpi_\btheta}_{\bpi_\btheta}$ is $L_\lambda$-smooth. 

Thus, following the Step 1 of part (a) of Theorem~\ref{thm:conv}, we get 
\begin{align}
    \min_{t\in [T-1]} \|\nabla \softV^{\bpi_{\btheta_t}}_{\bpi_{\btheta_t}}(\brho)\|^2 
    &\leq \frac{1}{T\eta\left(1-\frac{L_{\lambda}\eta}{2}\right)}\left(\softV^{\opt_o}_{\opt_o}(\brho) - \softV^{\bpi_{\btheta_0}}_{\bpi_{\btheta_0}}(\brho) \right)\leq \frac{R_{\max}+\lambda \log|\cA|}{T\eta\left(1-\frac{L_{\lambda}\eta}{2}\right)(1-\gamma)}\,.\label{eq:softmax_grad_bound}
\end{align}
The last inequality is true due to the fact that $\softV^{\opt_o}_{\opt_o}(\brho) - \softV^{\bpi_{\btheta_0}}_{\bpi_{\btheta_0}}(\brho) \leq \softV^{\opt_o}_{\opt_o}(\brho) \leq \frac{R_{\max}+\lambda \log|\cA|}{1-\gamma}$. 

\textbf{Step 2:} Now, from Part (b) of Lemma~\ref{lemma:grad_domm_generic}, we obtain that
\begin{align*}
  &\min_{t\in [T-1]} \left(\subopt(\btheta_t \mid \lambda)\right)^2 \nonumber\\
  \leq& 2 {|\cS||\cA|}{\rm Cov}^2 \frac{R_{\max}}{T\eta\left(1-\frac{L\eta}{2}\right)(1-\gamma)} + 2 \Bigg( \frac{2+\Cov}{(1-\gamma)^2}\Bigg(\lipreward + \liptransition (R_{\max}+\lambda \log |\cA|)\Bigg)\Bigg)^2
\end{align*}

Thus, we conclude that
\begin{align}
&~~\min_{t\in [T-1]} \subopt(\btheta_t \mid \lambda) \nonumber\\
&\leq \sqrt{\frac{2{|\cS||\cA|} {\Cov}^2 \left(R_{\max}+\lambda \log|\cA|\right)}{T\eta\left(1-\frac{L_{\lambda} \eta}{2}\right)(1-\gamma)}} \notag\\
&+ \sqrt{2} \Bigg( \frac{2+\Cov}{(1-\gamma)^2}\Bigg(\lipreward + \liptransition (R_{\max}+\lambda\log |\cA|)\Bigg)\Bigg),\label{eq:reg_conv_before_choosing_T_generic}\vspace*{-1em}
\end{align}

\textbf{Step 4:} Now, by setting the $T$-dependent term in Equation~\eqref{eq:reg_conv_before_choosing_T_generic} to $\epsilon$, we get $
    T \geq \frac{2{|\cS||\cA|} {\Cov}^2 \left(R_{\max}+\lambda \log|\cA|\right)}{\eta\left(1-\frac{L_{\lambda} \eta}{2}\right)(1-\gamma)\epsilon^2}\,.\vspace*{-0.5em}$
    
Choosing $\eta = \frac{1}{L_{\lambda}}$, $\lambda = \frac{(1-\gamma) R_{\max}}{(1+2\log|\cA|)}$,  we get the final expression $T \geq \frac{8{|\cS||\cA|} {\Cov}^2 L_{\lambda} R_{\max}}{(1-\gamma)\epsilon^2}\,$,
and
\begin{align*}
\min_{t\in [T-1]} \softV^{\opt_o}_{\opt_o}(\brho) - \softV^{\bpi_{\btheta_t}}_{\bpi_{\btheta_t}}(\brho) 
&\leq \epsilon + \bigO\left(\frac{\Cov}{(1-\gamma)^2}\right)\,.
\end{align*}

\end{proof}

\section{Convergence of \framework~ for Exponential PeMDPs} \label{app:conv_pepg}
\subsection{Proofs for Unregularised Value Function}

    \begin{lemma}[Performative Policy Gradient for Softmax PeMDPs]\label{lemm:grad_softmax}
    Given exponential family PeMDPs, for all $(s,a,s') \in (\mathcal{S},\mathcal{A},\mathcal{S})$, derivative of the performative value function w.r.t $\btheta_{s,a}$ satisfies:
    
\begin{align} \label{eq:simple_gradient_softmax}
        \frac{\partial V^{\bpi_\btheta}_{\bpi_\btheta}(\brho)}{\partial \btheta_{s,a}} \geq \frac{1}{1-\gamma}d_{\bpi_\btheta}^{\bpi_\btheta}(s,a|\brho) &\left(A_{\bpi_\btheta}^{\bpi_\btheta}(s, a) + \xi\right)\,.
    \end{align}  

\end{lemma}
\begin{proof}
   First, we note that    \begin{align}\label{eq:individual_grads}
        &\frac{\partial}{\partial \btheta_{s',a'}} \log \bpi_\btheta(a|s) = \indicator[s=s',a=a'] - \bpi_{\btheta}(a'|s)\indicator[s=s'] \notag\\
        &\frac{\partial}{\partial \btheta_{s',a'}} \log \transition_{\bpi_\btheta}(s''|s,a) = \psi(s'') \indicator[s=s',a=a']\left(1 - \transition_{\bpi_\btheta}(s''|s,a)\right) \notag \\
        &\frac{\partial}{\partial \btheta_{s',a'}} \reward_{\bpi_\btheta} (s,a) = \xi \indicator[s=s',a=a']\,.
    \end{align}
    In this proof, we further substitute the expressions of individual gradients in Equation~\eqref{eq:individual_grads} into Equation~\eqref{eq:perf_grad_unregularised}. 
    
    Therefore, for a given initial state distribution $\brho$, we get
    \begin{align*} 
        &\frac{\partial}{\partial \btheta_{s,a}} V_{\bpi_\btheta}^{\bpi_\btheta}(\brho)\\ 
        &= \expectation_{\tau \sim \prob_{\bpi_\btheta}^{\bpi_\btheta}} \Big[ \sum_{t=0}^{\infty} \gamma^t \Big(A_{\bpi_\btheta}^{\bpi_\btheta}(s_t, a_t) \frac{\partial}{\partial \btheta_{s,a}}\log \bpi_\btheta(a_t \mid s_t)+ A_{\bpi_\btheta}^{\bpi_\btheta}(s_t, a_t)  \frac{\partial}{\partial \btheta_{s,a}} \log P_{\bpi_\btheta}(s_{t+1}|s_t,a_t)+ \frac{\partial}{\partial \btheta_{s,a}} \reward_{\bpi_\btheta}(s_t, a_t)\Big)\Big]\\
        &= \expectation_{\tau \sim \prob_{\bpi_\btheta}^{\bpi_\btheta}} \Big[ \sum_{t=0}^{\infty} \gamma^t \Big(A_{\bpi_\btheta}^{\bpi_\btheta}(s_t, a_t) (\indicator[s_t=s,a_t=a]\\
        &\qquad\qquad-\bpi_\btheta(a|s)\indicator[s_t=s] ) A_{\bpi_\btheta}^{\bpi_\btheta}(s_t, a_t)  \psi(s_{t+1}) \indicator[s_t=s,a_t=a]\left(1 - \transition_{\bpi_\btheta}(s_{t+1}|s,a)\right) + \xi \indicator[s_t=s, a_t=a]\Big)\Big]\\
        \underset{(a)}{\geq}& \expectation_{\tau \sim \prob_{\bpi_\btheta}^{\bpi_\btheta}} \Big[ \sum_{t=0}^{\infty} \gamma^t  A_{\bpi_\btheta}^{\bpi_\btheta}(s_t, a_t) \indicator[s_t=s,a_t=a]\Big] 
        - \expectation_{\tau \sim \prob_{\bpi_\btheta}^{\bpi_\btheta}} \Big[ \sum_{t=0}^{\infty} \gamma^t \bpi_\btheta(a|s) \indicator[s_t=s] A_{\bpi_\btheta}^{\bpi_\btheta}(s_t, a_t)\Big]\\
        &\qquad\qquad+ \expectation_{\tau \sim \prob_{\bpi_\btheta}^{\bpi_\btheta}} \Big[ \sum_{t=0}^{\infty} \gamma^t \xi \indicator[s_t=s, a_t=a]\Big]\\
        \underset{(b)}{=}& \frac{1}{1-\gamma}d_{\bpi_\btheta,\brho}^{\bpi_\btheta}(s,a) A_{\bpi_\btheta}^{\bpi_\btheta}(s, a) 
        + \frac{1}{1-\gamma}\xi d_{\bpi_\btheta,\brho}^{\bpi_\btheta}(s,a)
\end{align*}  
(a) is due to the fact that $1- \transition_{\bpi_\btheta}(s,a) \geq 0$ for all $s,a$. 
(b) is due to $\expectation_{\tau \sim \prob_{\bpi_\btheta}^{\bpi_\btheta}} \Big[ \sum_{t=0}^{\infty} \gamma^t \bpi_\btheta(a|s)\indicator[s_t=s] A_{\bpi_\btheta}^{\bpi_\btheta}(s_t, a_t)\Big] = 0$.



\end{proof}

\begin{replemma}{lemma:grad_domm_softmax}{Performative Gradient Domination for Exponential PeMDPs}
	Let us consider PeMDPs with softmax policies, linear rewards and exponential transitions.

(a) For unregularised value function,
	\begin{align} \label{eq:grad_domm_softmax}
		 \subopt(\bpi_{\btheta})
		&\leq \sqrt{|\cS||\cA|} \Cov \|\nabla_\btheta V^{\bpi_\btheta}_{\bpi_\btheta}(\bnu)\|_2 + \frac{\gamma}{(1-\gamma)^2} R_{\max}\psi_{\max} \,.
	\end{align}
\end{replemma}
\begin{proof}[Proof of Lemma \ref{lemma:grad_domm_softmax}-- Part (a)]
This proof is divided into two parts. In the first part we bound the expected advantage term from Lemma~\ref{lemma:perf_grad_upperbound} with the norm of the gradient of value function. During this step, we need to express the expected advantage as a linear combination of the advantage itself and the occupancy measure over all states and actions like in equation \eqref{eq:simple_gradient_softmax}. The expectation however is taken w.r.t the occupancy measure $\occupancy{\bpi_\btheta}{\opt_o}$, thus we need to perform a change of measure which introduces a coverage term as shown below. In the second step we directly use the bound of rewards and transitions obtained from their Lipchitzness in lemma \ref{lemma:perf_grad_upperbound}. We know by Lemma~\ref{lemma:perf_perf_diff} that
\begin{align*}
    \subopt(\bpi_{\btheta})
    = &~\frac{1}{1-\gamma} \expectation_{(s,a) \sim \occupancy{\bpi_\btheta,\brho}{\opt_o}(\cdot\mid \brho)}[A_{\bpi_\btheta}^{\bpi_\btheta}(s,a)] \notag\\
    &+ \qquad \frac{1}{1-\gamma} \expectation_{(s,a)\sim \occupancy{\bpi_\btheta,\brho}{\opt_o}}\Big[(\reward_{\opt_o}(s,a) - \reward_{\bpi_\btheta}(s,a)) + \gamma (\transition_{\opt_o}(\cdot|s,a) - \transition_{\bpi_\btheta}(\cdot|s,a))^\top V^{\opt_o}_{\opt_o}(\cdot) \Big] \,.
\end{align*}

\textbf{Step 1: Upper bounding Term 1.}  
\begin{align} \label{eq:advantage_bound}
        \text{Term 1} \defn \expectation_{(s,a)\sim \occupancy{\bpi_\btheta,\brho}{\opt_o}}[A_{\bpi_\btheta}^{\bpi_\btheta}(s,a)] 
        = \sum_{s,a} \occupancy{\bpi_\btheta}{\opt_o}(s,a|\brho)A^{\bpi_\btheta}_{\bpi_\btheta}(s,a)
        &= \sum_{s,a} \frac{\occupancy{\bpi_\btheta}{\opt_o}(s,a|\brho)}{\occupancy{\bpi_\btheta}{\bpi_\btheta}(s,a|\nu)}\occupancy{\bpi_\btheta}{\bpi_\btheta}(s,a|\nu)A^{\bpi_\btheta}_{\bpi_\btheta}(s,a) \nonumber\\
        &\leq \Bigg\|\frac{\occupancy{\bpi_\btheta,\brho}{\opt_o}}{\occupancy{\bpi_\btheta, \nu}{\bpi_\btheta}}\Bigg\|_{\infty} \sum_{s,a} \occupancy{\bpi_\btheta}{\bpi_\btheta}(s,a|\nu)A^{\bpi_\btheta}_{\bpi_\btheta}(s,a)
    \end{align}
    Now, we leverage the gradient of softmax performative MDPs to obtain
    \begin{align*}
        \sum_{s,a} \occupancy{\bpi_\btheta}{\bpi_\btheta}(s,a|\nu)A^{\bpi_\btheta}_{\bpi_\btheta}(s,a) &\leq  (1-\gamma) \sum_{s,a} \frac{\partial V^{\bpi_\btheta}_{\bpi_\btheta}(\nu)}{\partial \btheta_{s,a}}- \xi\\
        &=  (1-\gamma)  \mathbf{1}^\top \nabla_\btheta V^{\bpi_\btheta}_{\bpi_\btheta}(\nu) - \xi \\
        &\leq  (1-\gamma) \sqrt{|\cS||\cA|} \|\nabla_\btheta V^{\bpi_\btheta}_{\bpi_\btheta}(\nu)\|_2 -  \xi
    \end{align*}
    The last inequality is obtained by applying Cauchy-Schwarz inequality.
    
    Now, substituting the above result back in Equation~\eqref{eq:advantage_bound}, we get
    \begin{align}\label{eq:term1_lemma3}
        \frac{1}{1-\gamma}\expectation_{(s,a) \sim \occupancy{\bpi_\btheta,\brho}{\opt_o}(\cdot\mid s_0)}[A_{\bpi_\btheta}^{\bpi_\btheta}(s,a)] &\leq \sqrt{|\cS||\cA|} \Bigg\|\frac{\occupancy{\bpi_\btheta,\brho}{\opt_o}}{\occupancy{\bpi_\btheta, \nu}{\bpi_\btheta}}\Bigg\|_{\infty} \|\nabla_\btheta V^{\bpi_\btheta}_{\bpi_\btheta}(\nu)\|_2 - \Bigg\|\frac{\occupancy{\bpi_\btheta,\brho}{\opt_o}}{\occupancy{\bpi_\btheta, \nu}{\bpi_\btheta}}\Bigg\|_{\infty} \frac{\xi}{1-\gamma}
    \end{align}

    \textbf{Step 2: Upper bounding Term 2.} For softmax rewards and transitions, we further obtain from Lemma \ref{lemma:perf_grad_upperbound},
     \begin{align}
        \text{Term }2 &\defn  \frac{1}{1-\gamma} \expectation_{(s,a)\sim \occupancy{\bpi_\btheta,\brho}{\opt_o}}\Big[(\reward_{\opt_o}(s,a) - \reward_{\bpi_\btheta}(s,a)) 
    + \gamma (\transition_{\opt_o}(\cdot|s,a) - \transition_{\bpi_\btheta}(\cdot|s,a))^\top V^{\opt_o}_{\opt_o}(\cdot) \Big] \,\notag\\
    &\leq \frac{1}{1-\gamma} (\xi + \frac{\gamma}{1-\gamma} R_{\max}\psi_{\max} )\|\opt_o(\cdot|\brho)-\bpi_\btheta(\cdot|\brho)\|_\infty\notag\\
    &\leq \frac{1}{1-\gamma} (\xi + \frac{\gamma}{1-\gamma} R_{\max}\psi_{\max}) \,.\label{eq:term2_lemma3}
    \end{align}

    \textbf{Step 3:} Now, if we use Equation~\eqref{eq:term1_lemma3} and~\eqref{eq:term2_lemma3} together, we get
    \begin{align*}
        \subopt(\bpi_{\btheta}) &\leq \sqrt{|\cS||\cA|} \Bigg\|\frac{\occupancy{\bpi_\btheta,\brho}{\opt_o}}{\occupancy{\bpi_\btheta, \nu}{\bpi_\btheta}}\Bigg\|_{\infty} \|\nabla_\btheta V^{\bpi_\btheta}_{\bpi_\btheta}(\nu)\|_2 +\left(1- \Bigg\|\frac{\occupancy{\bpi_\btheta,\brho}{\opt_o}}{\occupancy{\bpi_\btheta, \nu}{\bpi_\btheta}}\Bigg\|_{\infty}\right) \frac{\xi}{1-\gamma} \notag\\
        &+ \frac{\gamma}{(1-\gamma)^2} R_{\max}\psi_{\max}\,\\
        &\leq \sqrt{|\cS||\cA|} \Bigg\|\frac{\occupancy{\bpi_\btheta,\brho}{\opt_o}}{\occupancy{\bpi_\btheta, \nu}{\bpi_\btheta}}\Bigg\|_{\infty} \|\nabla_\btheta V^{\bpi_\btheta}_{\bpi_\btheta}(\nu)\|_2  + \frac{\gamma}{(1-\gamma)^2} R_{\max}\psi_{\max}\\
    \end{align*}
    The last inequality is true since $\Bigg\|\frac{\occupancy{\bpi_\btheta,\brho}{\opt_o}}{\occupancy{\bpi_\btheta, \nu}{\bpi_\btheta}}\Bigg\|_{\infty} \geq 1$ (Lemma~\ref{lemma:coverage_lower_bound}). For the final step, we use $\Bigg\|\frac{\occupancy{\bpi_\btheta,\brho}{\opt_o}}{\occupancy{\bpi_\btheta, \nu}{\bpi_\btheta}}\Bigg\|_{\infty} \leq \Cov$.
    \end{proof}
    \textbf{Note:} Since we are working with exponential family transitions, we can substitute $\psi_{\max} = \bigO\left(\frac{1-\gamma}{\gamma}\right)$ in \eqref{eq:grad_domm_softmax} to get the final version of the unregularised gradient domination reported in the main paper.

\begin{theorem}[Convergence of \framework{}~for unregularised Exponential Family PeMDPs ]\label{thm:conv_softmax_a}
    Let ${\rm Cov} \triangleq \max_{\btheta, \nu} \Bigg\|\frac{\occupancy{\bpi_\btheta,\brho}{\opt_o}}{\occupancy{\bpi_\btheta,\nu}{\bpi_\btheta}}\Bigg\|_{\infty}$. The gradient ascent algorithm on $V^{\bpi_\btheta}_{\bpi_\btheta}(\brho)$ (Equation \eqref{eqn:ascent_step}) with step size 
    $\eta = \Omega(\min\lbrace \frac{(1-\gamma)^2}{\gamma |\cA|}, \frac{(1-\gamma)^3}{\gamma^2}\rbrace)$ satisfies, for all distributions $\brho \in \Delta(\mathcal{S})$. Then, for unregularised case,
\begin{align*}\min_{t < T} \subopt(\bpi_{\btheta_t})  \;\; \leq \;\; \epsilon + \bigO\left(\frac{1}{1-\gamma}\right)\\
\end{align*}
  when 
  $$T = \Omega\left(\frac{|\cS||\cA|{\rm  Cov}^2}{\epsilon^2} \max\Bigg\lbrace\frac{\gamma R_{\max}\mid\cA\mid}{(1-\gamma)^3}, \frac{\gamma^2}{(1-\gamma)^4} \Bigg\rbrace\right)\, $$
\end{theorem}
\begin{proof}
We proceed with this proof by dividing it in four steps. In the first step, we use the smoothness of the value function to prove an upper bound for the minimum squared gradient norm of the value over time which is a constant times $1/T$ . In the second step, we derive a lower bound on the norm of gradient of value function using Lemma \ref{lemma:grad_domm_softmax}. In the final two steps, we combine the bounds obtained from the first two steps to derive lower bounds for $T$ and $\epsilon$, i.e. the error threshold. 

\textbf{Step 1:} As $V^{\bpi_\btheta}_{\bpi_\btheta}$ is $L$-smooth (Lemma~\ref{lemm:perf_smooth}), it satisfies  
    \begin{align*}
    \Big|V^{\bpi_\btheta}_{\bpi_\btheta}(\brho) - V^{\bpi_\btheta'}_{\bpi_\btheta'}(\brho )- \langle \nabla_\btheta V^{\bpi_\btheta}_{\bpi_\btheta}(\brho), \btheta - \btheta'\rangle \Big| \leq \frac{L}{2}\|\btheta- \btheta'\|^2
\end{align*}


Thus, taking $\btheta$ as $\btheta_{t+1}$ and $\btheta'$ as $\btheta_{t}$ and using the gradient ascent expression (Equation~\eqref{eqn:ascent_step}) yields
\begin{align*}
    \Big|V^{\bpi_{\btheta_{t+1}}}_{\bpi_{\btheta_{t+1}}}(\brho) - V^{\bpi_{\btheta_t}}_{\bpi_{\btheta_t}}(\brho) -\eta\|\nabla_{\btheta} V^{\bpi_{\btheta_t}}_{\bpi_{\btheta_t}}(\brho)\|^2\Big| &\leq \frac{L}{2}\|\btheta_{t+1}-\btheta_t\|^2 \\
    \implies\qquad\qquad\qquad\qquad\qquad V^{\bpi_{\btheta_{t+1}}}_{\bpi_{\btheta_{t+1}}}(\brho) - V^{\bpi_{\btheta_t}}_{\bpi_{\btheta_t}}(\brho) &\geq \eta\|\nabla V^{\bpi_{\btheta_t}}_{\bpi_{\btheta_t}}(\brho)\|^2 - \frac{L}{2}\|\btheta_{t+1}-\btheta_t\|^2
\end{align*}

This further implies that
\begin{align}\label{eq:grad_ineq}
   V^{\bpi_{\btheta_{t+1}}}_{\bpi_{\btheta_{t+1}}}(\brho) - V^{\opt_o}_{\opt_o}(\brho) &\geq V^{\bpi_{\btheta_t}}_{\bpi_{\btheta_t}}(\brho) -V^{\opt_o}_{\opt_o}(\brho) + \eta\|\nabla_\btheta V^{\bpi_{\btheta_t}}_{\bpi_{\btheta_t}}(\brho)\|^2 - \frac{L}{2}\|\btheta_{t+1}-\btheta_t\|^2 \nonumber\\
    &= V^{\bpi_{\btheta_t}}_{\bpi_{\btheta_t}}(\brho) -V^{\opt_o}_{\opt_o}(\brho) + \eta(1-\frac{L\eta}{2})\|\nabla V^{\bpi_{\btheta_t}}_{\bpi_{\btheta_t}}(\brho)\|^2 
\end{align}
The last equality is due to Equation~\eqref{eqn:ascent_step}.

Now, telescoping Equation~\eqref{eq:grad_ineq} leads to 
\begin{align}\label{eq:weighted_grad_bound}
    \eta(1-\frac{L\eta}{2})\sum_{t=0}^{T-1} \|\nabla V^{\bpi_{\btheta_t}}_{\bpi_{\btheta_t}}(\brho)\|^2 &\leq \left(V^{\opt_o}_{\opt_o}(\brho) - V^{\bpi_{\btheta_0}}_{\bpi_{\btheta_0}}(\brho) \right) - \left(V^{\opt_o}_{\opt_o}(\brho) - V^{\bpi_{\btheta_T}}_{\bpi_{\btheta_T}}(\brho)  \right) \\
    &\leq \left(V^{\opt_o}_{\opt_o}(\brho) - V^{\bpi_{\btheta_0}}_{\bpi_{\btheta_0}}(\brho) \right)
\end{align}
Since $\sum_{t=0}^{T-1} \|\nabla V^{\bpi_{\btheta_t}}_{\bpi_{\btheta_t}}(\brho)\|^2 \geq  T \min_{t\in [T-1]} \|\nabla V^{\bpi_{\btheta_t}}_{\bpi_{\btheta_t}}(\brho)\|^2$, we obtain 
\begin{align*}
    \min_{t\in [T-1]} \|\nabla V^{\bpi_{\btheta_t}}_{\bpi_{\btheta_t}}(\brho)\|^2 \leq \frac{1}{T\eta\left(1-\frac{L\eta}{2}\right)}\left(V^{\opt_o}_{\opt_o}(\brho) - V^{\bpi_{\btheta_0}}_{\bpi_{\btheta_0}}(\brho) \right) \leq \frac{R_{\max}}{T\eta\left(1-\frac{L\eta}{2}\right)(1-\gamma)}\,.
\end{align*}
The last inequality comes from $V^{\opt_o}_{\opt_o}(\brho) \leq \frac{R_{\max}}{1-\gamma}$ (Assumption~\ref{ass:bounded_r}).

\textbf{Step 2:} We derive from Equation~\eqref{eq:grad_domm_softmax} that
\begin{align*}
 (\subopt(\bpi_{\btheta}))^2
&\leq \Big(\sqrt{|\cS||\cA|} \Bigg\|\frac{\occupancy{\bpi_\btheta,\brho}{\opt_o}}{\occupancy{\bpi_\btheta, \nu}{\bpi_\btheta}}\Bigg\|_{\infty} \|\nabla_\btheta V^{\bpi_\btheta}_{\bpi_\btheta}(\nu)\|_2 + \frac{\gamma R_{\max} }{(1-\gamma)^2} \psi_{\max} \Big)^2 \\
&\leq 2 {|\cS||\cA|} \Bigg\|\frac{\occupancy{\bpi_\btheta,\brho}{\opt_o}}{\occupancy{\bpi_\btheta, \nu}{\bpi_\btheta}}\Bigg\|_{\infty}^2  \|\nabla_\btheta V^{\bpi_\btheta}_{\bpi_\btheta}(\nu)\|_2^2 + \frac{2 \gamma^2 R_{\max}^2}{(1-\gamma)^4} \psi_{\max}^2
\end{align*}
Thus, we further get
\begin{align*}
\min_{t \in [T-1]} (\subopt(\bpi_{\btheta_t}))^2 
&\leq 2 {|\cS||\cA|} \min_{t \in [T-1]} \Bigg\|\frac{\occupancy{\bpi_{\btheta_t},\brho}{\opt_o}}{\occupancy{\bpi_{\btheta_t}, \nu}{\bpi_\btheta}}\Bigg\|_{\infty}^2  \|\nabla_\btheta V^{\bpi_{\btheta_t}}_{\bpi_{\btheta_t}}(\nu)\|_2^2 + \frac{2 \gamma^2 R_{\max}^2}{(1-\gamma)^4} \psi_{\max}^2  \\
&\leq 2 {|\cS||\cA|}{\rm Cov}^2 \min_{t \in [T-1]} \|\nabla_\btheta V^{\bpi_{\btheta_t}}_{\bpi_{\btheta_t}}(\nu)\|_2^2 + \frac{2 \gamma^2 R_{\max}^2}{(1-\gamma)^4} \psi_{\max}^2 \\
&\leq 2 {|\cS||\cA|}{\rm Cov}^2 \frac{R_{\max}}{T\eta\left(1-\frac{L\eta}{2}\right)(1-\gamma)} + \frac{2 \gamma^2 R_{\max}^2}{(1-\gamma)^4} \psi_{\max}^2\,.
\end{align*}
\textbf{Step 3:} Now, we set
\begin{align*}
    \min_{t
    \in [T-1]}(\subopt(\bpi_{\btheta_t}))^2 &\leq 2 {|\cS||\cA|}{\rm Cov}^2 \frac{R_{\max}}{T\eta\left(1-\frac{L\eta}{2}\right)(1-\gamma)} + \frac{2 \gamma^2 R_{\max}^2}{(1-\gamma)^4} \psi_{\max}^2 \\
    &\leq \left(\sqrt{2 {|\cS||\cA|} \frac{R_{\max}}{T\eta\left(1-\frac{L\eta}{2}\right)(1-\gamma)}} {\rm Cov} + \frac{\sqrt{2} \gamma R_{\max}}{(1-\gamma)^2} \psi_{\max}\right)^2\\
    \implies \min_{t
    \in [T-1]}(\subopt(\bpi_{\btheta_t})) &\leq \left(\epsilon + \frac{\sqrt{2}\gamma R_{\max}}{(1-\gamma)^2}  \psi_{\max}\right)\,,
\end{align*}
and solve for $T$ to get
\begin{align}
    T \ge \frac{2 {|\cS||\cA|}{\rm Cov}^2 R_{\max}}{\eta (1-\frac{L\eta}{2}) (1-\gamma) \epsilon^2} \label{eq:T_lower_bound}
\end{align}

Choosing $\eta = \frac{1}{L}$, we get the final expression
\begin{align}
    T \ge \frac{4 L {|\cS||\cA|}{\rm Cov}^2 R_{\max}}{\epsilon^2(1-\gamma)} \,.
\end{align}
for any $\epsilon > 0$ and the smoothness constant  $L=\bigO\left( \max\Bigg\lbrace\frac{\gamma R_{\max}\mid\cA\mid}{(1-\gamma)^2}, \frac{\gamma^2}{(1-\gamma)^3} \Bigg\rbrace \right)$.

Hence, we conclude that for $T = \Omega\left(\frac{|\cS||\cA|}{\epsilon^2} \max\Bigg\lbrace\frac{\gamma R_{\max}\mid\cA\mid}{(1-\gamma)^3}, \frac{\gamma^2}{(1-\gamma)^4} \Bigg\rbrace\right)$ and $\psi_{\max}=\bigO(\frac{1-\gamma}{\gamma})$, 
\begin{align*}
    \min_{t
    \in [T-1]}(\subopt(\bpi_{\btheta_t})) 
    &\leq \epsilon + \bigO\left(\frac{1}{1-\gamma}\right)\,.
\end{align*}

\end{proof}


\subsection{Proofs for Entropy-regularised or Soft Value Function}

\begin{lemma}[Regularized Performative Policy Difference: Upper Bound for PeMDPs with linear rewards, exponential transition and softmax policy class]\label{lemma:reg_perf_grad_upperbound}
  Under Assumption~\ref{ass:bounded_r}, the sub-optimality gap of a policy $\bpi_\btheta$ is 
  \begin{align} \label{eq:reg_perf_grad_upperbound}
    \subopt(\bpi_\btheta|\lambda)
    \leq &~\frac{1}{1-\gamma}\expectation_{(s,a)\sim \occupancy{\bpi_\btheta,\brho}{\opt_o}}[ \softA_{\bpi_\btheta}^{\bpi_\btheta}(s,a)] \nonumber\\
    &+ \frac{1}{1-\gamma} \Big(\xi  + \frac{\gamma}{1-\gamma} \psi_{\max} (R_{\max}+\lambda \log |\cA|)\Big) \notag\\
    &- \frac{\lambda}{1-\gamma} \sum_s \occupancy{\bpi_\btheta}{\opt_o}(s|\brho) \KL{\opt_o(\cdot|s)}{\bpi_\btheta(\cdot|s)}
  \end{align}
\end{lemma}

\begin{proof}
   This lemma follows the same sketch as Lemma \ref{lemma:reg_perf_grad_upperbound_generic} but replacing the constants $\lipreward$ and $\liptransition$ with constants $\xi$ and $\psi_{\max}$ specific to the given choice of rewards and transitions. 
\end{proof}

\begin{lemma}[Regularized Performative Policy gradient for softmax policies and softmax MDPs]\label{lemm:reg_grad_softmax}
    For a class of PeMDPs $\mathcal{M} \defn (\mathcal{S},\mathcal{A},\bpi,\transition_\bpi,\reward_\bpi, \btheta,\brho)$ consider softmax parametrization for policy $\bpi_\btheta \in \Delta(\btheta \in \mathbf{\Theta})$ and transition dynamics $\transition_{\bpi_\btheta}$ and linear parametrization for reward $\reward_{\bpi_\btheta}$. For all $(s,a,s') \in (\mathcal{S},\mathcal{A},\mathcal{S})$, derivative of the expected return w.r.t $\btheta_{s,a}$ satisfies:
    \begin{align}
        \frac{\partial \tilde V^{\bpi_\btheta}_{\bpi_\btheta}(\brho)}{\partial \btheta_{s,a}} &\geq \frac{1}{1-\gamma}\occupancy{\bpi_\btheta,\brho}{\bpi_\btheta}(s,a) \left(\tilde A_{\bpi_\btheta}^{\bpi_\btheta}(s, a) + \xi\right) - \frac{\lambda}{1-\gamma} \occupancy{\bpi_\btheta,\brho}{\bpi_\btheta}(s,a)\log|\cA|\,.
    \end{align}
\end{lemma}

\begin{proof}
    This proof follows the same sketch as the proof of Lemma ~\ref{lemma:grad_domm_softmax}.  However, we get two additional $\lambda$-dependent terms-- (a) one from the log policy term in the soft advantage, and (b) the other from the log policy term in the soft rewards. We then simplify these terms to obtain the final expression.
    
    First, let us note that
    \begin{align}\label{eq:indiv_grad_regularised}
        &\frac{\partial}{\partial \btheta_{s',a'}} \log \bpi_\btheta(a|s) = \indicator[s=s',a=a'] - \bpi_{\btheta}(a'|s')\indicator[s=s'] \notag\\
        &\frac{\partial}{\partial \btheta_{s',a'}} \log \transition_{\bpi_\btheta}(s''|s,a) = \psi(s'') \indicator[s=s',a=a']\left(1 - \transition_{\bpi_\btheta}(s''|s,a)\right) \notag \\
        &\frac{\partial}{\partial \btheta_{s',a'}} \reward_{\bpi_\btheta} (s,a) = \xi \indicator[s=s',a=a']\,.
    \end{align}
    Now, we get from Theorem~\ref{thm:perf_pg_theorem},
    \begin{align*} 
        &\frac{\partial}{\partial \btheta_{s,a}} \tilde V_{\bpi_\btheta}^{\bpi_\btheta}(\brho)\\ 
        =&\expectation_{\tau \sim \prob^{\bpi_\btheta}_{\bpi_\btheta}} \Big[ \sum_{t=0}^{\infty} \gamma^t \Big(\tilde A_{\bpi_\btheta}^{\bpi_\btheta}(s_t, a_t) \frac{\partial}{\partial \btheta_{s,a}}\log \bpi_\btheta(a_t \mid s_t)+ \tilde A_{\bpi_\btheta}^{\bpi_\btheta}(s_t, a_t)  \frac{\partial}{\partial \btheta_{s,a}} \log P_{\bpi_\btheta}(s_{t+1}|s_t,a_t)\\
        &\qquad\qquad+ \frac{\partial}{\partial \btheta_{s,a}} \reward_{\bpi_\btheta}(s_t, a_t) - \lambda \frac{\partial}{\partial \btheta_{s,a}}\log \bpi_\btheta(a_t \mid s_t)\Big)\Big]\\
        =& \expectation_{\tau \sim \prob^{\bpi_\btheta}_{\bpi_\btheta}} \Big[ \sum_{t=0}^{\infty} \gamma^t \Big(\tilde A_{\bpi_\btheta}^{\bpi_\btheta}(s_t, a_t) \left(\indicator[s_t=s,a_t=a]-\bpi_\btheta(a|s)\indicator[s_t=s] \right) + \tilde A_{\bpi_\btheta}^{\bpi_\btheta}(s_t, a_t)  \psi(s_{t+1}) \indicator[s_t=s,a_t=a] \\
        &\qquad\qquad\left(1 - \transition_{\bpi_\btheta}(s_{t+1}|s,a)\right)+ \xi \indicator[s_t=s, a_t=a] - \lambda\indicator[s_t=s,a_t=a]+\lambda\bpi_\btheta(a|s)\indicator[s_t=s]\Big)\Big]\\
        &\underset{(a)}{\geq} \expectation_{\tau \sim \prob^{\bpi_\btheta}_{\bpi_\btheta}} \Big[ \sum_{t=0}^{\infty} \gamma^t  \tilde A_{\bpi_\btheta}^{\bpi_\btheta}(s_t, a_t) \indicator[s_t=s,a_t=a]\Big] - \expectation_{\tau \sim \prob^{\bpi_\btheta}_{\bpi_\btheta}} \Big[ \sum_{t=0}^{\infty} \gamma^t \bpi_\btheta(a|s)\indicator[s_t=s] \tilde A_{\bpi_\btheta}^{\bpi_\btheta}(s_t, a_t)\Big]\\
        &\qquad\qquad+ \expectation_{\tau \sim \prob^{\bpi_\btheta}_{\bpi_\btheta}} \Big[ \sum_{t=0}^{\infty} \gamma^t \xi \indicator[s_t=s, a_t=a]\Big] - \lambda \expectation_{\tau \sim \prob^{\bpi_\btheta}_{\bpi_\btheta}}\Big[ \sum_{t=0}^{\infty} \gamma^t \indicator[s_t=s,a_t=a] \Big] 
        \\
        &\qquad\qquad + \lambda \expectation_{\tau \sim \prob^{\bpi_\btheta}_{\bpi_\btheta}}\Big[ \sum_{t=0}^{\infty} \gamma^t \bpi_\btheta(a|s) \indicator[s_t=s] \Big] \\
        &\underset{(b)}{=} \frac{1}{1-\gamma}d_{\bpi_\btheta,\brho}^{\bpi_\btheta}(s,a) \tilde A_{\bpi_\btheta}^{\bpi_\btheta}(s, a) + \lambda \expectation_{\tau \sim \prob^{\bpi_\btheta}_{\bpi_\btheta}}\Big[ \sum_{t=0}^{\infty} \gamma^t \bpi_\btheta(a|s)\log \bpi_\btheta(a_t|s_t) \indicator[s_t=s] \Big]+ \frac{1}{1-\gamma}\xi d_{\bpi_\btheta,\brho}^{\bpi_\btheta}(s,a)  \\
        &= \frac{1}{1-\gamma}d_{\bpi_\btheta,\brho}^{\bpi_\btheta}(s,a) \tilde A_{\bpi_\btheta}^{\bpi_\btheta}(s, a) 
        +  \lambda \expectation_{\tau \sim \prob^{\bpi_\btheta}_{\bpi_\btheta}}\Big[ \sum_{t=0}^{\infty} \gamma^t \bpi_\btheta(a|s)\log \bpi_\btheta(a_t|s_t) \sum_{a}  \indicator[s_t=s,a_t=a] \Big]
        + \frac{1}{1-\gamma}\xi d_{\bpi_\btheta,\brho}^{\bpi_\btheta}(s,a) \\
        &= \frac{1}{1-\gamma}d_{\bpi_\btheta,\brho}^{\bpi_\btheta}(s,a) \tilde A_{\bpi_\btheta}^{\bpi_\btheta}(s, a) 
        +  \lambda \bpi_\btheta(a|s)\expectation_{\tau \sim \prob^{\bpi_\btheta}_{\bpi_\btheta}}\Big[ \sum_{a} \log \bpi_\btheta(a|s) \sum_{t=0}^{\infty} \gamma^t  \indicator[s_t=s,a_t=a] \Big]
        + \frac{1}{1-\gamma}\xi d_{\bpi_\btheta,\brho}^{\bpi_\btheta}(s,a) \\
        &= \frac{1}{1-\gamma}d_{\bpi_\btheta,\brho}^{\bpi_\btheta}(s,a) \left(\tilde A_{\bpi_\btheta}^{\bpi_\btheta}(s, a) + \xi\right)+ \frac{\lambda}{1-\gamma} \bpi_\btheta(a|s)\sum_{a} \occupancy{\bpi_\btheta,\brho}{\bpi_\btheta}(s,a)   \log \bpi_\btheta(a|s)\,\\
        &\underset{(c)}{\geq} \frac{1}{1-\gamma}d_{\bpi_\btheta,\brho}^{\bpi_\btheta}(s,a) \left(\tilde A_{\bpi_\btheta}^{\bpi_\btheta}(s, a) + \xi\right) - \frac{\lambda}{1-\gamma}\occupancy{\bpi_\btheta,\brho}{\bpi_\btheta}(s,a)\log|\cA|\,.
\end{align*}

(b) holds as:
\begin{align*}
    \expectation_{\tau \sim \prob^{\bpi_\btheta}_{\bpi_\btheta}}\Big[ \sum_{t=0}^{\infty} \gamma^t \indicator[s_t=s,a_t=a] \Big] 
        - \expectation_{\tau \sim \prob^{\bpi_\btheta}_{\bpi_\btheta}}\Big[ \sum_{t=0}^{\infty} \gamma^t \bpi_\btheta(a|s) \indicator[s_t=s] \Big] 
        = d_{\bpi_\btheta,\brho}^{\bpi_\btheta}(s,a) - d_{\bpi_\btheta,\brho}^{\bpi_\btheta}(s)\bpi_\btheta (a|s) = 0
\end{align*}

(c) holds from the following:
\begin{align*}
    -\sum_{a} \occupancy{\bpi_\btheta}{\bpi_\btheta}(s,a|s_0) \log {\bpi_\btheta(a|s)} &= \occupancy{\bpi_\btheta}{\bpi_\btheta}(s|s_0)  \Big( - \sum_{a} \bpi_\btheta(a|s) \log \bpi_\btheta(a|s) \Big) \\
    &\underset{(d)}\leq \occupancy{\bpi_\btheta}{\bpi_\btheta}(s|s_0)  \log |\cA| 
\end{align*}
and (d) holds as entropy is upper bounded by $\log |\cA|$~\citep[Theorem 2.6.4]{Cover2006}. 


\end{proof}    

\begin{replemma}{lemma:grad_domm_softmax}[Regularized Performative Gradient Domination: Part(b) of Lemma~\ref{lemma:grad_domm_softmax}] For regularized PeMDPs the following inequality holds:
\begin{align} 
    \subopt(\bpi_{\btheta}\mid \lambda) \leq \sqrt{|\cS||\cA|} \Cov\|\nabla_\btheta \softV^{\bpi_\btheta}_{\bpi_\btheta}(\bnu)\|_2 +\frac{\lambda\log|\cA|}{1-\gamma}\left( \frac{\gamma\psi_{\max}}{1-\gamma}+2\right)+\frac{\gamma\psi_{\max}R_{\max}}{(1-\gamma)^2}\,.
\end{align}    
\end{replemma}

\begin{proof}
\textbf{Step 1.} First, we observe that
\begin{align*} 
     -\KL{\opt_o(\cdot|s)}{\bpi_\btheta(\cdot|s)} \leq - \sum_{a\in\cA} \opt_o(a|s) \log \opt_o(a|s) \leq \log |\cA|
\end{align*}
 Hence, we get
 \begin{align}\label{eq:kl_bound}
    - \sum_s \occupancy{\bpi_\btheta}{\opt_o}(s|s_0) \KL{\opt_o(\cdot|s)}{\bpi_\btheta(\cdot|s)} \leq \log |\cA|
\end{align}

    \textbf{Step 2.} Using Lemma~\ref{lemm:reg_grad_softmax} and applying Cauchy-Schwarz inequality, we get
    \begin{align}
        \sum_{s,a} \occupancy{\bpi_\btheta}{\bpi_\btheta}(s,a) \softA^{\bpi_\btheta}_{\bpi_\btheta}(s,a) 
        &\leq \sqrt{|\cS||\cA|}{(1-\gamma)}  \|\nabla_\btheta \softV^{\bpi_\btheta}_{\bpi_\btheta}(\bnu)\|_2 - \xi +{\lambda}\log|\cA|
         \label{eq:adv_bound_regularised}
    \end{align}
    
    \textbf{Step 3.} Now, substituting Equation~\eqref{eq:kl_bound} and~\eqref{eq:adv_bound_regularised} in Equation~\eqref{eq:reg_perf_grad_upperbound}, we finally get
    \begin{align*} 
    \subopt(\bpi_{\btheta}\mid \lambda) 
    &\leq {\sqrt{|\cS||\cA|}} \Cov \|\nabla_\btheta \softV^{\bpi_\btheta}_{\bpi_\btheta}(\bnu)\|_2 - \Cov \frac{\xi}{1-\gamma}+ \frac{\lambda}{1-\gamma}\log|\cA|\nonumber\\
    &\qquad\qquad+ \frac{1}{1-\gamma} \Big(\lipreward  + \frac{\gamma}{1-\gamma} \liptransition (R_{\max}+\lambda \log |\cA|)\Big) + \frac{\lambda}{1-\gamma}\log|\cA|\\
    &\underset{(a)}{=} {\sqrt{|\cS||\cA|}} \Cov \|\nabla_\btheta \softV^{\bpi_\btheta}_{\bpi_\btheta}(\bnu)\|_2 -\Cov \frac{\xi}{1-\gamma}+\frac{\lambda}{1-\gamma}2\log|\cA|\nonumber\\
    &\qquad\qquad+ \frac{1}{1-\gamma} \Big( \xi + \frac{\gamma}{1-\gamma} \psi_{\max} (R_{\max} +\lambda \log |\cA|)\Big) \\
    &\underset{(b)}{\leq}  {\sqrt{|\cS||\cA|}} \Cov \|\nabla_\btheta \softV^{\bpi_\btheta}_{\bpi_\btheta}(\bnu)\|_2 \\
    &\qquad\qquad+ \frac{ \gamma}{(1-\gamma)^2} \psi_{\max}(R_{\max} + \lambda \log|\cA|)
    +\frac{\lambda}{1-\gamma}2\log|\cA| \\
    &={\sqrt{|\cS||\cA|}} \Cov \|\nabla_\btheta \softV^{\bpi_\btheta}_{\bpi_\btheta}(\bnu)\|_2 + \frac{\lambda\log|\cA|}{1-\gamma}\left( \frac{\gamma\psi_{\max}}{1-\gamma}+2\right)+\frac{\gamma\psi_{\max}R_{\max}}{(1-\gamma)^2}
\end{align*}

In (a), we substitute the values of $\lipreward$ and $\liptransition$ for softmax PeMDPs, and in (b), we use $\Cov \geq 1$ (Lemma~\ref{lemma:coverage_lower_bound}). 

\end{proof}
\textbf{Note:} Analogous to the unregularised case, we can substitute $\psi_{\max}$ with $\bigO\left(\frac{1-\gamma}{\gamma}\right)$ to get a simpler bound as reported in the main paper.

\begin{theorem}[Convergence of \framework{}~for entropy-regularised Exponential Family PeMDPs ]\label{thm:conv_softmax_b}

    Let ${\rm Cov} \triangleq \max_{\btheta, \nu} \Bigg\|\frac{\occupancy{\bpi_\btheta,\brho}{\opt_o}}{\occupancy{\bpi_\btheta,\nu}{\bpi_\btheta}}\Bigg\|_{\infty}$. The gradient ascent algorithm on $V^{\bpi_\btheta}_{\bpi_\btheta}(\brho)$ (Equation \eqref{eqn:ascent_step}) with step size 
    $\eta = \Omega(\min\lbrace \frac{(1-\gamma)^2}{\gamma |\cA|}, \frac{(1-\gamma)^3}{\gamma^2}\rbrace)$ satisfies, for all distributions $\brho \in \Delta(\mathcal{S})$. Then, for entropy regularised case, if we set $\lambda = \frac{(1-\gamma) R_{\max}}{1+2\log|\cA|}$, we get
$$
    \min_{t < T} \subopt(\bpi_{\btheta_t}\mid \lambda) \;\; \leq \;\; \epsilon +\bigO\left(\frac{1}{1-\gamma}\right) \text{  when  } T = \Omega\left( \frac{R_{\max}|\cS||\cA|^2}{\epsilon^2 (1-\gamma)^3}{\Cov}^2\right), 
$$
\end{theorem}

\begin{proof}

This proof follows similar steps of Theorem~\ref{thm:conv_softmax_a} with two additional changes: (i) We have a $\lambda$, i.e. regularisation coefficient, dependent term due to the entropy regulariser. (ii) The maximum value of the soft value function is $\frac{R_{\max}+\lambda \log|\cA|}{1-\gamma}$ instead of $\frac{R_{\max}}{1-\gamma}$ for the unregularised value function. 

\textbf{Step 1:} From Equation~\eqref{eq:smoothness_constant_regularised}, we observe that the soft-value function $\softV^{\bpi_\btheta}_{\bpi_\btheta}$ is $L_\lambda$-smooth. 

Thus, following the Step 1 of Theorem~\ref{thm:conv_softmax_a}, we get 
\begin{align}
    \min_{t\in [T-1]} \|\nabla \softV^{\bpi_{\btheta_t}}_{\bpi_{\btheta_t}}(\brho)\|^2 
    &\leq \frac{1}{T\eta\left(1-\frac{L_{\lambda}\eta}{2}\right)}\left(\softV^{\opt_o}_{\opt_o}(\brho) - \softV^{\bpi_{\btheta_0}}_{\bpi_{\btheta_0}}(\brho) \right)\notag\\
    &\leq \frac{R_{\max}+\lambda \log|\cA|}{T\eta\left(1-\frac{L_{\lambda}\eta}{2}\right)(1-\gamma)}\,.\label{eq:softmax_grad_bound}
\end{align}
The last inequality is true due to the fact that $\subopt(\bpi_{\btheta_0}) \leq \softV^{\opt_o}_{\opt_o}(\brho) \leq \frac{R_{\max}+\lambda \log|\cA|}{1-\gamma}$. 

\textbf{Step 2:} Now, from Part (b) of Lemma~\ref{lemma:grad_domm_softmax}, we obtain that
\begin{align*}
  &\min_{t\in [T-1]} \left(\subopt(\bpi_{\btheta_t}\mid \lambda)\right)^2 \nonumber\\
  &\leq \min_{t\in [T-1]} \Bigg(\sqrt{|\cS||\cA|} \Bigg\|\frac{\occupancy{\bpi_{\btheta_t},\brho}{\opt_o}}{\occupancy{\bpi_{\btheta_t},\bnu}{\bpi_{\btheta_t}}}\Bigg\|_{\infty} \|\nabla_\btheta \softV^{\bpi_{\btheta_t}}_{\bpi_{\btheta_t}}(\bnu)\|_2 
  + \frac{\lambda\log|\cA|}{1-\gamma}\left( \frac{\gamma\psi_{\max}}{1-\gamma}+2\right)+\frac{\gamma\psi_{\max}R_{\max}}{(1-\gamma)^2}  \Bigg)^2\\
  &\leq 2{|\cS||\cA|} \min_{t\in [T-1]} \Bigg\|\frac{\occupancy{\bpi_{\btheta_t},\brho}{\opt_o}}{\occupancy{\bpi_{\btheta_t},\bnu}{\bpi_{\btheta_t}}}\Bigg\|_{\infty}^2 \|\nabla_\btheta \softV^{\bpi_{\btheta_t}}_{\bpi_{\btheta_t}}(\bnu)\|_2^2 
  \\
  &+ 2\Bigg(\frac{\lambda\log|\cA|}{1-\gamma}\left( \frac{\gamma\psi_{\max}}{1-\gamma}+2\right)+\frac{\gamma\psi_{\max}R_{\max}}{(1-\gamma)^2} \Bigg)^2\\
  &\leq \frac{2{|\cS||\cA|} {\Cov}^2 \left(R_{\max}+\lambda \log|\cA|\right)}{T\eta\left(1-\frac{L_{\lambda} \eta}{2}\right)(1-\gamma)} \\
  &+ 2\left(\frac{\lambda\log|\cA|}{1-\gamma}\left( \frac{\gamma\psi_{\max}}{1-\gamma}+2\right)+\frac{\gamma\psi_{\max}R_{\max}}{(1-\gamma)^2}\right)^2\,.
\end{align*}
The last inequality is due to the upper bound on the minimum gradient norm as in Equation~\eqref{eq:softmax_grad_bound} and by definition of the coverage parameter ${\Cov}$.

Thus, we conclude that
\begin{align}
&\min_{t\in [T-1]} \subopt(\bpi_{\btheta_t}\mid \lambda) \nonumber\\
&\leq \sqrt{\frac{2{|\cS||\cA|} {\Cov}^2 \left(R_{\max}+\lambda \log|\cA|\right)}{T\eta\left(1-\frac{L_{\lambda} \eta}{2}\right)(1-\gamma)}} \\
&+ \sqrt{2}\left(\frac{\lambda\log|\cA|}{1-\gamma}\left( \frac{\gamma\psi_{\max}}{1-\gamma}+2\right)+\frac{\gamma\psi_{\max}R_{\max}}{(1-\gamma)^2} \right)\,.\label{eq:reg_conv_before_choosing_T}
\end{align}



\textbf{Step 4:} Now, by setting the $T$-dependent term in Equation~\eqref{eq:reg_conv_before_choosing_T} to $\epsilon$, we get $
    T \geq \frac{2{|\cS||\cA|} {\Cov}^2 \left(R_{\max}+\lambda \log|\cA|\right)}{\eta\left(1-\frac{L_{\lambda} \eta}{2}\right)(1-\gamma)\epsilon^2}\,.$
    
Choosing $\eta = \frac{1}{L_{\lambda}}$, $\lambda = \frac{(1-\gamma) R_{\max}}{(1+2\log|\cA|)}$, and $\psi_{\max}=\bigO(\frac{1-\gamma}{\gamma})$, we get the final expression $T \geq \frac{8{|\cS||\cA|} {\Cov}^2 L_{\lambda} R_{\max}}{(1-\gamma)\epsilon^2}\,$,
and
\begin{align*}
\min_{t\in [T-1]} \softV^{\opt_o}_{\opt_o}(\brho) - \softV^{\bpi_{\btheta_t}}_{\bpi_{\btheta_t}}(\brho) 
&\leq \epsilon + \bigO(\frac{1}{1-\gamma})\,.
\end{align*}

Finally, noting that $L_{\lambda} = \bigO\left( \max\Bigg\lbrace\frac{\gamma R_{\max}\mid\cA\mid \psi_{\max}^2}{(1-\gamma)^2}, \frac{R_{\max} \psi_{\max}^2}{(1-\gamma)^2} \Bigg\rbrace \right)\,$, we get
\begin{align*}
    T = \Omega \left( \frac{|\cS||\cA|}{\epsilon^2(1-\gamma)^3} \max\lbrace 1, \gamma |\cA| \rbrace\right)\,.
\end{align*}
\end{proof}

 \clearpage

\section{Technical Lemmas}

\begin{lemma}[Lower Bound of Coverage] \label{lemma:coverage_lower_bound}
    For any $\bpi,\bpi' \in \Pi(\Theta)$, the following non-trivial lower bound holds,
    \begin{align*}
        \Bigg\|\frac{\occupancy{\bpi'}{}}{\occupancy{\bpi}{}}\Bigg\|_{\infty} \geq 1
    \end{align*}
\end{lemma}

\begin{proof}
    \begin{align*}
        \Bigg\|\frac{\occupancy{\bpi'}{}}{\occupancy{\bpi}{}}\Bigg\|_{\infty}  = \max_{s,a} \frac{\occupancy{\bpi'}{}(s,a)}{\occupancy{\bpi}{}(s,a)} \geq \frac{1}{\sum_{s,a}w_{s,a}}\sum_{s,a} \frac{\occupancy{\bpi'}{}(s,a)}{\occupancy{\bpi}{}(s,a)} \cdot w_{s,a}
    \end{align*}
    Choose $w_{s,a} = \occupancy{\bpi}{}(s,a)$
    Hence, we get,
    \begin{align*}
        \max_{s,a} \frac{\occupancy{\bpi'}{}(s,a)}{\occupancy{\bpi}{}(s,a)} \geq \frac{\sum_{s,a}\occupancy{\bpi'}{}(s,a)}{\sum_{s,a}\occupancy{\bpi}{}(s,a)} = 1
    \end{align*}
    The last equality holds from the fact that the state-action occupancy measure is a distribution over $\cS \times \cA$. Hence, $\sum_{s,a}\occupancy{\bpi'}{}(s,a)= \sum_{s,a}\occupancy{\bpi}{}(s,a)$
\end{proof}

\begin{lemma} \label{lemm:state_occupancy_distribution}
    The discounted state occupancy measure
    \begin{align*}
       \occupancy{\bpi'}{\bpi}(s|s_0) \defn (1-\gamma)\,\expectation_{\tau\sim\prob_{\bpi'}^{\bpi}}\Big[\sum_{t=0}^{\infty}\gamma^t\indicator\{s_t=s\}\Big]
    \end{align*}
    is a probability mass function over the state-space $\cS$. 
\end{lemma}

\begin{proof}
For each fixed $s$ the integrand $\sum_{t=0}^\infty\gamma^t\indicator\{s_t=s\}\ge0$, hence $\occupancy{\bpi'}{\bpi}(s|s_0)\ge0$.

To check normalization, we sum over all states and use Tonelli/Fubini (permitted because the summand is non-negative) to exchange sums and expectation:
\begin{align*}
\sum_{s\in\mathcal S}\occupancy{\bpi'}{\bpi}(s|s_0)
&=(1-\gamma)\,\expectation_{\tau\sim\prob_{\bpi'}^{\bpi}(\cdot|s_0)}\Big[\sum_{t=0}^\infty\gamma^t\sum_{s\in\mathcal S}\indicator\{s_t=s\}\Big]\\
&=(1-\gamma)\,\expectation_{\tau\sim\prob_{\bpi'}^{\bpi}(\cdot|s_0)}\Big[\sum_{t=0}^\infty\gamma^t\cdot 1\Big]
=(1-\gamma)\sum_{t=0}^\infty\gamma^t = 1.
\end{align*}
Therefore $\brho$ is a probability mass function on $\cS$.
\end{proof}

A similar argument holds for the discounted state-action occupancy measure $\occupancy{\bpi'}{\bpi}(s,a|s_0)$ as well.  \clearpage
\section{Experiment Details} \label{app:exp_details}

\textbf{Environment.} We evaluate \framework{} in the Gridworld test-bed~\citep{mandal2023performativereinforcementlearning}, which has become a benchmark in performative RL. This environment consists of a grid where two agents $A_1$ (the principal) and $A_2$ (the follower), jointly control an actor navigating from start positions (S) to the goal (G) while avoiding hazards. 
The environment dynamics are as follows: Agent $A_1$ proposes a control policy for the actor by selecting one of four directional actions. Agent $A_2$ can either accept this action (not intervene) or override it with its own directional choice. \textit{This creates a performative environment for $A_1$, as its effective policy outcomes depend on $A_2$'s responses to its deployed strategy.}

The cost structure follows: visiting blank cells (S) incurs penalty of $-0.01$, goal cells (F) cost $-0.02$, hazard cells (H) impose a severe penalty of $-0.5$, and any intervention by $A_2$ results in an additional cost of $-0.05$ for the intervening agent. 
The response model also follows that of~\cite{mandal2023performativereinforcementlearning}, i.e., the agent $A_2$ responds to $A_1$'s policy using a Boltzmann softmax operator. 
Given $A_1$'s current policy $\bpi_1$, we compute the optimal Q-function $Q^{*|\bpi_1}$ for each follower agent $A_j$ relative to a perturbed version of the grid world, where each cell types matches $A_1$'s environment with probability $0.7$. 
We then define an average Q-function over the follower agents and determine the collective response policy via Boltzmann softmax
$Q^{*|\pi_1}(s,a) = \frac{1}{n} \sum_{j=2}^{n+1} Q^{*|\pi_1}_j(s,a),    \pi_2(a|s) = \frac{\exp(\beta \cdot Q^{*|\pi_1}(s,a))}{\sum_{a'} \exp(\beta \cdot Q^{*|\pi_1}(s,a'))}$. 

Note that our experimental setup deliberately uses the immediate response model from the original performative RL framework, rather than the gradually shifting environment of~\cite{rank2024performative} that assumes slow shifts in the environment. 
Our choice to use the immediate response model presents a more challenging performative setting where the environment responds instantaneously to policy changes. This allows us to demonstrate that unlike MDRR~\citep{rank2024performative}, \framework~can handle the fundamental performative challenge without requiring environmental assumptions that artificially slows down the feedback loop, thereby highlighting the robustness of the proposed \framework~approach.

\textbf{Experimental Setup.} 
We evaluate \framework~(with and without entropy regularisation) alongside Mixed Delayed Repeated Retraining (MDRR), which represents the current state-of-the-art in performative reinforcement learning under gradually shifting environments~\citep{rank2024performative}, and Repeated Policy Optimization with Finite Samples (RPO FS). MDRR has demonstrated significant improvements over traditional repeated retraining methods, by leveraging historical data from multiple deployments, while RPO FS is included as the baseline method from \citep{mandal2023performativereinforcementlearning} for direct comparison with the original performative RL approach.

All experiments use a $8 \times 8$ grid with $\gamma = 0.9$, exploration parameter $\epsilon = 0.5$ for initial policy construction, one follower agent $A_2$, and 100 trajectory samples per iteration. The algorithms share common parameters of $T = 1000$ iterations. For regularization, RPO FS and MDRR use $\lambda = 0.1$ from their original experiments, while entropy-regularized \framework{} uses $\lambda = 2.0$ (ablation studies for this choice are provided in the appendix). \framework~uses learning rate $\eta = 0.1$, MDRR employs memory weight $v=1.1$ for historical data utilization, delayed round parameter $k = 3$, and FTRL parameters $N = B = 10$, while RPO FS follows the finite-sample optimization from \citep{mandal2023performativereinforcementlearning}.

\textbf{Computational Cost.} 
We report the wall-clock runtime of each algorithm on a single CPU.
Table~\ref{tab:runtime} summarizes the per-iteration cost and the total
runtime extrapolated to 200 iterations based on measurements over 50
iterations.

\begin{table}[h]
\centering
\caption{Runtime comparison of algorithms (single CPU).}
\label{tab:runtime}
\begin{tabular}{lcc}
\toprule
Algorithm & ms/iter & Total (200 iter) \\
\midrule
\framework~(Regularized) & 10{,}635 & $\sim$35 min \\
\framework & 11{,}664 & $\sim$39 min \\
RPO FS~\citep{mandal2023performativereinforcementlearning} & 13{,}039 & $\sim$43 min \\
MDRR~\citep{rank2024performative} & 24{,}317 & $\sim$81 min \\
\bottomrule
\end{tabular}
\end{table}

All algorithms were run on CPU with identical environment configurations.
\framework~(with and without regularization) is the most efficient, with per-iteration costs roughly
half that of MDRR. RPO sits between the two groups.

\section{Ablation Studies}\label{app:ablation}
\subsection{Entropy regularisation}\label{app:ablation_er}
\begin{figure}[h!]
    \centering    \includegraphics[width=0.5\linewidth]{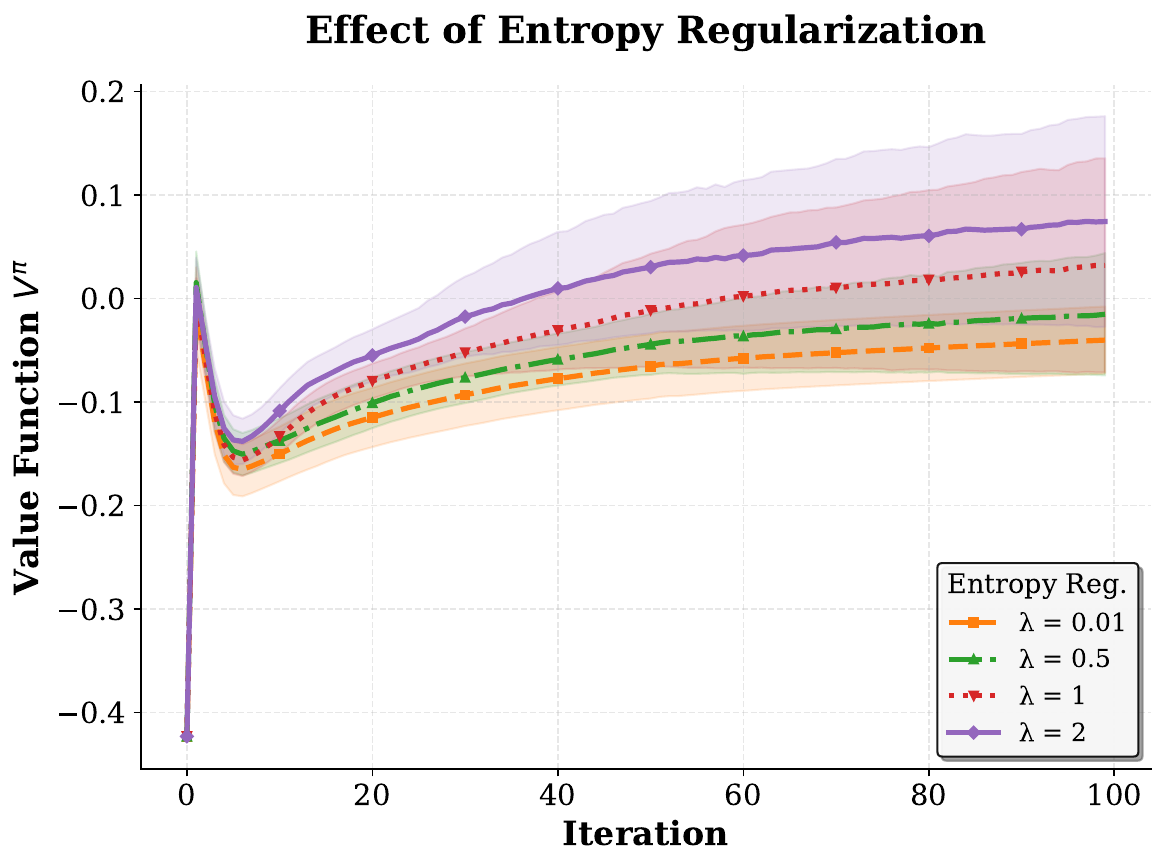}
    \caption{Ablation study for \framework~ for different values of regularised $\lambda$ with 20 random seeds, each for 100 iterations}
    \label{fig:ablation_lambda}
\end{figure}
We conducted an ablation study across four entropy regularization strengths ($\lambda \in \{0.01, 0.5, 1, 2\}$ to determine the optimal balance between exploration and convergence stability in entropy regularised \framework{}. The results demonstrate that $\lambda = 2$ achieves the highest final performance (~0.05), while smaller values ($\lambda \leq 1)$ converge to similar suboptimal levels around $-0.01$ to $0$, indicating that stronger entropy regularization enables more effective exploration of the policy space in performative settings.

\subsection{Learning rate}\label{app:ablation_lr}
\begin{figure}[h!]
    \centering    \includegraphics[width=0.5\linewidth]{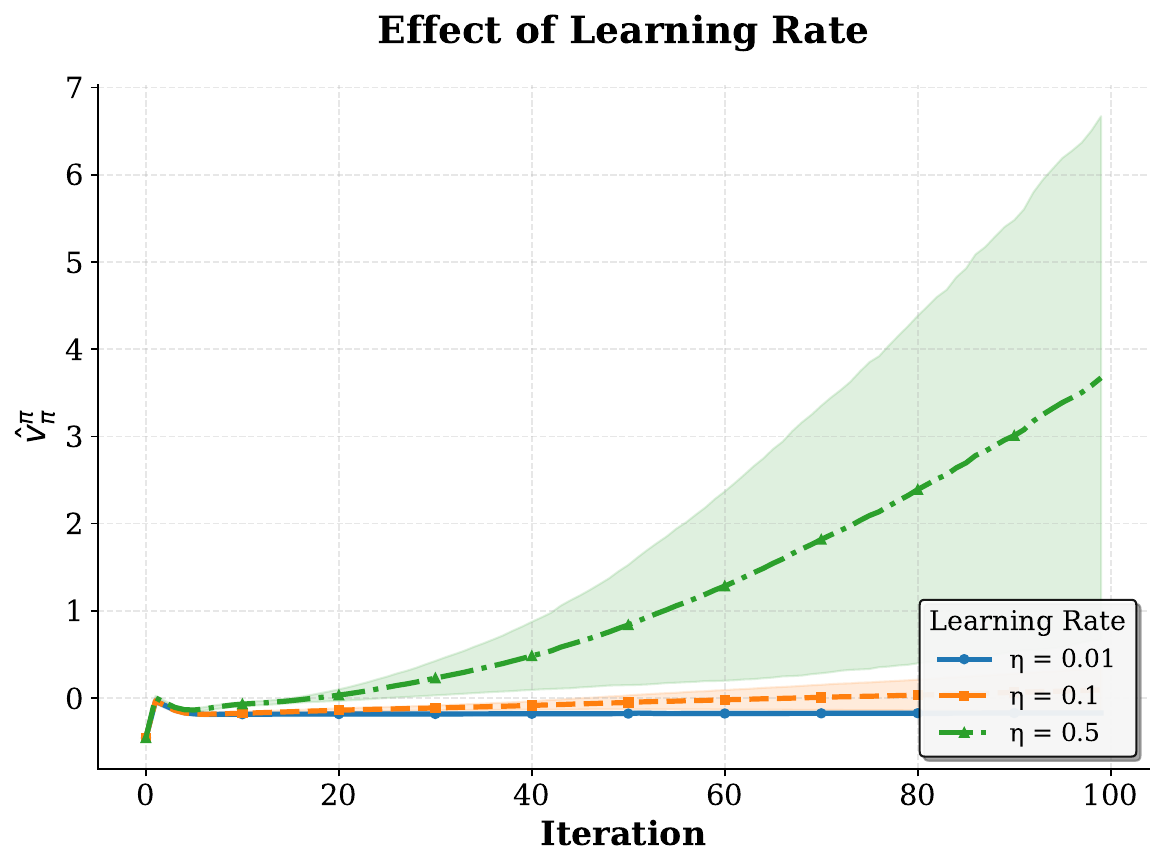}
    \caption{Ablation study for \framework~ for different values of $\eta$ with 20 random seeds across 100 iterations}
    \label{fig:ablation_eta}
\end{figure}

 We additionally performed an ablation study on the learning rate, considering $\eta \in \{0.01,0.1,0.5\}$, to examine its effect on convergence behavior and performance stability in \framework. The results indicate that the largest learning rate ($\eta= 0.5$) attains the highest final performance ($\hat{V_{\bpi}^{\bpi}}=4$); however, it is also accompanied by substantially higher variance across runs. In contrast, smaller learning rates yield more stable learning dynamics but converge to comparatively lower performance levels. This highlights the classical trade-off between fast convergence and stability in policy optimization.


\end{document}